\documentclass[letterpaper]{article} % DO NOT CHANGE THIS

\PassOptionsToPackage{table,dvipsnames}{xcolor} % 需要什么选项就写什么

\usepackage{aaai2026}  % DO NOT CHANGE THIS
\usepackage{times}  % DO NOT CHANGE THIS
\usepackage{helvet}  % DO NOT CHANGE THIS
\usepackage{courier}  % DO NOT CHANGE THIS
\usepackage[hyphens]{url}  % DO NOT CHANGE THIS
\usepackage{graphicx} % DO NOT CHANGE THIS
\usepackage{natbib}  % DO NOT CHANGE THIS AND DO NOT ADD ANY OPTIONS TO IT
\usepackage{caption} % DO NOT CHANGE THIS AND DO NOT ADD ANY OPTIONS TO IT
\usepackage{algorithm}
\usepackage{algorithmic}

\usepackage{amsmath}
\usepackage{amssymb}

\usepackage{caption}
\usepackage{subcaption}

\usepackage{tikz}
\usetikzlibrary{matrix}

\usepackage{booktabs}
\usepackage{tabularx}

\usepackage{newfloat}
\usepackage{listings}
\DeclareCaptionStyle{ruled}{labelfont=normalfont,labelsep=colon,strut=off} % DO NOT CHANGE THIS
\floatstyle{ruled}
\newfloat{listing}{tb}{lst}{}
\floatname{listing}{Listing}
\title{Rhetorical Text-to-Image Generation via Two-layer Diffusion Policy Optimization}

\author{%
  Yuxi Zhang\textsuperscript{1} \quad
  Yueting Li\textsuperscript{2} \quad
  Xinyu Du\textsuperscript{3} \quad
  Sibo Wang\textsuperscript{3} \\[1ex]
  \textsuperscript{1}The Chinese University of Hong Kong, Shenzhen \quad
  \textsuperscript{2}University of California, Berkeley \\
  \textsuperscript{3}The Chinese University of Hong Kong \\[1ex]
}

\begin{document}

\maketitle

\begin{abstract}
Generating images from rhetorical languages remains a critical challenge for text-to-image models. Even state-of-the-art (SOTA) multimodal large language models (MLLM) fail to generate images based on the hidden meaning inherent in rhetorical language—despite such content being readily mappable to visual representations by humans. A key limitation is that current models emphasize object-level word embedding alignment, causing rhetorical expressions to steer image generation towards their literal visuals and overlook the intended semantic meaning. To address this, we propose Rhet2Pix, a framework that formulates rhetorical text-to-image generation as a multi-step policy optimization problem, incorporating a two-layer MDP diffusion module. In the outer layer, Rhet2Pix converts the input prompt into incrementally elaborated sub-sentences and executes corresponding image-generation actions, constructing semantically richer visuals. In the inner layer, Rhet2Pix mitigates reward sparsity during image generation by discounting the final reward and optimizing every adjacent action pair along the diffusion denoising trajectory. Extensive experiments demonstrate the effectiveness of Rhet2Pix in rhetorical text-to-image generation. Our model outperforms SOTA MLLMs such as GPT-4o, Grok-3, and leading academic baselines, across both qualitative and quantitative evaluations. The code and dataset used in this work are publicly available at: \url{https://anonymous.4open.science/r/Rhet2Pix-2D52/}.

\end{abstract}


\section{Introduction}

Rhetorical language infuses text with vivid sensory and imaginative depth, and visualizing linguistic metaphors amplifies multimodal understanding while deepening engagement with rhetorical expression \cite{chakrabarty2023spy}. Advanced text-to-image models offer a promising approach for this visualization task. Models such as GPT-4o \citep{openai2024gpt4ocard}, DALL-E 3 \citep{betker2023improving}, Stable Diffusion \citep{rombach2022highresolutionimagesynthesislatent}, and Imagen \citep{saharia2022photorealistictexttoimagediffusionmodels} are capable of generating highly realistic images from textual input. However, these models often prioritize general-purpose generation and may struggle with capturing the abstract and figurative nature of rhetorical expressions. Rhetorical language visualization, framed as a task-specific text-to-image generation problem, can benefit from reinforcement learning (RL), which is increasingly used to fine-tune diffusion models for specialized generative objectives \citep{black2024trainingdiffusionmodelsreinforcement,%
fan2023dpokreinforcementlearningfinetuning,%
hu2025towards,%
lee2023aligningtexttoimagemodelsusing,%
prabhudesai2024aligningtexttoimagediffusionmodels,%
wallace2023diffusionmodelalignmentusing,%
xu2023imagerewardlearningevaluatinghuman}. Recasting the denoising process in diffusion models as a Markov Decision Process (MDP) enables the application of RL and policy optimization to achieve task-specific generative objectives. \citep{kaelbling1996reinforcementlearningsurvey,%
rombach2022highresolutionimagesynthesislatent,%
wang2023patchdiffusionfasterdataefficient}.

Despite recent progress, generating faithful visualizations of rhetorical expressions remains a significant challenge. Existing text-to-image models typically encode input text as a guiding condition for image generation, treating individual words as discrete visual components. While this approach enables the image generation of recognizable objects and attributes, it results in fragmented visuals that fails to capture the compositional semantics and figurative structures essential for rhetorical meaning.
Even models with strong language capabilities, such as GPT-4o, are limited by this overly strict alignment between text and image, often producing unnatural fusions of subject and metaphorical vehicle that distort the intended interpretation. 

\begin{figure*}[t]
    \centering
    \includegraphics[width=1\linewidth]{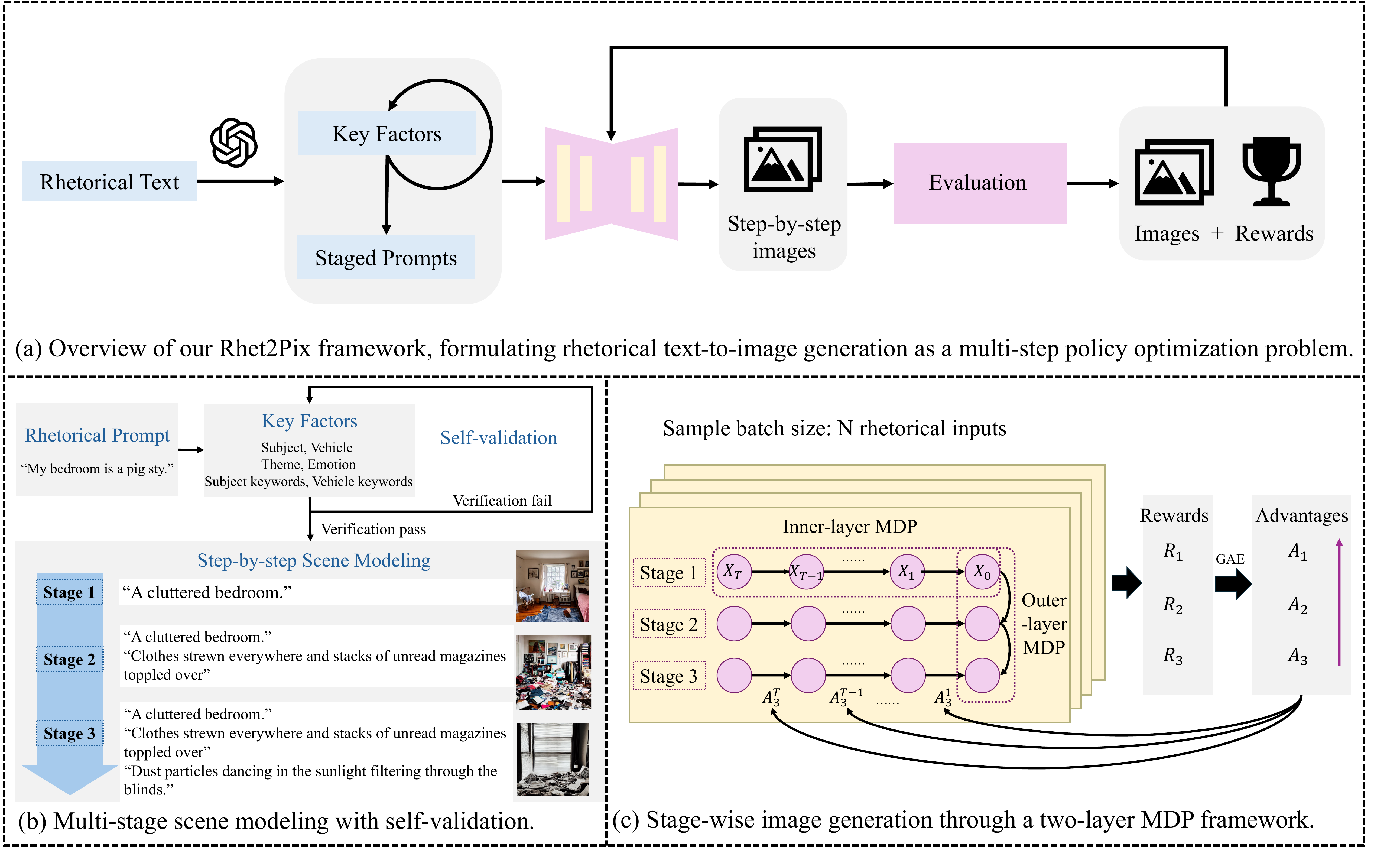}
    \caption{\textbf{Rhet2Pix framework.} 
    Our approach formulates rhetorical image generation as a two-layer Markov Decision Process.} %The outer layer controls the progressive construction of the image, transitioning from simple object-level representations to increasingly detailed scenes. The inner layer performs stepwise denoising to refine visual outputs conditioned on intermediate prompts.
    \label{fig:enter-label}
    \vspace{-10pt}
\end{figure*}

To build a high-quality text-to-image model for rhetorical language, we propose Rhet2Pix. 
% \textemdash 
Inspired by the incremental process by which a painter develops a scene,
Rhet2Pix models rhetorical image generation as a multi-step, two-layer Markov Decision Process (MDP), as illustrated in Figure \ref{fig:enter-label}. In the outer layer, the input rhetorical prompt is progressively decomposed into a sequence of semantically enriched sub-prompts, guiding step-wise image generation that incrementally refines the visual scene. In the inner layer,  Rhet2Pix models the denoising process of the diffusion model as a trajectory of actions and addresses the sparse reward problem by discounting the final reward along this trajectory, enabling fine-grained policy optimization over consecutive action pairs. To facilitate learning across both layers, we introduce a rhetorical-specific reward function that jointly evaluates semantic alignment and visual element consistency.

To summarize, we make the following key contributions:
\begin{itemize}
    \item 
    To the best of our knowledge, our work is the first to identify the misalignment issue in rhetorical text-to-image generation and trace its root cause: the word embedding of the vehicle object introduces a bias that leads the model to generate its literal visual counterpart. This creates a counter-effect — stronger object-level correspondence results in images that deviate further from the intended semantic meaning. This issue highlights a fundamental limitation in current MLLMs, which predominantly depend on surface-level text–image similarity.
    
    \item 
    Our Rhet2Pix framework formulates rhetorical image generation as a multi-step policy optimization problem based on a two-layer Markov Decision Process (MDP),  progressively generating images that faithfully reflect the intended rhetorical meaning. In the outer layer, it incrementally converts the input into semantically enriched prompts to guide progressive visual refinement, while in the inner layer, it optimizes adjacent denoising steps in the diffusion process using discounted rewards to address the challenge of reward sparsity.
    
    \item 
    In the rhetorical image generation task,
    Rhet2Pix outperforms state-of-the-art academic baselines and leading MLLMs such as GPT-4o and Grok-3, demonstrating superior performance in both qualitative and quantitative evaluations.

\end{itemize}

\section{Related work}

\noindent\textbf{Rhetoric research.}
Current research on rhetorical language mainly focuses on natural language processing tasks \citep{chakrabarty2021mermaid, veale2016round, chakrabarty2022flute}, particularly the detection and interpretation of metaphors \citep{gong2020illinimet, leong2018report, tsvetkov2014metaphor}. Studies that bridge rhetorical text and image generation are limited. \citet{akula2023metaclue} introduced the MetaCLUE framework for identifying metaphorical concepts in images, while \citet{chakrabarty2023spy} explored the integration of large language models with diffusion models for metaphor understanding. Despite these advances, a clear gap remains in generating images that accurately reflect the underlying semantics of rhetorical input. Our method addresses this challenge by generating images that align with the intended rhetorical meaning and accurately represent key semantic elements.

\noindent\textbf{Automated prompt construction.}
Automatically transforming user input into effective prompts is essential for building high-performing and user-friendly text-to-image generation systems \citep{xie2023prompt}. Traditional approaches—ranging from hand-crafted templates \citep{dai2021knowledge, petroni2019language}, to interactive refinement tools \citep{brade2023promptify, feng2023promptmagician, liu2022design}, and prompt optimization via reinforcement learning or genetic algorithms \citep{gao2023can, hao2023optimizing, mo2024dynamic}—rely heavily on expert intervention and iterative tuning, thereby limiting scalability and hindering full automation.
Recent advancements in large language models (LLMs) have enabled autonomous semantic decomposition \citep{yang2024masteringtexttoimagediffusionrecaptioning}, task planning \citep{dai2024optimalscenegraphplanning, rana2023sayplangroundinglargelanguage}, and structured scene layout \citep{prasad2024adaptasneededdecompositionplanning} in multimodal tasks, opening new possibilities for more efficient text-to-image prompt construction.
Building on these developments, our method leverages LLMs to automatically convert rhetorical inputs into a sequence of staged prompts with progressively enriched semantic content, providing structured guidance for stepwise image generation.

\noindent\textbf{Text-to-image generation.}
In recent years, diffusion models \citep{ho2020denoising, sohl2015deep} and autoregressive models \citep{tian2024visualautoregressivemodelingscalable, jiang2025t2ir1reinforcingimagegeneration, sun2024autoregressivemodelbeatsdiffusion} have gained significant attention for their outstanding performance in text-to-image generation. Compared to autoregressive models, diffusion models excel in generating high-quality and diverse images \citep{batzolis2021conditional, ho2022cascaded, ramesh2021zero, rombach2022high}. By refining random noise through multiple denoising steps, diffusion models produce more coherent images that better align with the input text \citep{black2024trainingdiffusionmodelsreinforcement, hu2025towards}. This method has quickly become the dominant framework for text-to-image generation \citep{betker2023improving, saharia2022photorealistictexttoimagediffusionmodels, zhang2023text, yang2025mmadamultimodallargediffusion}.

\noindent\textbf{Reinforcement Learning for Text-to-Image Generation.}
Despite advancements in diffusion models, aligning generated images with the complex semantics of input text remains challenging. Several studies have framed the denoising process of diffusion models as a reinforcement learning (RL) problem \citep{black2024trainingdiffusionmodelsreinforcement, fan2023dpokreinforcementlearningfinetuning, hu2025towards, lee2023aligningtexttoimagemodelsusing, prabhudesai2024aligningtexttoimagediffusionmodels, wallace2023diffusionmodelalignmentusing, xu2023imagerewardlearningevaluatinghuman}, enabling fine-tuning to better align outputs with human preferences \citep{lee2023aligning, zhang2024hive} and semantic intent \citep{dong2023raft, huang2023t2i}. A key issue with RL-based diffusion models is reward sparsity, where feedback is only given after the full image is generated, leading to high gradient variance and slow convergence \citep{arriola2025block}. Recent solutions, including progressive backward training and branch-based sampling, have aimed to address this \citep{hu2025towards}. Our approach further reduces reward sparsity by discounting the final reward across the diffusion trajectory and optimizing each adjacent action pair.

\section{Preliminaries}
\noindent\textbf{Text-to-image diffusion models. }
The diffusion model process consists of two stages: the forward process and the reverse process. In the forward process, the image $x_0$ is gradually corrupted into pure noise $x_T$ over $T$ steps, with Gaussian noise being added at each step. The reverse process generates an image from pure noise by iteratively denoising, conditioned on a textual description $\mathbf{c}$. It is parameterized by a neural network $\boldsymbol{\mu}_\theta\left(\mathbf{x}_t, \mathbf{c}, t\right)$, predicting the noise $\epsilon$ that transforms $x_0$ to $x_T$. Sampling starts from $x_T \sim N(0, I)$ and generates denoised samples iteratively \citep{ho2020denoising,%
song2020denoising}.
\[
    p_\theta\left(\mathbf{x}_{t-1} \mid \mathbf{x}_t, \mathbf{c}\right)=\mathcal{N}\left(\mathbf{x}_{t-1} \mid \boldsymbol{\mu}_\theta\left(\mathbf{x}_t, \mathbf{c}, t\right), \sigma_t^2 \mathbf{I}\right)
\]

where $\mu_\theta$ is predicted by a diffusion model parameterized by $\theta$, and $\sigma_t$ is the fixed timestep-dependent variance.

\noindent\textbf{Markov decision process and reinforcement learning. }
The denoising process in diffusion models can be viewed as a Markov Decision Process (MDP), formulated as a sequential decision-making problem. We define an MDP $\left(\mathcal{S}, \mathcal{A}, P_0, P, R\right)$ with states $s \in \mathcal{S}$, actions $a \in \mathcal{A}$, initial state distribution $P_0$, transition probabilities $P$, and reward $R$. At each timestep $t$, the agent (e.g., diffusion model) observes state $s_t \in \mathcal{S}$, selects action $a_t\sim\pi_\theta\left(a_t \mid s_t\right) \in \mathcal{A}$, transitions to the next state $s_{t+1}\sim P\left(s_{t+1} \mid s_t, a_t\right)$, and receives reward $R\left(s_t, a_t\right)$. As the agent interacts with the MDP, it generates a trajectory: a sequence of states and actions $\tau=\left(s_0, a_0, s_1, a_1, \ldots, s_T, a_T\right)$. The goal of Reinforcement Learning (RL) is to maximize expected cumulative reward from trajectories sampled according to its policy.

\[
    \mathcal{J}_{\mathrm{RL}}(\pi_\theta)=\mathbb{E}_{\tau \sim p(\tau \mid \pi_\theta)}\left[\sum_{t=0}^T R\left(s_t, a_t\right)\right]
\]

where the policy $\pi_\theta$ parameterized by $\theta$ defines the action selection strategy and $\theta$ is updated by gradient descent.

% \section{Method}

\section{Rhet2Pix framework}
In this section, we introduce the Rhet2Pix framework, designed to automatically and efficiently generate high-quality images from rhetorical text. It captures the underlying semantics and expressive intent of the input without requiring human intervention.

\subsection{Problem statement}
Given a dataset $\left\{T_1, T_2, \ldots, T_n\right\}$ of rhetorical texts, the objective is to generate a corresponding set of images $\left\{I_1, I_2, \ldots, I_n\right\}$ that accurately reflect the semantic meanings of each $T_i$ without manual intervention.
This task presents two key challenges:
\begin{itemize}
    \item \textbf{Semantic-Level Alignment.} 
    % Current text-to-image pipelines can only handle simple, brief prompts and lack the capacity to represent complex or extended textual semantics.
    Models often collapse distinct ideas or oversimplify complex rhetorical expressions into literal interpretations, resulting in visuals that overlook deeper contextual meanings. 
    \item \textbf{Object and Image-Level Alignment.} 
    The alignment of semantic meaning sometimes causes the destruction of the original images. Models must balance meaning preservation with aesthetic integrity.
    \end{itemize}

\subsection{Multi-stage Scene Modeling}
Given each rhetorical text $T_i$, we decompose it into a compact set of key factors $F_i$ and derive a sequence of detailed scene instruction $\big\{P_i^j\big\}_{j=1}^C$, where $C$ denotes the total number of stages (i.e., the number of prompts).

In the first stage, we leverage a 
% fine-tuned 
powerful
large language model (GPT-4o) to extract seven semantic dimensions, including the rhetorical device, literal subject, metaphorical vehicle, overarching theme, emotional tone, subject keywords, and vehicle keywords. The first five are used for the subsequent scene description generation, while the last two, containing elements related to the subject and vehicle, are employed for object detection in the reward evaluation.
$$
F_i=\left[d_i^{\text {device }}, d_i^{\text {sub }}, d_i^{\text {veh }}, d_i^{\text {theme }}, d_i^{\text {emotion }}, d_i^{\text {sub list}}, d_i^{\text {veh list}}\right] 
$$

To guarantee the accuracy of factor extraction, we employ a generate-verify-retry loop:
for each candidate set of key factors $F$,
% for each candidate $F_i^{(k)}$, where $F_i^{(k)}$ denotes the candidate produced in the $k$-th iteration of the loop, 
the LLM is asked to compute a semantic coherence score and a rhetorical consistency score, and we accept the first candidate $F$ satisfying the following criteria as the validated factor set $F_i$:
\[
\operatorname{verify}(F, T):=\mathbb{I}\left[s_{\operatorname{coh}}(F, T) \geq \tau_c \wedge s_{\text {rhet }}(F, T) \geq \tau_r\right],
\]
% where $\tau_c$ and $\tau_r$ represent two small constants as thresholds.
where $\tau_c$ and $\tau_r$
denote predefined threshold values for coherence and rhetorical alignment, respectively.

In the second stage,
inspired by the incremental process by which a painter develops a scene, we utilize reinforcement learning and construct an agentic framework to model the text-to-image generation process that simulates the multi-steps task in control\citep{ren2024diffusion}. Specifically, we define a staged prompting sequence $\left\{P_i^1, P_i^2, \ldots, P_i^C\right\}$, where each prompt $P_i^j$ introduces additional semantic elements to the scene. The sequence begins with a core subject description and is followed by progressively enriched content such as environmental context, lighting conditions, spatial layout, emotional tone, and other such attributes.

During training, the prompts $\big\{P_i^j \big\}_{j=1}^C$ are fed into the diffusion-based generator to yield images $\big\{I_i^j \big\}_{j=1}^C$, while $F_i$ supplies the RL reward guiding prompt optimization and boosting semantic fidelity. At inference, only the final-stage prompts ${P_i^C}$ will be used for image generation.

\subsection{Stage-wise Image Generation}
We formulate rhetorical image generation as a two-layer Markov Decision Process (MDP), where the inner layer models the denoising trajectory of the stable diffusion model and the outer layer captures the semantic progression guided by staged prompt refinement.

In the inner-layer MDP, the denoising process begins from pure Gaussian noise $\mathbf{x}_T \sim \mathcal{N}(0, I)$ and proceeds through $T$ iterative steps to produce a final image $\mathbf{x}_0$. At each denoising step $t$, the components of the MDP are defined as follows:
\[
\begin{aligned}
\text { State: } & \mathbf{s}_t:=\left(\mathbf{c}, t, \mathbf{x}_t\right) \\
\text { Action: } & \mathbf{a}_t:=\mathbf{x}_{t-1} \\
\text { Policy: } & \pi_\theta\left(\mathbf{a}_t \mid \mathbf{s}_t\right):=p_\theta\left(\mathbf{x}_{t-1} \mid \mathbf{x}_t, \mathbf{c}\right)
\end{aligned}
\]

where $\mathbf{c}$ is the conditioning prompt, $t$ is the diffusion timestep, and $\mathbf{x}_t$ is the intermediate latent. This process yields a trajectory $\tau=\left\{\left(\mathbf{s}_t, \mathbf{a}_t\right)\right\}_{t=T}^0$, over which we assign credit and optimize the policy.

In the outer-layer MDP, a rhetorical input $T_i$ is decomposed into a sequence of staged prompts $\big\{P_i^j \big\}_{j=1}^C$, progressively enriching the semantic content. At outer timestep $j$, the state is defined as $\mathbf{s}_j^{\text {outer }}:=I_i^j$, the image generated from $P_i^j$, and the action $\mathbf{a}_j^{\text {outer }}$ represents the semantic substances added to $P_i^{j}$ to obtain $P_i^{j+1}$. Each generated image $I_i^j$, i.e., outer-layer state $\mathbf{s}_j^{\text {outer }}$, will be evaluated by the reward function.

\subsection{Optimization over Adjacent Image Pairs}

\begin{figure}[t]
    \centering
    \includegraphics[width=1\linewidth]{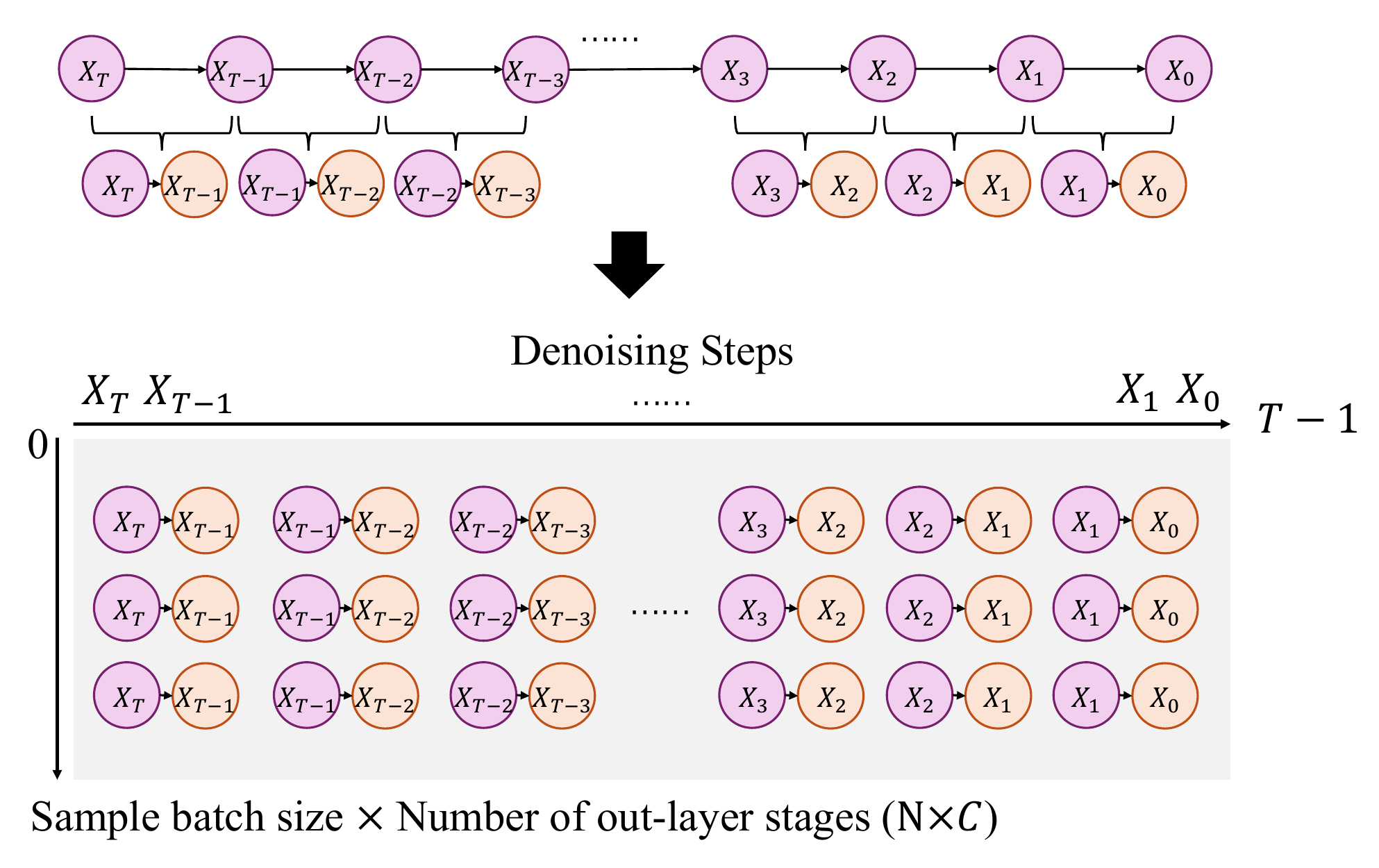}
    \caption{\textbf{Sampling and optimization over adjacent action pairs.} We collect denoising trajectories across $N$ rhetorical inputs, each with $C$ staged prompts and $T$ diffusion steps, yielding $N \times C \times T$ adjacent action pairs. These pairs are shuffled across input, stage, and timestep dimensions to remove correlation. Sampled pairs are used to compute action likelihoods under current and past policies, and the policy is updated via PPO using reweighted semantic advantages.}
    \label{fig:pool}
\end{figure}

To optimize the denoising policy $\pi_\theta$, we leverage the two-layer MDP structure by first computing advantages of generated images at the outer layer, and then propagating these advantages to the inner-layer action space via trajectory-level discounting.

Given $N$ rhetorical inputs, each decomposed into $C$ staged prompts, we obtain $N \times C$ images and their corresponding outer-layer states $\big\{\mathbf{s}_j^{\text {outer }}\big\}_{j=1}^C$. For each outer timestep $j$, a reward $r(I_i^j)$ is computed, and advantages $\hat{A}_i$ are estimated using Generalized Advantage Estimation (GAE):
\begin{align}
\delta_j &:= r_j + \gamma V^{\pi_{\theta_{\text{old}}}}\left(I_{j+1}\right)
               - V^{\pi_{\theta_{\text{old}}}}\left(I_j\right), \notag \\
\hat{A}_j &:= \sum_{l=0}^{C-j} (\gamma \lambda)^l \delta_{j+l} .
\end{align}

These outer-layer advantages are then discounted backward across their corresponding denoising trajectories. For each inner timestep $t$ 
% in trajectory $i$
, the step-wise advantage is:
\[
\hat{A}_t^{(j)}:=\gamma_{\text {denoise }}^t \cdot \hat{A}_j, \quad t=T, T-1, \ldots, 1 .
\]

As shown in Figure \ref{fig:pool}, we collect all denoising trajectories across the dataset, yielding $N \times C \times T$ adjacent action pairs ($\mathbf{s}_t, \mathbf{s}_{t+1}$). Each pair corresponds to a denoising action, where the latent is updated from $\mathbf{x}_t$ to $\mathbf{x}_{t-1}$. All action pairs are shuffled across rhetorical input, prompt stage, and timestep dimensions to remove temporal correlation, and mini-batches are sampled for policy updates.

For each sampled action pair ($\mathbf{s}_t, \mathbf{s}_{t+1}$), a single DDIM denoising step is applied to $\mathbf{s}_t$, producing the predicted mean $\mu_t$ and variance $\sigma_t^2$. A Gaussian $\mathcal{N}\left(\mu_t, \sigma_t^2\right)$ defines the distribution over possible actions. The likelihood of generating $\mathbf{a}_t=\mathbf{x}_{t-1}$ is computed under both current and old policies:
\begin{align}
\pi_\theta\left(\mathbf{a}_t \mid \mathbf{s}_t\right) 
  &= \mathcal{N}\left(\mathbf{x}_{t-1} \mid \mu_t, \sigma_t^2\right), \notag \\
\pi_{\theta_{\text{old}}}\left(\mathbf{a}_t \mid \mathbf{s}_t\right) 
  &= \mathcal{N}\left(\mathbf{x}_{t-1} \mid \mu_t^{\text{old}}, (\sigma_t^2)^{\text{old}}\right).
\end{align}

The importance weight is then computed as:
\[
w_t=\frac{\pi_\theta\left(\mathbf{a}_t \mid \mathbf{s}_t\right)}{\pi_{\theta_{\text {old }}}\left(\mathbf{a}_t \mid \mathbf{s}_t\right)}.
\]

The PPO update is applied over the sampled action pairs, using the clipped surrogate loss
\begin{align}
\nabla_\theta \mathcal{L}_{\mathrm{PPO}}(\theta) 
&= \mathbb{E}_{(\mathbf{s}_t, \mathbf{s}_{t+1}) \sim \mathcal{D}} \Big[
      \nabla_\theta \min \big( \notag w_t \cdot \hat{A}_t^{(j)},\\\
&\operatorname{clip}(w_t, 1{-}\epsilon, 1{+}\epsilon) \cdot \hat{A}_t^{(j)}
\big) \Big] .
\end{align}

where $\epsilon$ is the PPO clipping parameter, and $\mathcal{D}$ is the distribution of sampled action pairs. This formulation enables efficient reinforcement learning over diffusion-based generation, guided by rhetorical structure and semantic alignment.

\subsection{Tailored Reward Mechanism}
We design a tailored reward mechanism for rhetorical text-to-image generation, consisting of three modules: staged semantic alignment, elemental alignment, and aesthetic quality. Together, these modules guide the model to generate high-quality images that capture both the coarse-to-fine structure of the rhetorical prompt and its true metaphorical intent.

\noindent\textbf{Staged semantic alignment.}
To overcome the limitations of using a single global similarity metric-which treats all parts of the prompt equally and tends to overweight incidental visual attributes such as background or lighting while neglecting the hierarchical importance of semantic components-we propose a staged alignment strategy. Specifically, each scene prompt $P_i^j$ is decomposed into an ordered sequence of $j$ subsentences $\big\{S_i^k\big\}_{k=1}^j$, designed to reflect the prompt's progressive semantic elaboration from core concepts to contextual details.

We compute the unit-normalized image embedding $\mathbf{v}_i^j$, $j=1,2,\cdots ,C$, for each training image $I_i^j$, and the unit-normalized text embeddings $\mathbf{u}_i^k$ for each sub-sentence.
We then introduce a monotonic weight vector $\mathbf{w}^{(j)} \in \mathbb{R}^j$ satisfying $w_1>w_2>\cdots>w_j$ and $\sum_{k=1}^j w_k=1$, to reflect the decreasing semantic centrality of later sub-sentences. The staged alignment reward is computed as
\[
r^{\text {stage }}\left(I_i^j, P_j^j\right)=\sum_{k=1}^j w_k\left\langle\mathbf{v}_i^j, \mathbf{u}_i^k\right\rangle,
\]

where $\langle\cdot, \cdot\rangle$ denotes the cosine similarity between the image and text embeddings.
By assigning greater weight to earlier (core) components of the prompt, this reward formulation encourages the model to prioritize essential semantic elements in the generation process, while still attending to finer contextual details in later stages.

\noindent\textbf{Elemental alignment.}
To correct the overemphasis on metaphorical vehicles that leads to semantic misalignment in generated images, we propose an elemental alignment mechanism that separately evaluates the vehicle and the true subject.  
Specifically, for each text input $T_i$, we construct two keyword sets: $d_i^{\text {sub list }}$, which includes a collection of subject-related elements such as core entities, associated states, and descriptive attributes; and $d_i^{\text {veh list }}$ which contains the metaphorical vehicle. Given a generated image $I_i^j$ with its unit-normalized embedding $\mathbf{v}_i^j$, we compute the subject reward as
\[
r^{\mathrm{subject}}\left(I_i^j\right)=\sum_{k \in d_i^{\text {sub list }}}\left\langle\mathbf{v}_i^j, \frac{f_{\text {text }}(k)}{\left\|f_{\text {text }}(k)\right\|}\right\rangle,
\]

which reflects the cumulative semantic alignment between the image and all subject-relevant components. Each term in the summation corresponds to the CLIP-based similarity between the image and a subject-related element, directly reflecting its semantic presence in the visual output.

To penalize inappropriate emphasis on the metaphorical vehicle, we define the vehicle penalty as
\[
r^{\mathrm{vehicle}}\left(I_i\right)= \begin{cases}-1.0, & \max _{k \in d_i^{\text {veh list }}}\left\langle\mathbf{v}_i^j, \frac{f_{\text {text }}(k)}{\left\|f_{\text {text }}(k)\right\|}\right\rangle>\tau \\ 0.0, & \text { otherwise }\end{cases},
\]

where $\tau$ is a fixed similarity threshold, manually determined by empirical analysis of CLIP similarity scores to approximate the point at which a concept can be considered visually present in the image.

\noindent\textbf{Aesthetic quality.} 
To avoid generating visually unappealing images due to excessive emphasis on semantic and elemental alignment, we introduce an aesthetic quality score as a supplementary measure. This score is based on the LAION aesthetics predictor \citep{schuhmann2022laion5bopenlargescaledataset}, which is trained on 176,000 human-rated images to evaluate the aesthetic quality of the generated images.

\begin{figure}[htbp]
  \centering
  \begin{subfigure}[b]{0.3\linewidth}
    \centering
    \includegraphics[width=\linewidth]{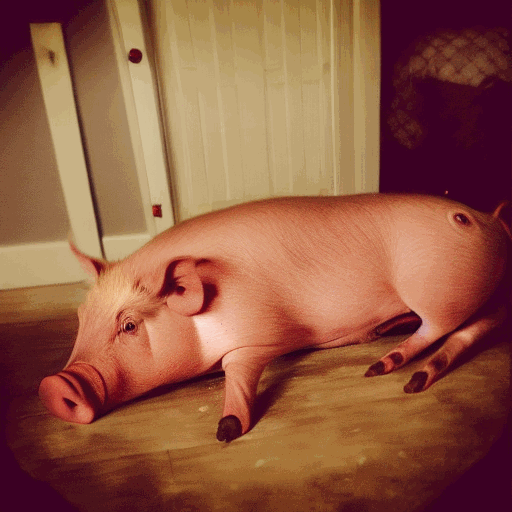}
    \caption{Bad}
    \label{fig:sub4}
  \end{subfigure}
  \hspace{0.03\linewidth}
  \begin{subfigure}[b]{0.3\linewidth}
    \centering
    \includegraphics[width=\linewidth]{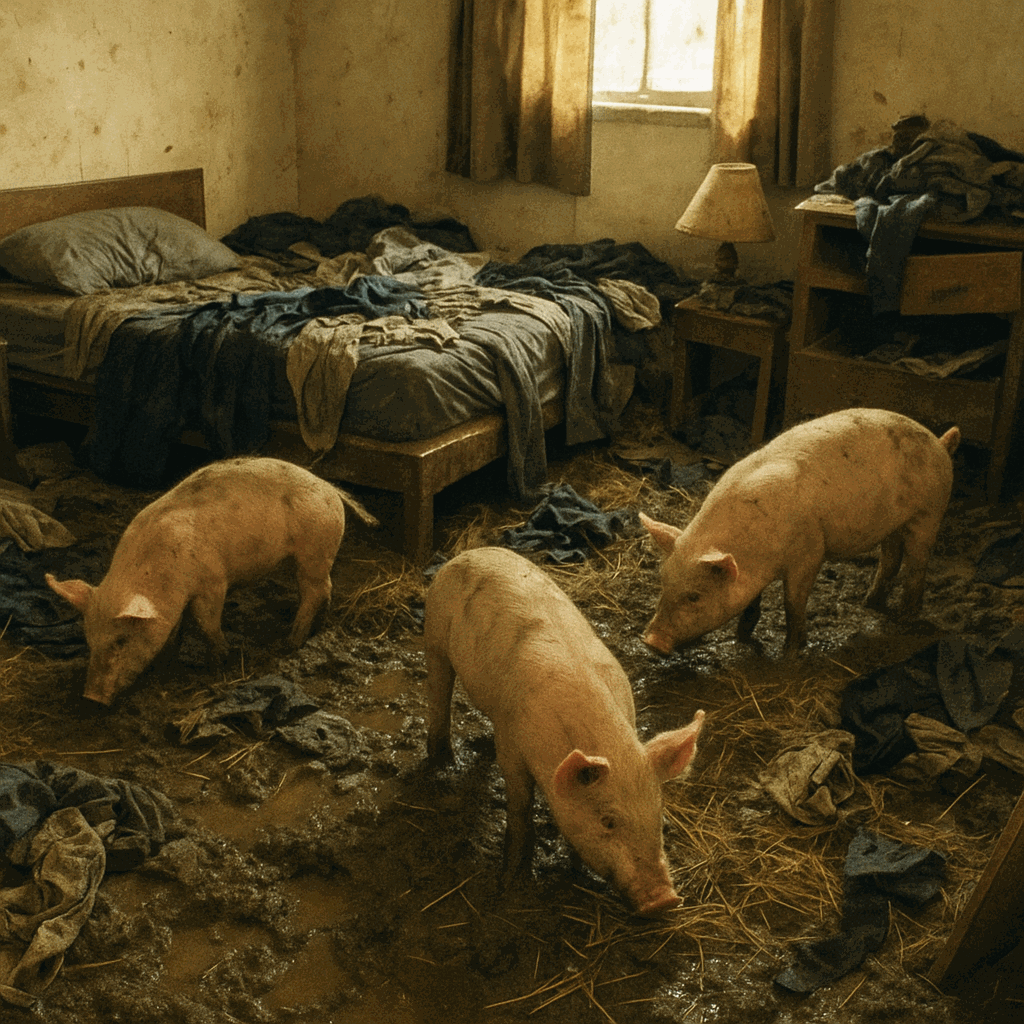}
    \caption{Medium}
    \label{fig:sub2}
  \end{subfigure}
  \hspace{0.03\linewidth}
  \begin{subfigure}[b]{0.3\linewidth}
    \centering
    \includegraphics[width=\linewidth]{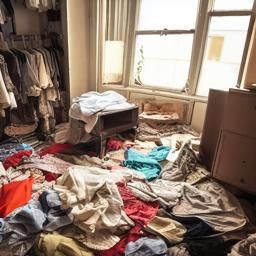}
    \caption{Good}
    \label{fig:sub1}
  \end{subfigure}
  
  \caption{\textbf{Effectiveness of our reward mechanism in capturing rhetorical semantics.} From left to right, the reward increases with images that better align with the underlying rhetorical meaning, receiving higher scores. All examples are generated from the prompt ``My bedroom is a pig sty.''}
  \label{fig:reward}
  \vspace{-10pt}
\end{figure}

To assess the effectiveness of our reward mechanism, we present examples as shown in Figure \ref{fig:reward}. Images with richer details that accurately reflect the intended semantic nuances are rewarded higher, while literal depictions of the metaphorical vehicle, such as pigs or pigsties, are penalized. These results demonstrate that our reward formulation effectively assesses both the semantic alignment and the appropriateness of visual elements in the generated images.

\section{Experiments}
\subsection{Experiment settings}

\begin{figure*}[tb]
  \centering
  \begin{tikzpicture}
    \matrix (m) [
      matrix of nodes,
      nodes in empty cells,
      column sep=1pt,
      row sep=1pt,
      nodes={anchor=center,
             inner sep=1pt,
             outer sep=1pt},
      row 1/.style={
        nodes={
          font=\small
        }
      }
    ]
    {
      %—— 第一行：模型标题 ——
        & \textbf{SD} & \textbf{DDPO} & \textbf{MMaDA} & \textbf{Imagen}
        & \textbf{GPT-4o} & \textbf{Grok-3} & \textbf{Rhet2Pix} \\
      %—— 第二行：a ——
      \textbf{(a)} 
        & \includegraphics[width=0.13\textwidth]{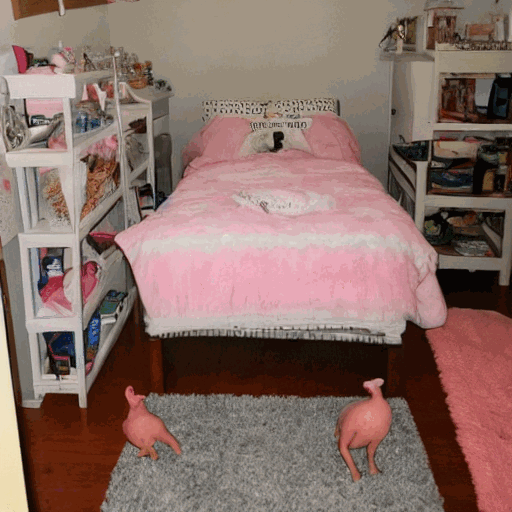}
        & \includegraphics[width=0.13\textwidth]{ddpo/bedroom.png}
        & \includegraphics[width=0.13\textwidth]{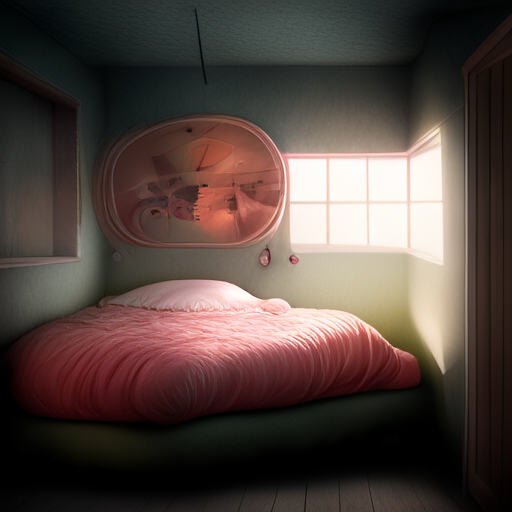}
        & \includegraphics[width=0.13\textwidth]{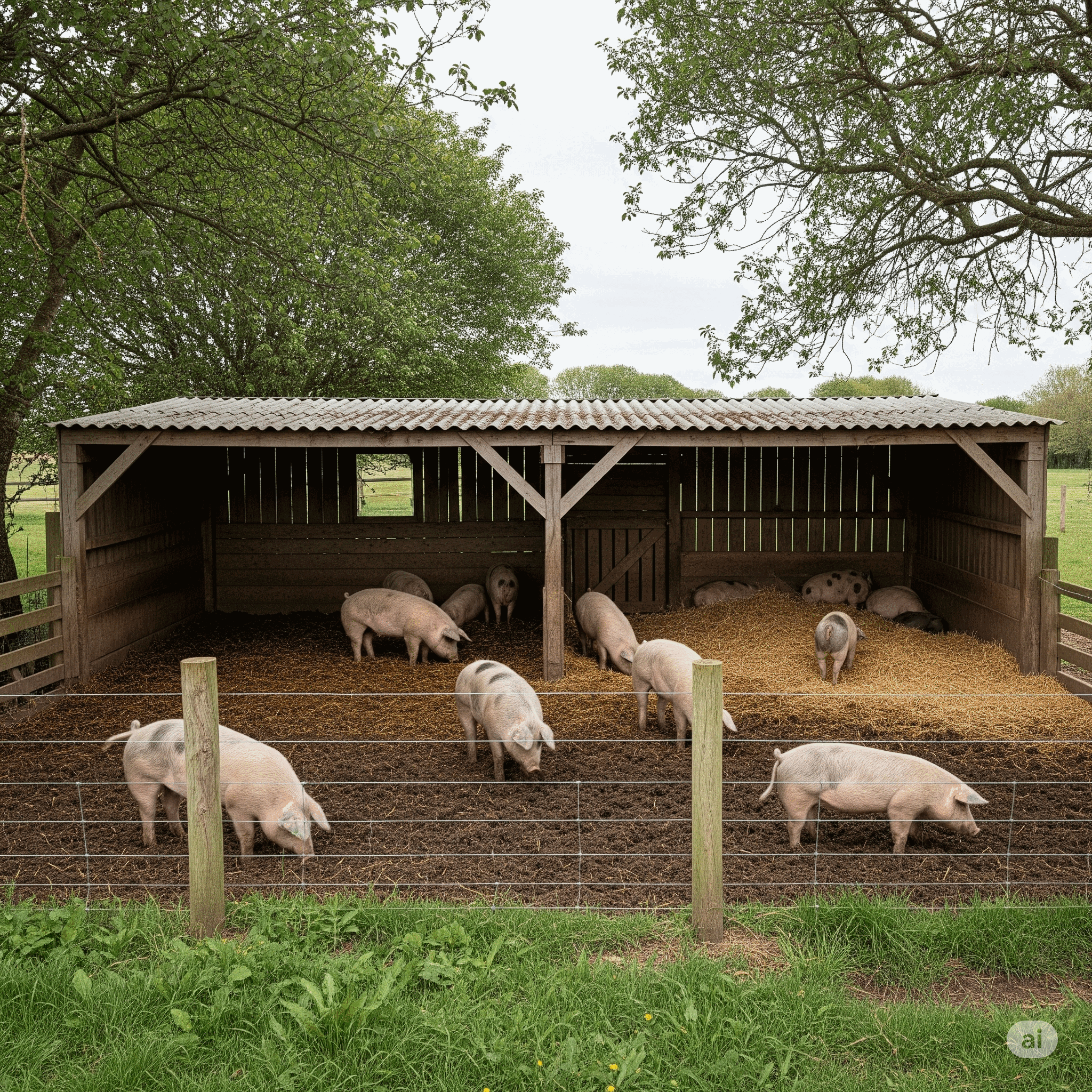}
        & \includegraphics[width=0.13\textwidth]{gpt-4o/bedroom.png}
        & \includegraphics[width=0.13\textwidth,height=0.13\textwidth, clip,trim=15pt 15pt 0pt 0pt]{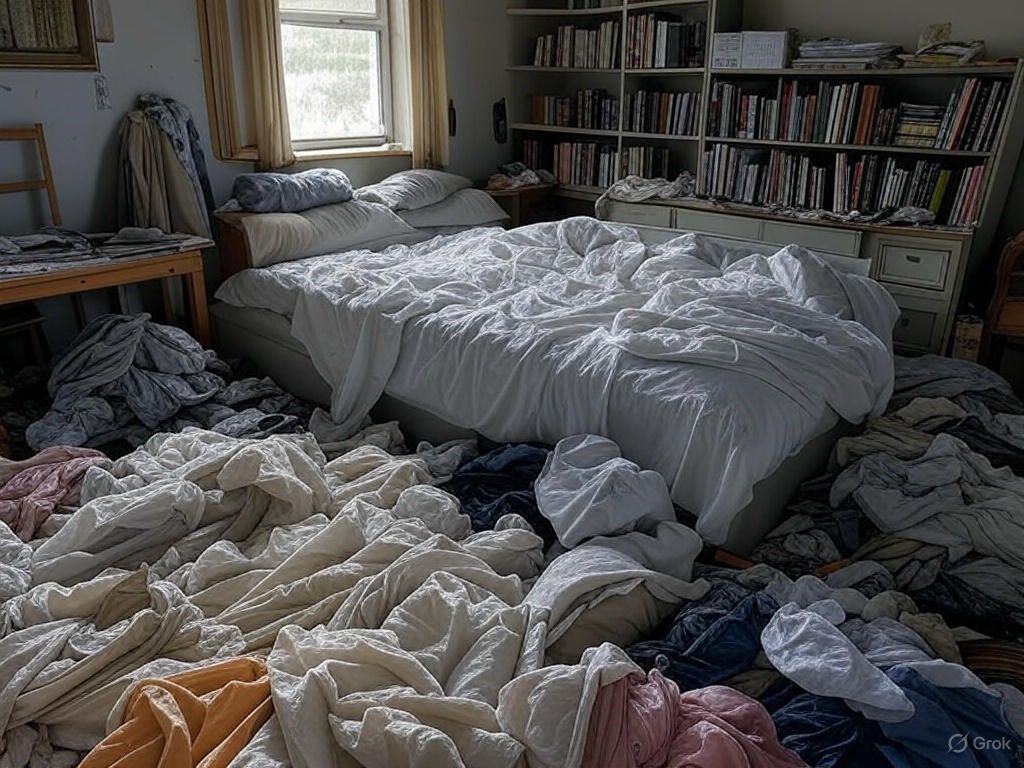}
        & \includegraphics[width=0.13\textwidth]{Rhet2Pix/bedroom.jpg} \\
      %—— 第三行：b ——
      \textbf{(b)} 
        & \includegraphics[width=0.13\textwidth]{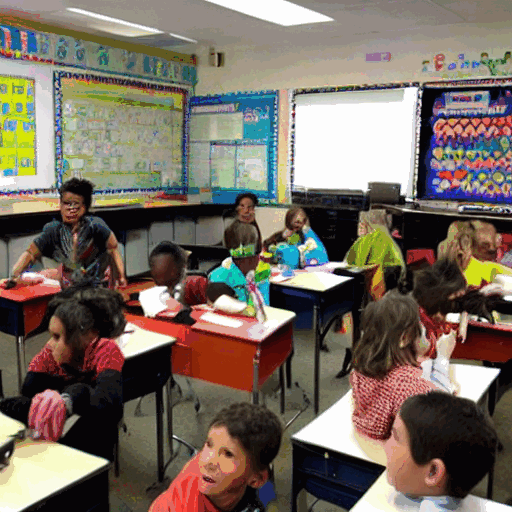}
        & \includegraphics[width=0.13\textwidth]{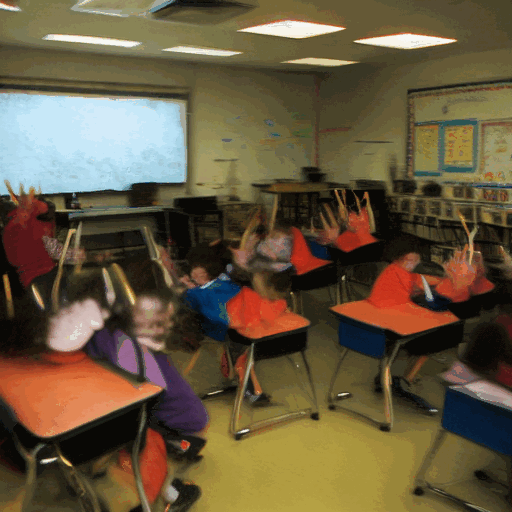}
        & \includegraphics[width=0.13\textwidth]{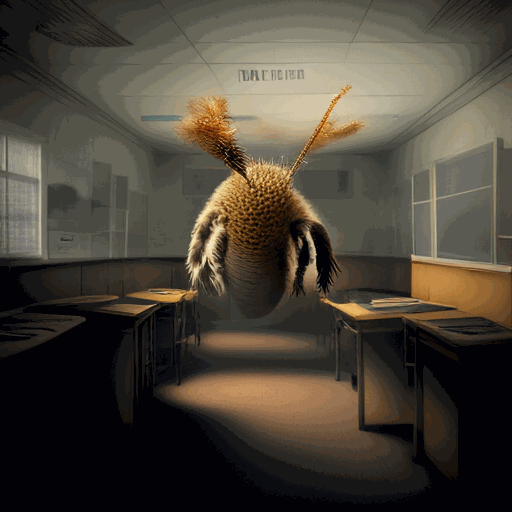}
        & \includegraphics[width=0.13\textwidth]{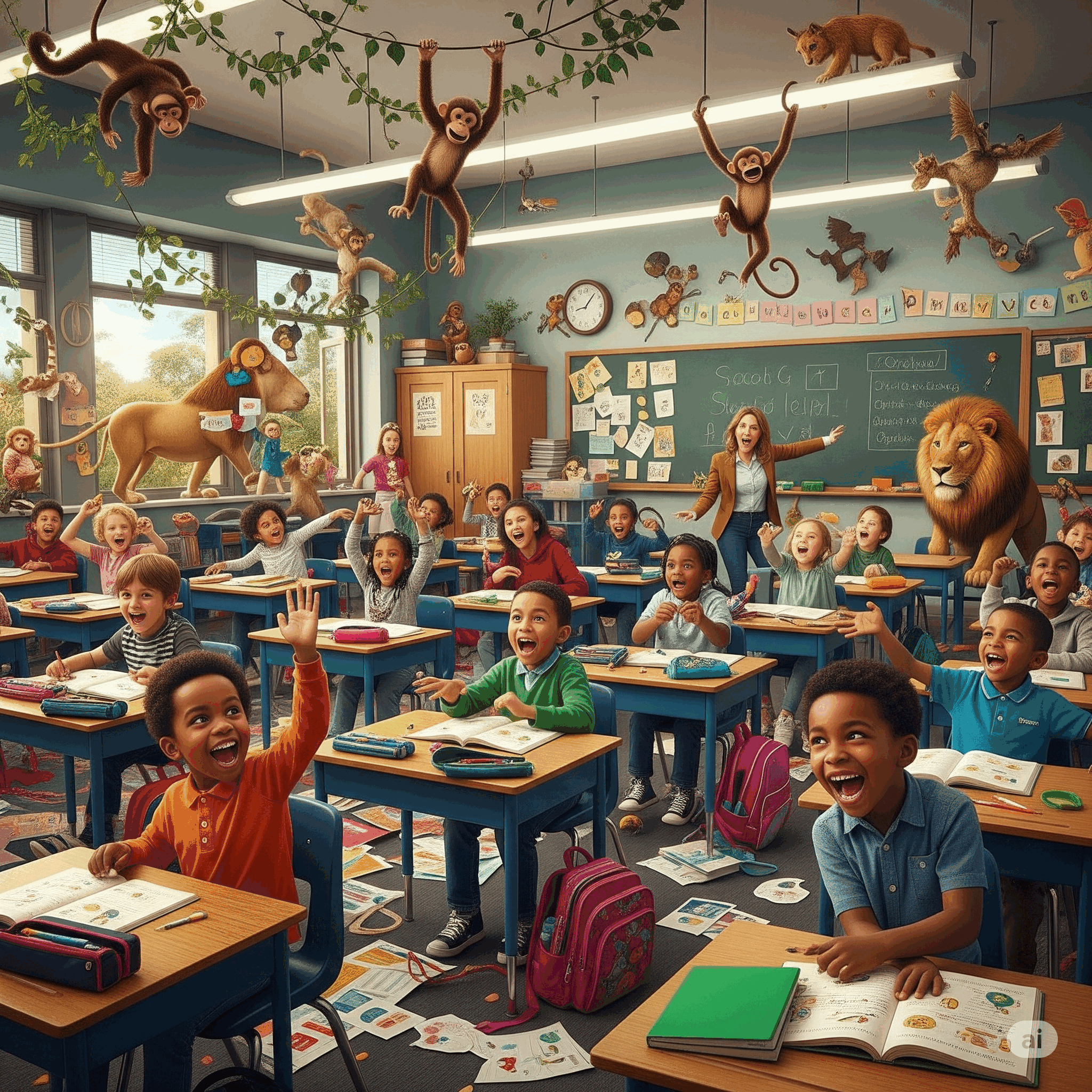}
        & \includegraphics[width=0.13\textwidth]{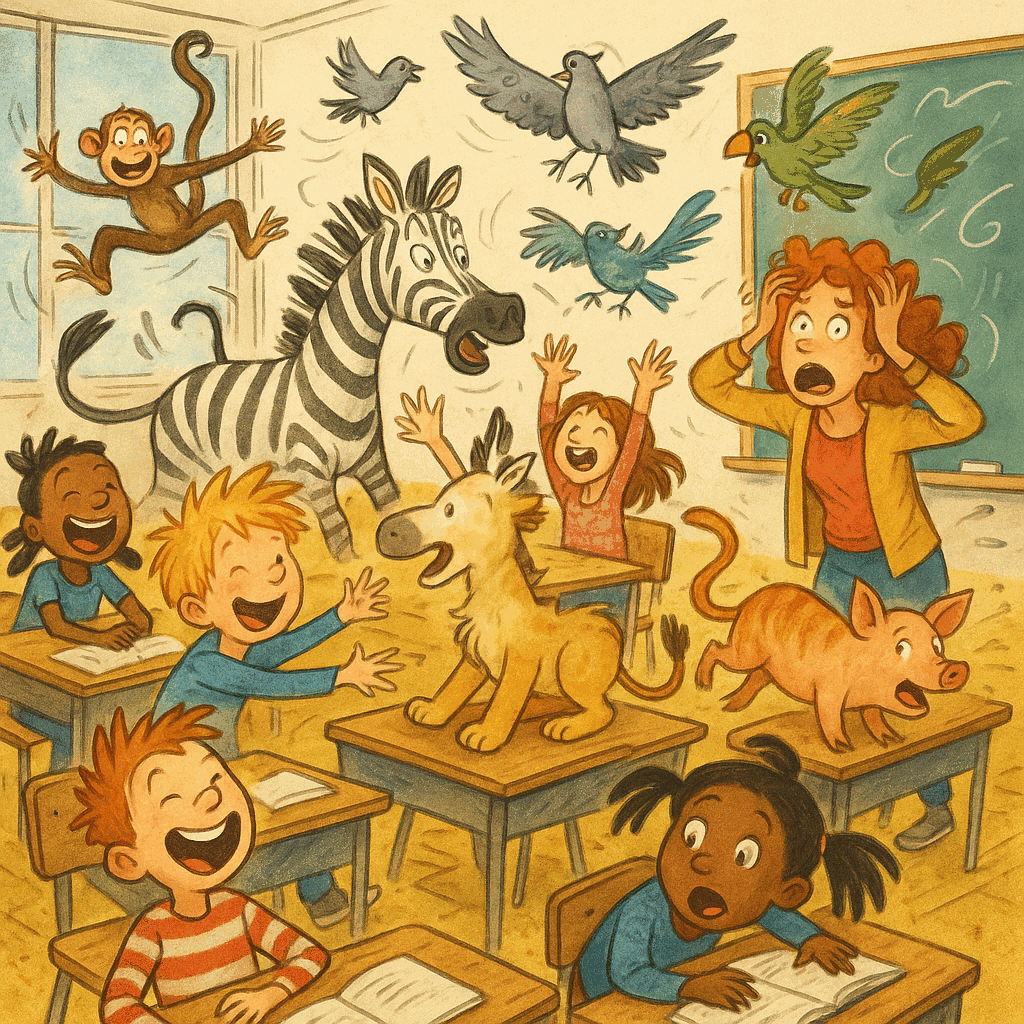}
        & \includegraphics[width=0.13\textwidth,height=0.13\textwidth, clip,trim=15pt 15pt 0pt 0pt]{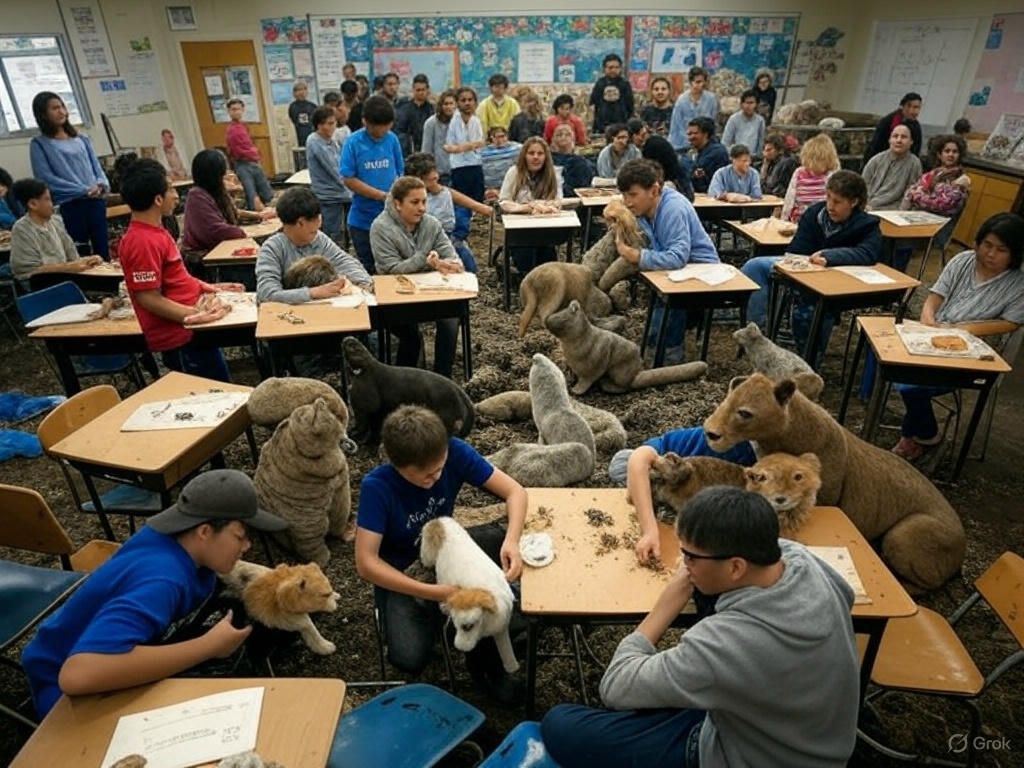}
        & \includegraphics[width=0.13\textwidth]{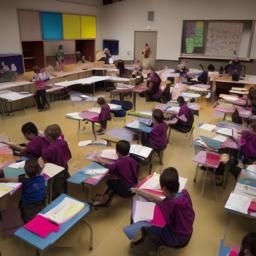} \\
      %—— 第四行：c ——
      \textbf{(c)} 
        & \includegraphics[width=0.13\textwidth]{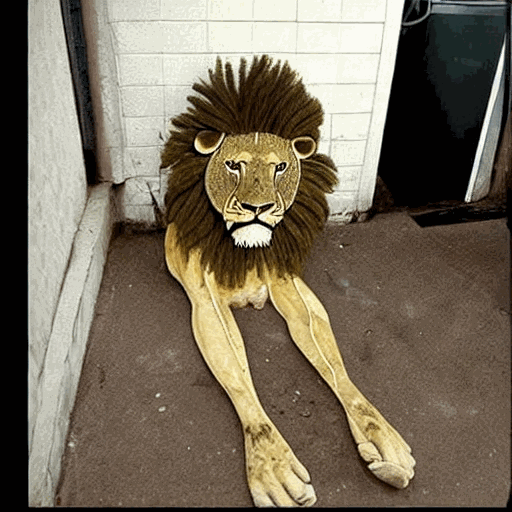}
        & \includegraphics[width=0.13\textwidth]{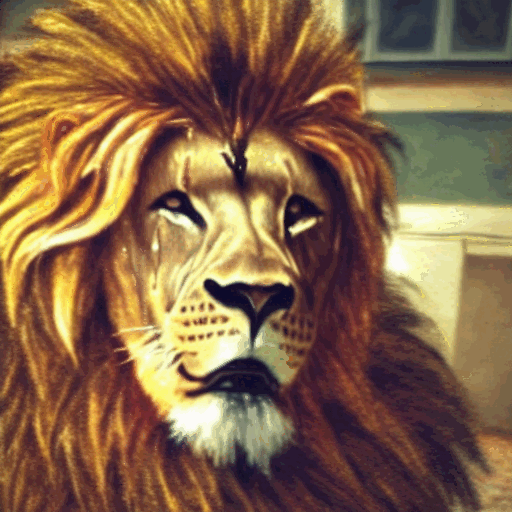}
        & \includegraphics[width=0.13\textwidth]{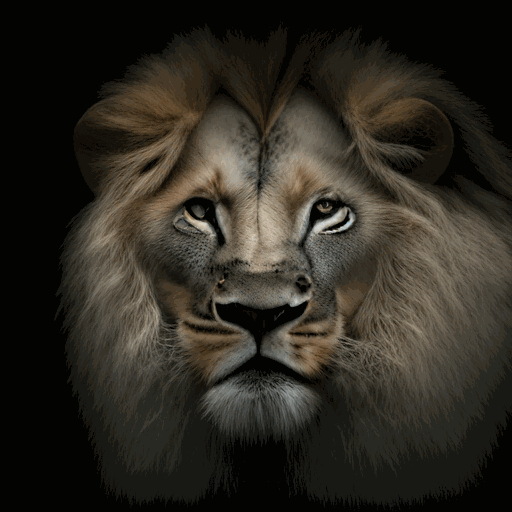}
        & \includegraphics[width=0.13\textwidth]{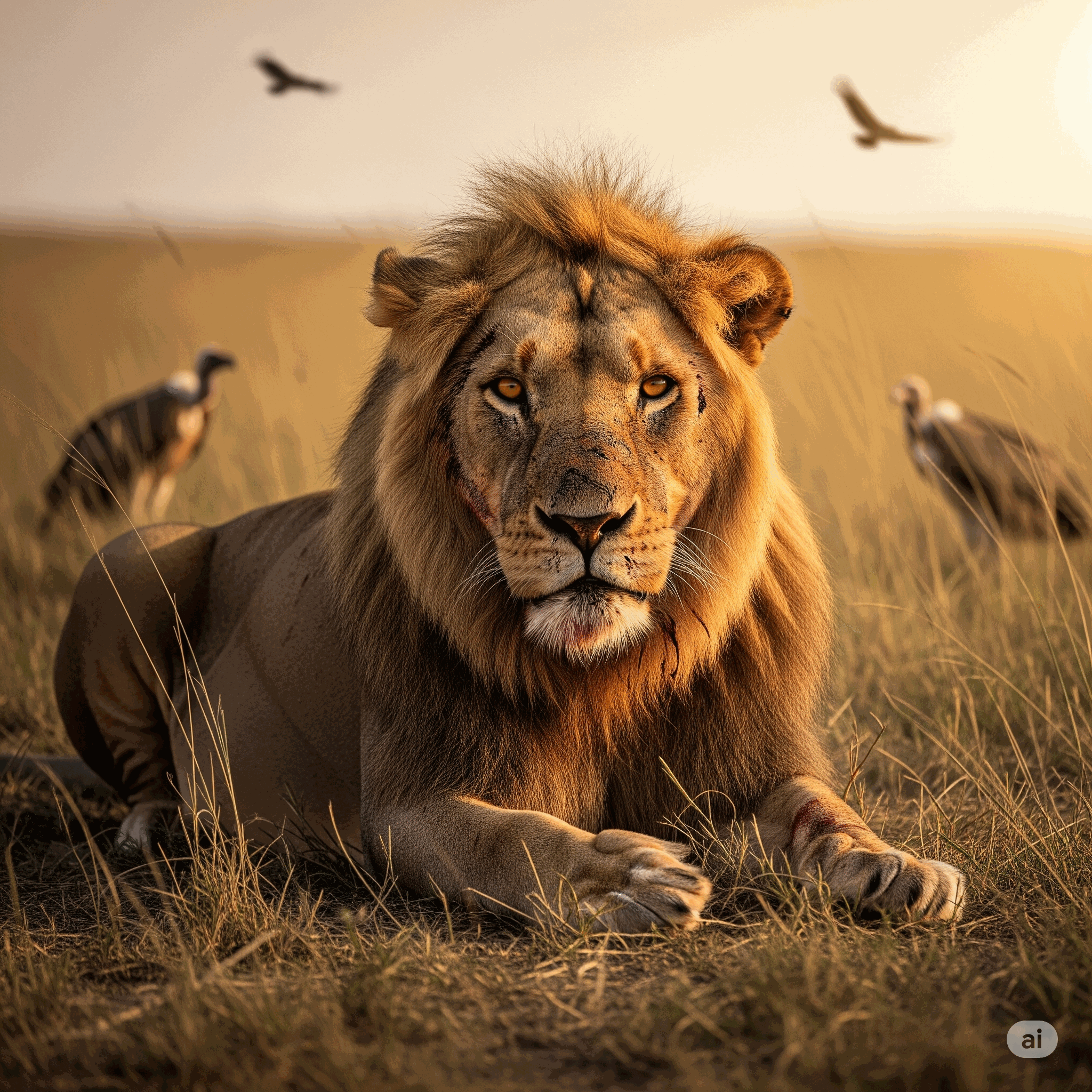}
        & \includegraphics[width=0.13\textwidth]{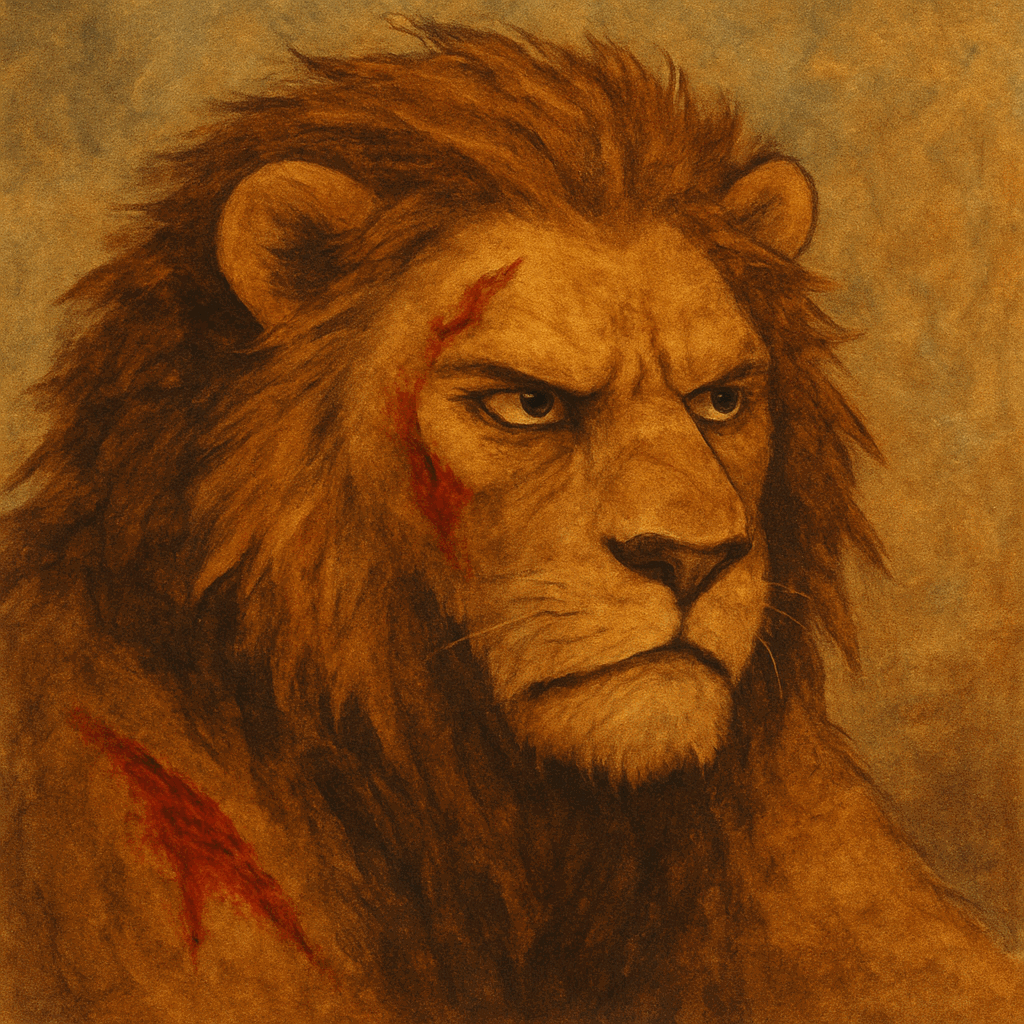}
        & \includegraphics[width=0.13\textwidth,height=0.13\textwidth, clip,trim=15pt 15pt 0pt 0pt]{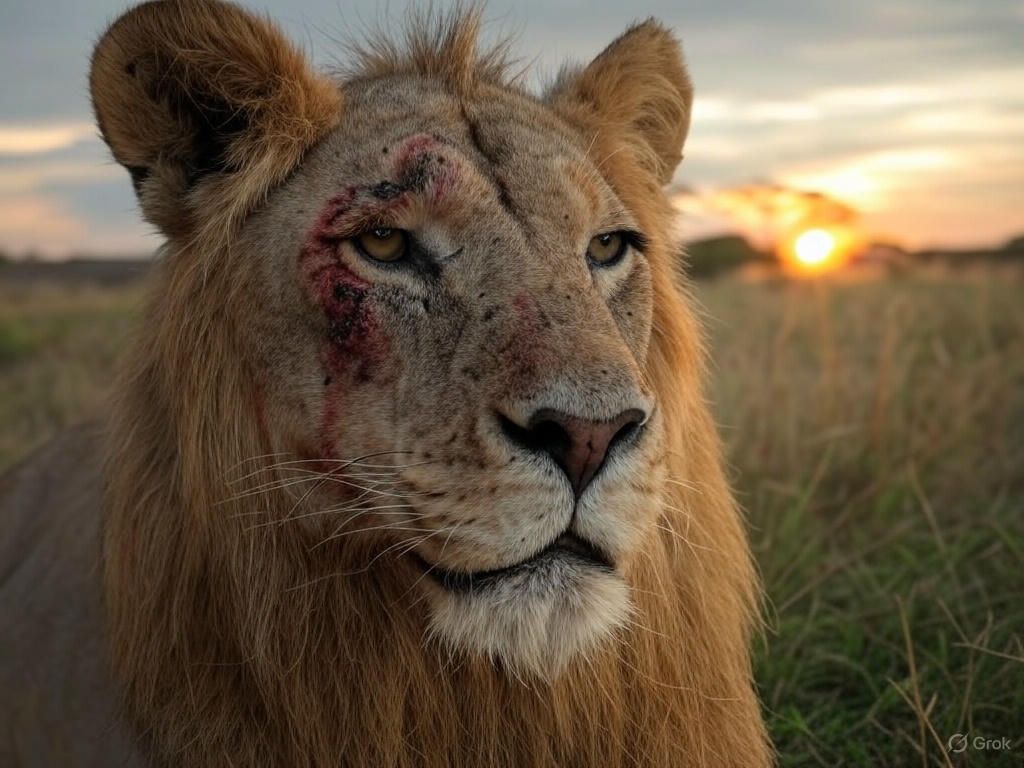}
        & \includegraphics[width=0.13\textwidth]{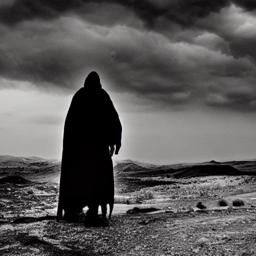} \\
      %—— 第五行：d ——
      \textbf{(d)} 
        & \includegraphics[width=0.13\textwidth]{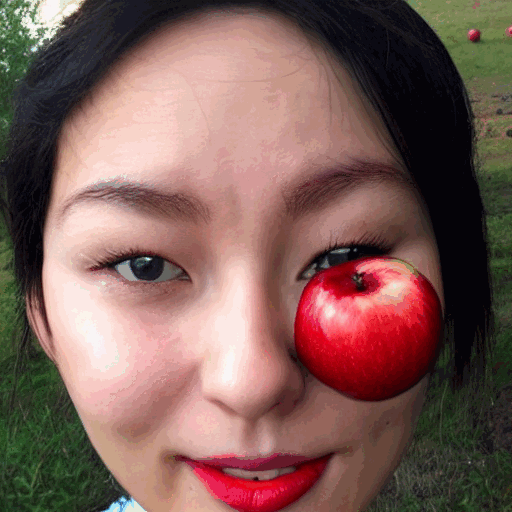}
        & \includegraphics[width=0.13\textwidth]{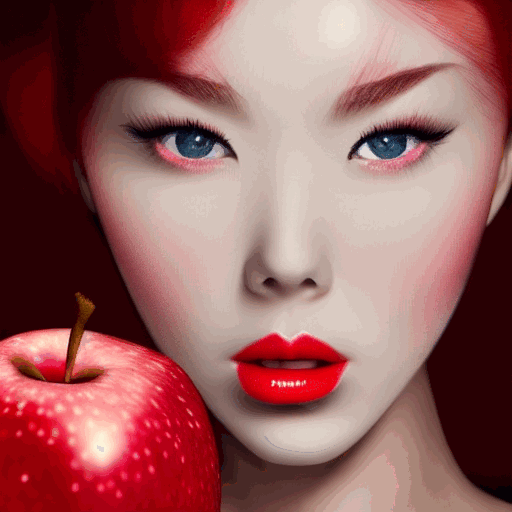}
        & \includegraphics[width=0.13\textwidth]{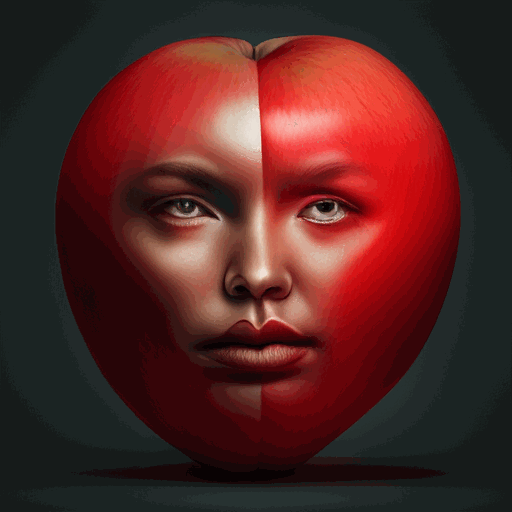}
        & \includegraphics[width=0.13\textwidth]{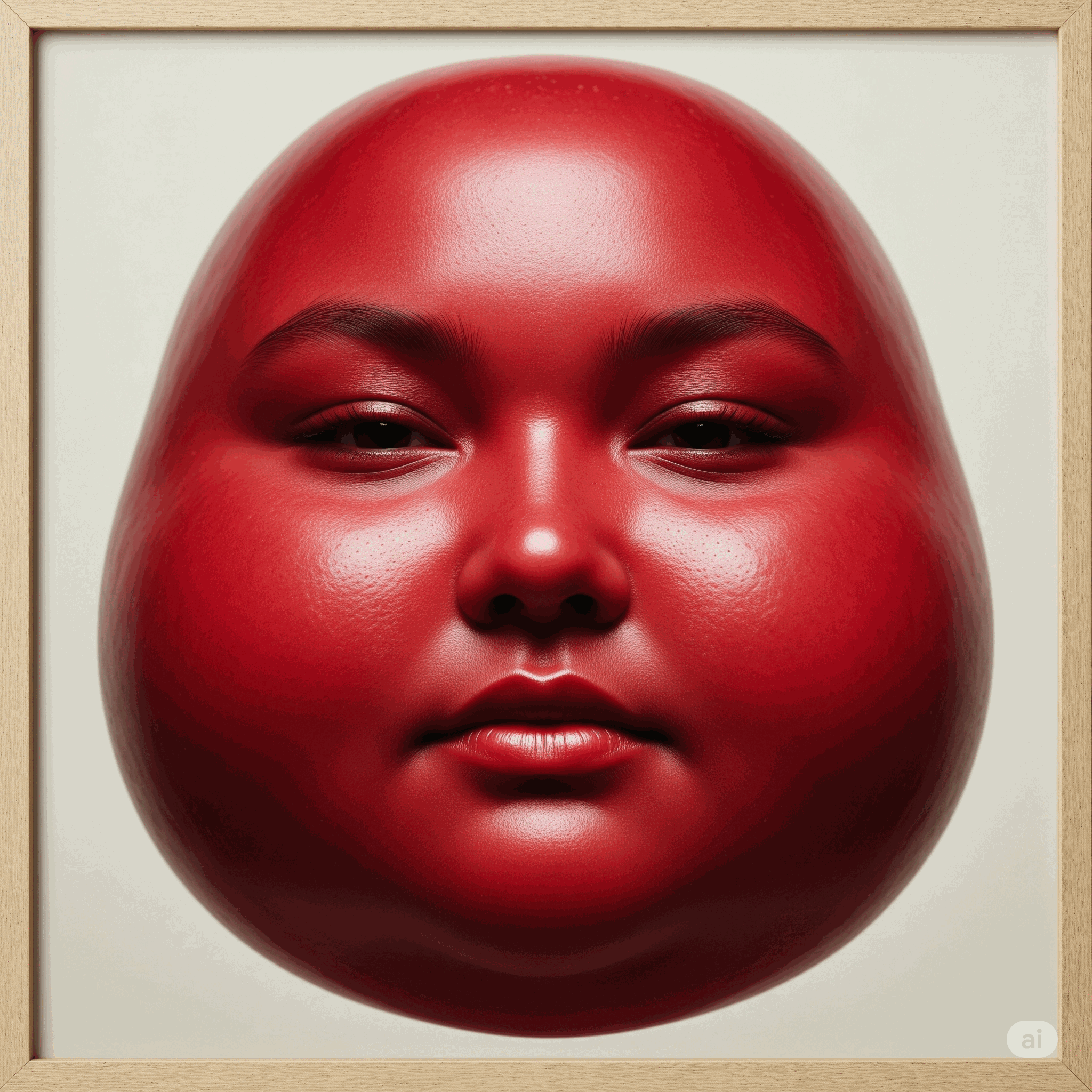}
        & \includegraphics[width=0.13\textwidth]{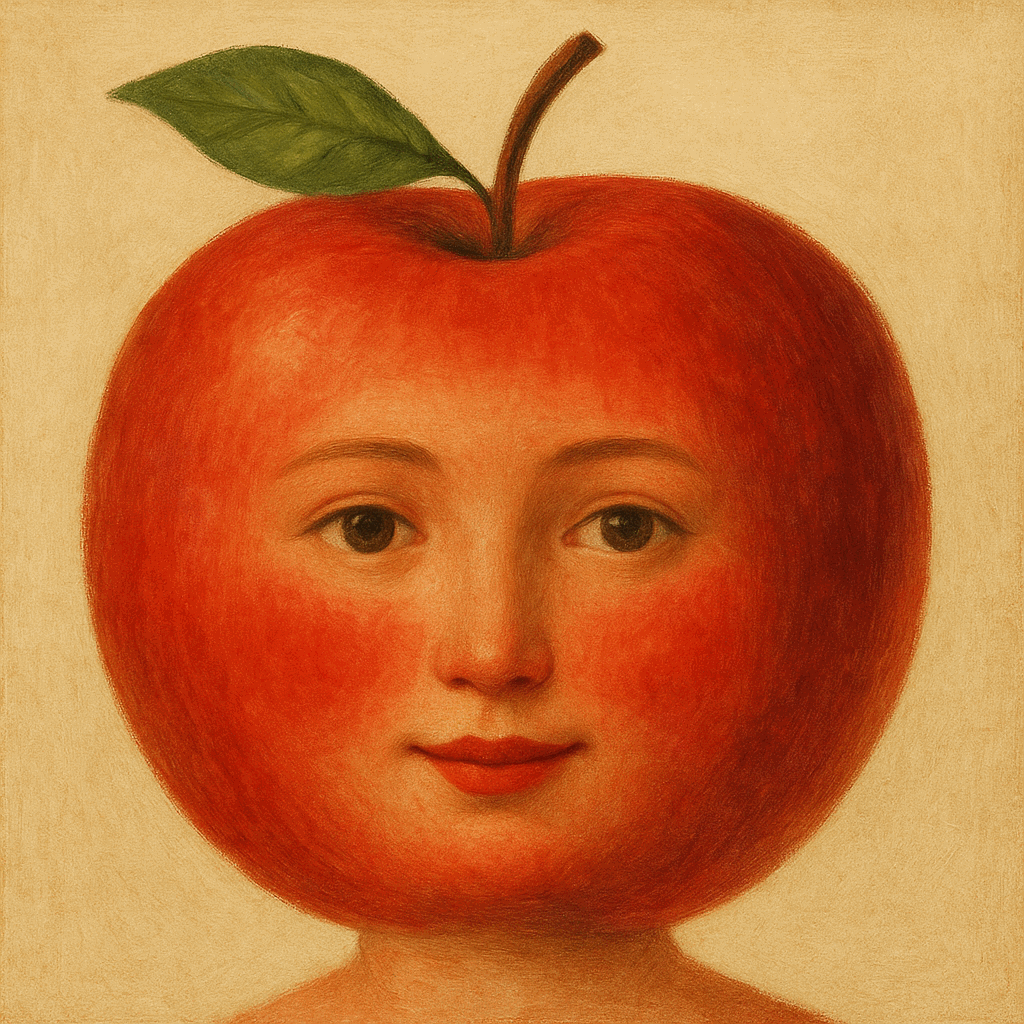}
        & \includegraphics[width=0.13\textwidth,height=0.13\textwidth, clip,trim=15pt 15pt 0pt 0pt]{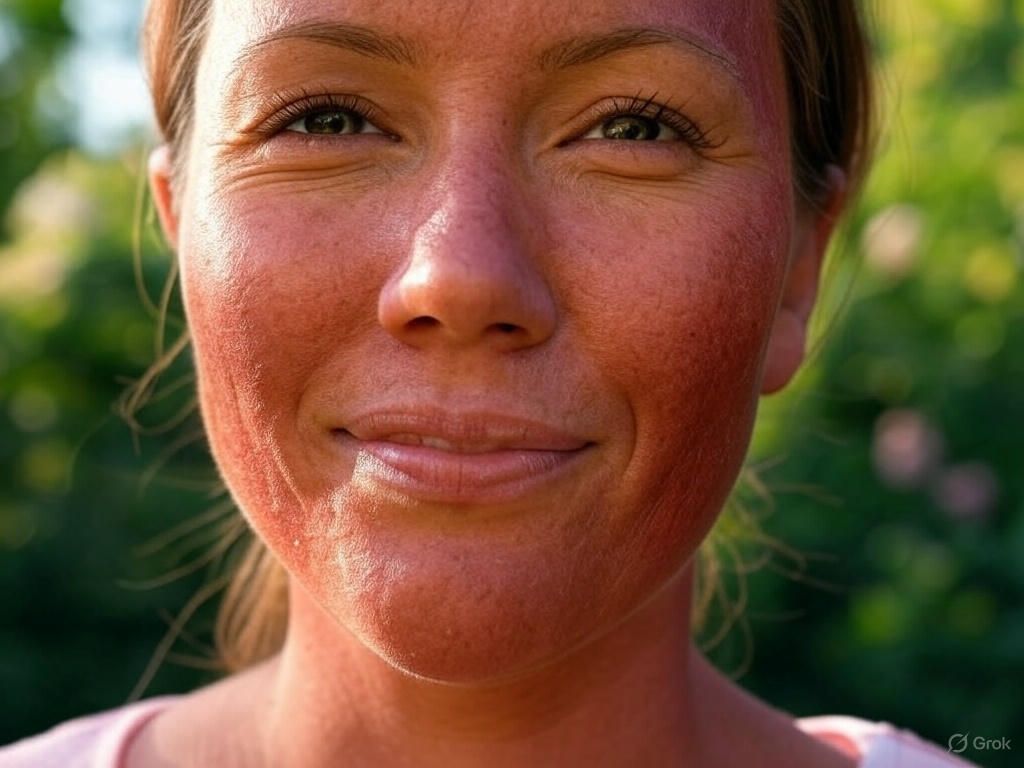}
        & \includegraphics[width=0.13\textwidth]{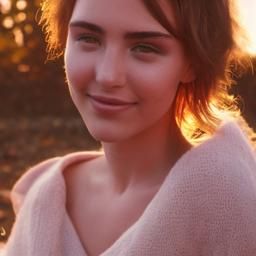} \\
    };
  \end{tikzpicture}

  \vspace{4pt}
  {\normalsize
  \begin{tabular}{ll}
    (a) & ``My bedroom is a pig sty.'' \\
    (b) & ``The classroom was like a zoo, buzzing with chaotic energy.'' \\
    (c) & ``He was looking like a wounded lion.'' \\
    (d) & ``Her face is like a red apple.'' \\
  \end{tabular}
  }

  \caption{\textbf{Qualitative comparison across models on rhetorical prompts.}
  Rhet2Pix generates semantically faithful and visually coherent images by correctly distinguishing metaphorical vehicles from intended subjects. Baseline models often fail to capture rhetorical meaning, producing literal or stylistically inconsistent outputs.}
  \label{fig:comparison}
\end{figure*}

\noindent\textbf{Dataset.}
% Many rhetorical devices—such as irony, allusion, and pun—either rely heavily on contextual or cultural knowledge or are grounded in linguistic form, making them difficult to represent visually. 
In this work, we focus on 
% rhetorical types that are more naturally aligned with visual expression. Specifically, we target 
metaphor (including personification) and simile, as they often describe concrete visual relationships between entities.
For data preparation, we adopt the FLUTE dataset \citep{chakrabarty2022flutefigurativelanguageunderstanding} and apply an additional filtering step to select high-quality metaphor and simile samples, ensuring both rhetorical clarity and visual interpretability for fine-tuning.

\noindent\textbf{Diffusion model.}
We use Stable Diffusion v1.4 \citep{rombach2022high} as the backbone generative model. Sampling is performed using the Denoising Diffusion Implicit Models (DDIM) algorithm \citep{song2020denoising}. For efficient parameter adaptation, we apply Low-Rank Adaptation (LoRA) \citep{hu2022lora} to the UNet \citep{ronneberger2015unetconvolutionalnetworksbiomedical} component of the diffusion model, enabling fine-tuning with minimal additional memory overhead.

\noindent\textbf{Experimental resources.}
Experiments were conducted using 8 NVIDIA A100 GPUs, each equipped with 80GB of HBM memory. Training each main baseline required approximately 8 GPU-days (1 day with 8 GPUs).

% \textbf{Additional Results.}
% Additional experimental results can be found in the appendix.

\subsection{Comparison Evaluation}
We compare our proposed method against representative baselines: Stable Diffusion (SD), a diffusion-based text-to-image model; DDPO, an RL-enhanced diffusion method fine-tuned by a VLM reward; MMaDA, Imagen, advanced multimodal diffusion foundation model; GPT-4o and Grok 3, advanced multimodal large language models.

\noindent\textbf{Qualitative evaluation.}
As shown in Figure~\ref{fig:comparison}, Rhet2Pix excels at generating images that capture the deep semantic intent of rhetorical language while maintaining precise control over visual content. Our method effectively distinguishes between the metaphorical vehicle and the intended subject, ensuring that the image emphasizes the correct visual referents and avoids misleading literal interpretations.

\begin{table*}[h]
  \centering
  
  \begin{tabular}{lccc}
    \toprule
    \textbf{Methods} & \textbf{Semantic‐alignment} & \textbf{Elemental‐alignment} & \textbf{Sum‐alignment} \\
    \midrule
    SD       & 0.2241 \textnormal{$\pm$ 0.0234}    & 0.0277 \textnormal{$\pm$ 0.2002} & 0.2518 \textnormal{$\pm$ 0.2191} \\
    DDPO     & 0.2190 \textnormal{$\pm$ 0.0405}    & 0.0168 \textnormal{$\pm$ 0.2148} & 0.2358 \textnormal{$\pm$ 0.2523} \\
    MMaDA    & 0.2160 \textnormal{$\pm$ 0.0160}    & 0.1746 \textnormal{$\pm$ 0.2754} & 0.3906 \textnormal{$\pm$ 0.2906} \\
    Imagen   & 0.2232 \textnormal{$\pm$ 0.0285}    & 0.1397 \textnormal{$\pm$ 0.2496}   & 0.3630 \textnormal{$\pm$ 0.2552} \\
    GPT-4o   & 0.2307 \textnormal{$\pm$ 0.0210}    & \(\mathbf{-}0.0965\) \textnormal{$\pm$ 0.0429} & 0.1342 \textnormal{$\pm$ 0.0499} \\
    Grok-3   & 0.2451 \textnormal{$\pm$ 0.0314}    & 0.1741 \textnormal{$\pm$ 0.1969}     & 0.4192 \textnormal{$\pm$ 0.2059} \\
    Rhet2Pix & 
      \colorbox{gray!20}{0.2654} \textnormal{$\pm$ 0.0172} & 
      \colorbox{gray!20}{0.4017} \textnormal{$\pm$ 0.0347} & 
      \colorbox{gray!20}{0.6671} \textnormal{$\pm$ 0.0517} \\
    \bottomrule
  \end{tabular}
  \caption{\textbf{Quantitative evaluation on alignment.} Rhet2Pix achieves strong performance in both semantic and elemental alignment and obtains the highest overall score.}
  \label{tab:quantitative}
\end{table*}

In contrast, baseline models have consistent limitations. SD, DDPO, MMaDA, and Imagen generate visually generic outputs that align only superficially with prompt keywords, lacking the ability to infer or represent metaphorical meaning. GPT-4o aligns strongly at the surface level but often merges subject and vehicle in unnatural ways (e.g., rendering a human head as an apple) and defaults to cartoon-style outputs for rhetorical prompts. Grok 3 occasionally captures figurative intent but frequently conflates subject and vehicle.

\begin{figure*}[h]
    \centering
    \includegraphics[width=0.7\linewidth]{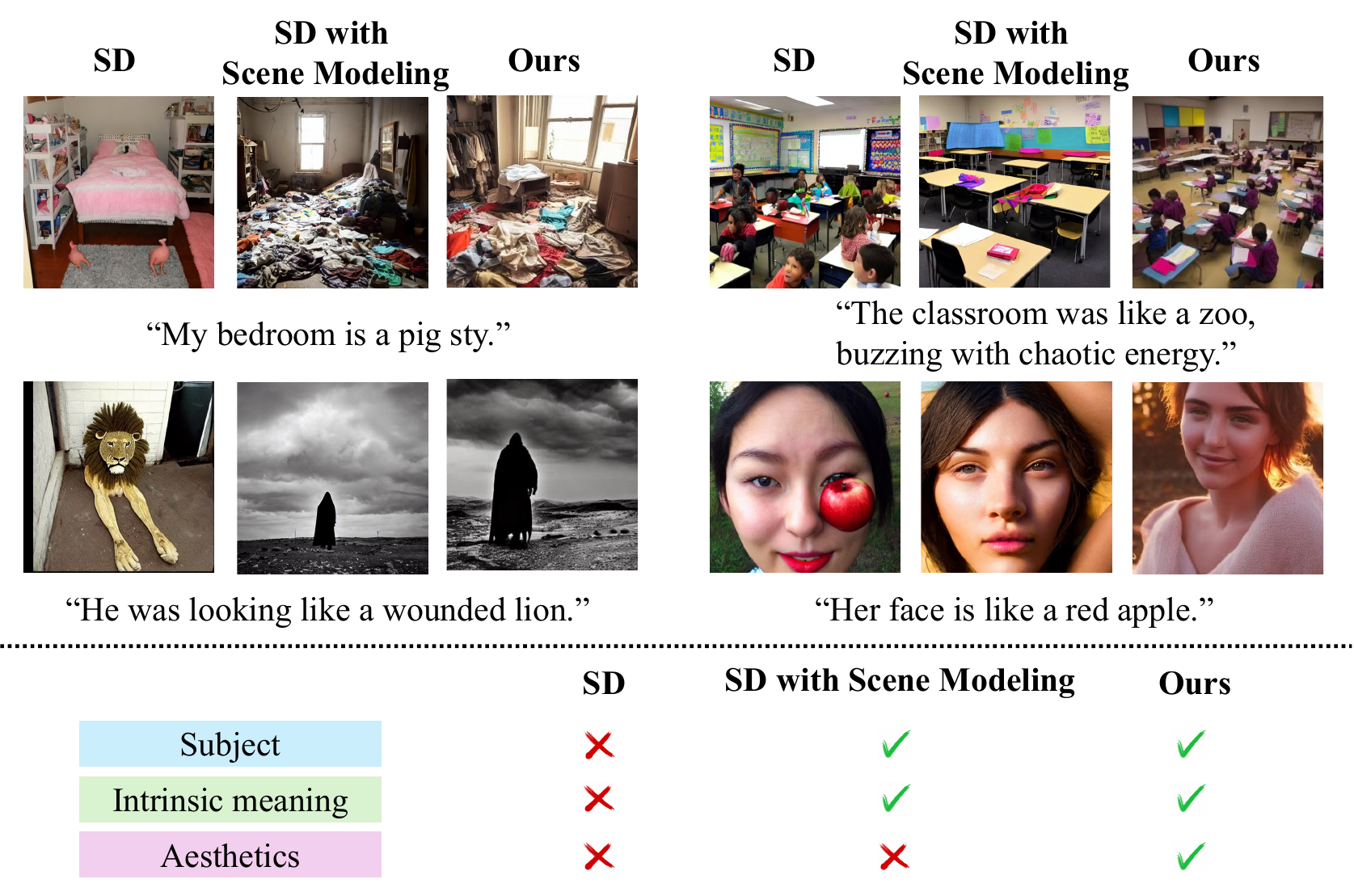}
    % \caption{\textbf{Ablation study results.} Ablation studies show that SD with scene modeling improves the interpretation of rhetorical text and refines visual elements, generating images that align better with the intended meaning. Our method builds on SD with scene modeling and further incorporates reinforcement learning to enhance both artistic quality and image details.}
    \caption{\textbf{Ablation study results.} Scene modeling enhances semantic alignment by refining the interpretation of rhetorical text, while our method further integrates reinforcement learning to improve artistic quality and image detail.}
    \label{fig:ablation}
\end{figure*}

\noindent\textbf{Quantitative evaluation.}
As reported in Table \ref{tab:quantitative}, Rhet2Pix performs consistently well across all alignment evaluation metrics, achieving a clear lead in overall performance. These results reflect its ability to balance high-level semantic understanding with precise control over visual elements when interpreting rhetorical prompts.

By contrast, SD, DDPO, MMaDA, and Imagen fail to reason about the overall intent of the input, producing images that do not reflect its full meaning. GPT-4o and Grok-3, while able to capture the general semantics thanks to large‐scale multimodal training, struggle to disentangle subject and metaphor. Their focus on literal text often elevates the metaphorical vehicle to the main subject, misrepresenting the intended rhetorical meaning.

\subsection{Ablation Studies}

As shown in Figure \ref{fig:ablation}, scene modeling (prompt engineering) and reinforcement learning (RL) are crucial for improving the alignment between rhetorical text and image generation. Scene modeling enhances the interpretation of rhetorical text by transforming it into more detailed and semantically enriched prompts, resulting in images that better reflect the intended meaning. However, using scene modeling alone still produces images with an unclear main subject (e.g., the person appears too small in the bottom-left panel) and erroneous details (e.g., the noisy classroom in the top-right panel contains no people). RL refines this process by guiding the model towards higher artistic quality and richer detail. Together, these components enable more precise and expressive image generation.

\section{Conclusions}

In this work, we identify a critical misalignment in rhetorical text-to-image generation, rooted in the direct use of metaphorical vehicle embeddings that drive literal imagery and obscure intended meanings. To overcome this, we propose Rhet2Pix, a framework that formulates rhetorical image generation as a two-layer, multi-step Markov Decision Process.  
% In the outer layer, our approach refines input rhetorical text into a sequence of semantically enriched prompts that guide the progressive refinement of visual detail. In the inner layer, Rhet2Pix discounts the final reward along the diffusion denoising trajectory to optimize adjacent action pairs and mitigate sparse feedback. 
Experimental results show that Rhet2Pix effectively captures the rhetorical meaning while precisely controlling the elements and their characteristics in the generated images.
This framework paves the way for more nuanced multimodal generation grounded in complex linguistic expressions.

% \newpage

\bibliography{main}


\newpage
\appendix

\section{Discussion on Multi-stage Image Generation}

\textit{(1) Why do we adopt multi-stage image generation?}

Existing text-to-image pipelines adopt the input text as prompts and convert it into word embeddings.
The lack of genuine semantic understanding causes generated images to merely depict isolated visual elements of individual words, missing the deeper meaning conveyed by the text.
Even approaches that use advanced multimodal models to parse the text into high-level semantic prompts before embedding still fall short: by feeding the literal subject, metaphorical vehicle, and environmental context simultaneously and without distinction, the model’s generation capacity is spread across competing objectives. Consequently, the intended focal subject loses prominence, often appearing blurred or de-emphasized, while background elements or metaphorical vehicles become the dominant factors shaping the generated image.

\begin{figure*}[tb]
    \centering
    \includegraphics[width=1\linewidth]{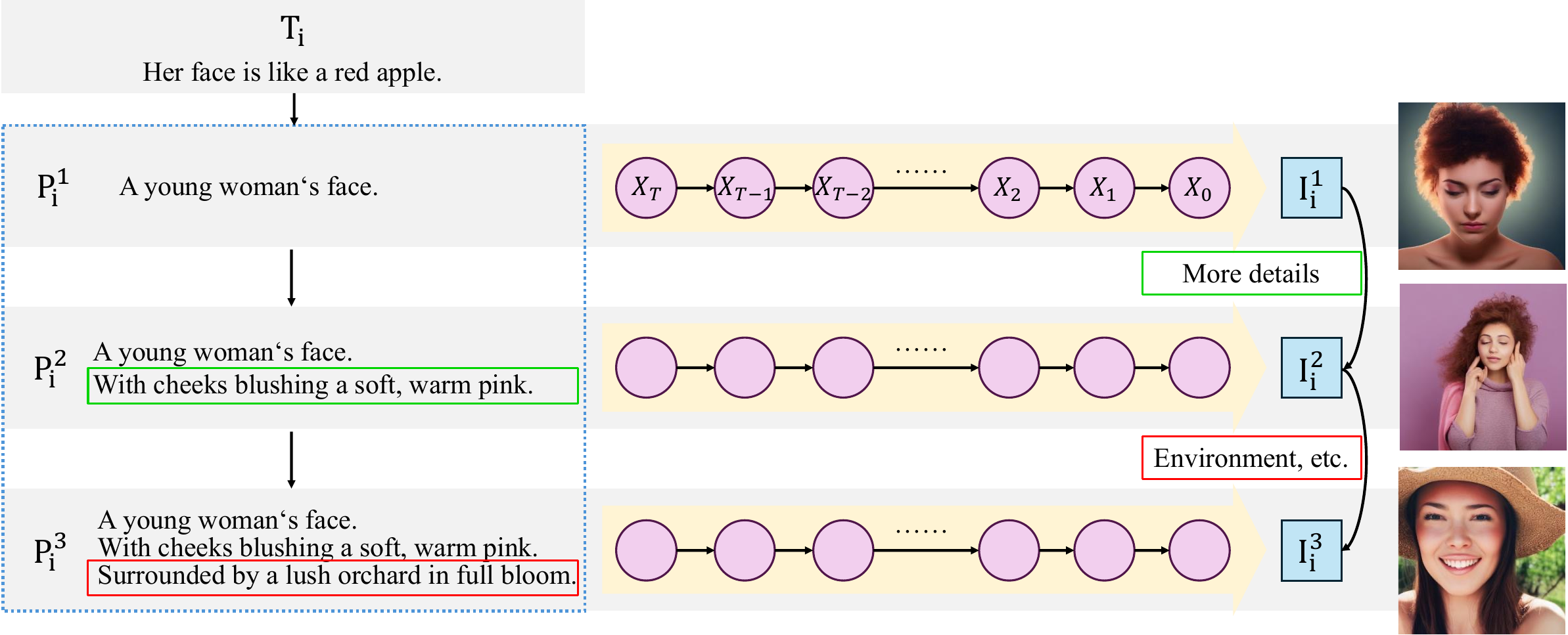}
    \caption{Illustrative example of the multi-stage image generation pipeline in Rhet2Pix.}
    \label{fig:multi-stage}
\end{figure*}

Taking these into consideration,
We propose a multi‐stage framework that structures both prompt generation and RL fine‐tuning into sequential phases, producing images that faithfully convey the text’s meaning and exhibit rich visual detail. Figure \ref{fig:multi-stage} provides an illustrative walkthrough of this process.

\begin{itemize}
    \item \textbf{Multi-Stage Prompting for Semantic Alignment.}
    We recursively derive a sequence of scene modeling prompts: starting with a prompt focused solely on the core subject, we then introduce prompts that gradually incorporate contextual, lighting, and compositional details—preserving subject clarity while progressively enhancing the scene's richness. During training, a stage‐wise semantic‐alignment reward preserves subject priority by assigning a higher weight to early, subject‐focused prompts.
    
    \item \textbf{Multi-Stage Dependency Optimization for Detail Enrichment.}
    As the multi-stage prompts drive the generation of a sequence of interdependent images, each image incrementally enriches the previous one by introducing additional semantic details. We model this as a MDP where each image is a state, and the semantic difference between prompts is the action driving state transitions. Using Generalized Advantage Estimation (GAE), we compute an advantage for each intermediate state, which corresponds to the generated images, and propagate these values through the diffusion denoising steps to update the policy. By emphasizing advantages from later stages, the model learns to focus on progressively enriching details, leading to final images with enhanced vividness and clarity.

\end{itemize}

\textit{(2) How do we construct a multi-stage image generation process?}

In Rhet2Pix, we implement multi-stage generation as a two-layer MDP that decouples semantic planning from pixel-level synthesis. For each rhetorical text input $T_i$, we first construct a sequence of $C$ scene instructions $\left\{P_i^1, P_i^2, \ldots, P_i^C\right\}$, where each prompt $P_i^j$ is generated by conditioning on both the validated semantic factors and the immediately preceding prompt $P_i^{j-1}$. This design ensures that each stage builds upon the last: $P_i^1$ establishes a precise description of the core subject, and each subsequent $P_i^j$ introduces additional details- environmental context, lighting cues, compositional elements, and emotional tone-while preserving earlier content.

Formally, we view this as an outer-layer MDP: the state at stage $j$ is the image $I_i^j$ synthesized from $P_i^j$, and the action is the semantic increment $\Delta_i^j$ that transforms $P_i^j$ into $P_i^{j+1}$. 
Through Generalized Advantage Estimation (GAE), we compute an advantage $\hat{A}_i^j$ for each state (i.e., each generated image) that quantifies its marginal benefit toward fulfilling the accumulated semantic objectives. 

Simultaneously, the inner-layer MDP governs the diffusion denoising trajectory for any given prompt $P_i^j$: at timestep $t$, the state $\mathbf{s}_t=\left(\mathbf{c}, t, \mathbf{x}_t\right)$ transitions via action $\mathbf{x}_{t-1}$ under policy $\pi_\theta\left(\mathbf{x}_{t-1} \mid \mathbf{x}_t, \mathbf{c}\right)$, where $\mathbf{c}$ is encoded from $P_i^j$. By backpropagating outer-layer advantages through this inner trajectory, we align low-level denoising updates with high-level semantic objectives.

\begin{figure}[htbp]
    \centering
    \includegraphics[width=1\linewidth]{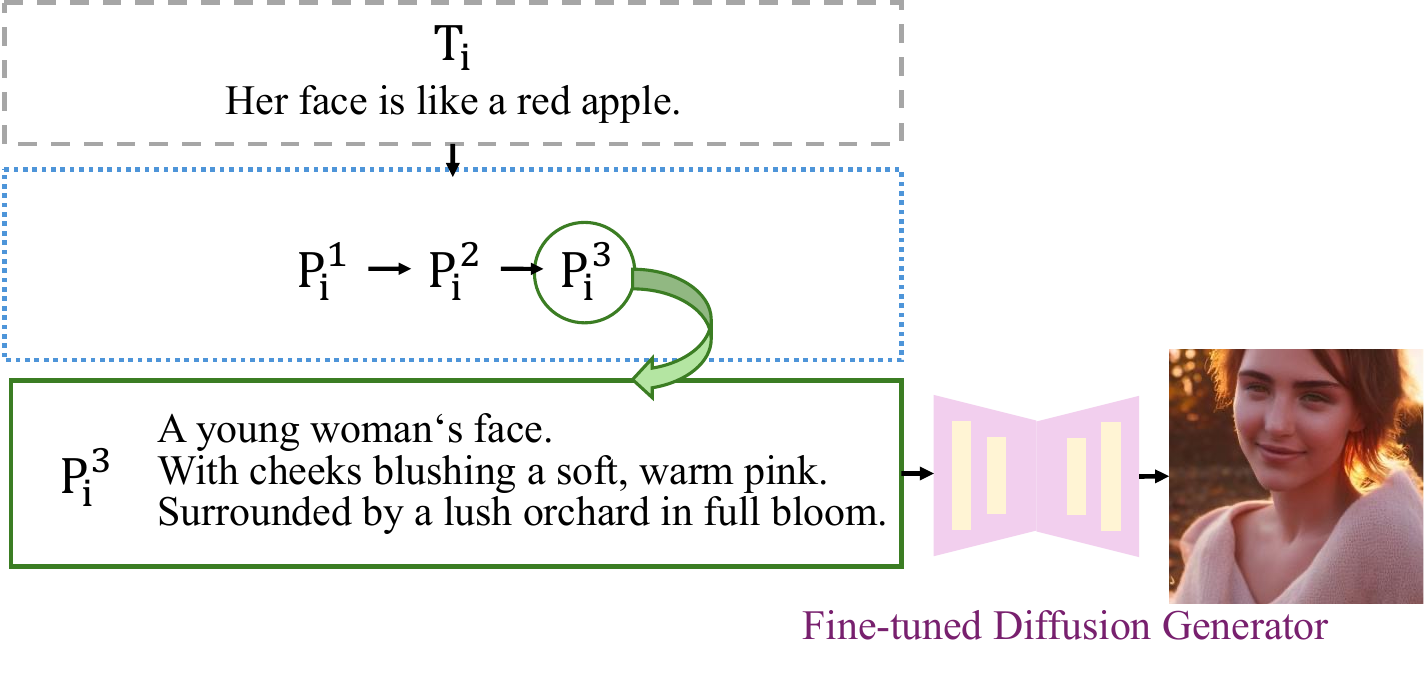}
    \caption{Illustrative example of the inference process of Rhet2Pix.}
    \label{fig:inference}
\end{figure}

At inference, as shown in Figure \ref{fig:inference}, only the final prompt $P_i^C$ is passed to the fine-tuned diffusion generator to produce the output image. All intermediate prompts and images serve exclusively for advantage computation and policy learning, rather than as visible outputs. Unlike video or comic generation-which present a sequence of frames or panels by issuing a lot of prompts-our method iteratively refines few images(e.g. 3 images) through successive semantic prompts and outputs only the final, richly detailed result.

\section{Discussion on sparse rewards}

\textit{(1) What is the sparse reward problem in diffusion-based text-to-image generation?}

When we use reinforcement learning to fine-tune a pre-trained Stable Diffusion model, the only reward signal originates from the final denoised image $\mathbf{x}_0$, leaving all intermediate denoising actions without direct feedback. Methods like DDPO address this by assigning the same terminal reward $R\left(\mathbf{x}_0\right)$ to every timestep.
\[
r_{T-1}=r_{T-2}=\cdots=r_0=R\left(\mathbf{x}_0\right)
\]

However, the denoising actions $a_t$ at different timesteps have different effects on alignment: early iterations establish layout and overall style, whereas later iterations refine texture and fine detail \citep{hu2025towards}. Treating each step identically disperses the reward across tasks of unequal importance, resulting in noisy updates, slower convergence, and reduced capacity to capture subtle visual features.

\textit{(2) How does our method effectively address sparse reward propagation?}

As noted above, denoising actions at different timesteps contribute unevenly to the final image—early iterations set global layout, middle iterations refine style, and late iterations enhance object-level detail.  Given our RL objective of semantic and visual alignment, we solve the reward sparisty along the trajectory by propagating the terminal advantage back through all diffusion steps via
\[
\hat A_t^{(i,j)} = \gamma_{\mathrm{denoise}}^t\,\hat A_i^j.
\]

By propagating the terminal advantage backwards along the diffusion trajectory with a decay factor, diffusion steps closer to the final image receive proportionally greater advantage, reflecting their critical role in semantic and visual refinement. This mechanism overcomes the sparse reward problem in intermediate denoising steps and guides the model more efficiently toward our alignment objectives.

% \newpage

\section{Limitations}

Rhet2Pix leverages reinforcement learning to fine-tune the alignment between rhetorical semantics and visual output. However, fully addressing the challenges of rhetorical text-to-image generation may necessitate a fundamentally new backbone—one that surpasses conventional text-image embedding similarity. Achieving true end-to-end rhetorical generation requires a deep understanding of the rhetorical intent embedded in the input text. This presents a substantial challenge and remains an ambitious avenue for future research.

\section{Notation}
The list of important symbols and their corresponding definition in this paper goes as Table \ref{tab:notation}.

\begin{table*}[htb]
  \centering
  \renewcommand{\arraystretch}{1.2}
  \caption{Notation table}
  \begin{tabular}{cl}
    \toprule
    \textbf{Symbol} & \textbf{Description} \\
    \midrule
    $T_i$                             & Input rhetorical texts\\
    $F_i$                             & Key factors decomposed from $T_i$ for scene modeling and reward evaluation \\
    $P_i^j$                           & Multi-stage prompts for input text $T_i$ \\
    $I_i^j$                           & Image generated from prompt $P_i^j$ \\
    $r_i^j$                           & Reward of image $I_i^j$ \\
    $\mathcal{D}$                     & Replay buffer storing denoising trajectories of generated images \\
    $\mathbf{c}$                      & Embedding vector produced from prompt $P_i^j$ \\
    $\mathbf{x}_t$                    & Diffusion model’s latent tensor at timestep $t$ \\
    $\mathbf{s}_t = (\mathbf{c},\,t,\,\mathbf{x}_t)$
                                      & Tuple of prompt embedding, timestep index, and latent tensor \\
    $\mathbf{a}_t = \mathbf{x}_{t-1}$ & Updated latent tensor after one diffusion step \\
    $V_\phi(I_i^j)$                   & State value of image $I_i^j$ estimated by the critic network with parameters $\phi$ \\
    $\delta_i^j$                      & TD residual used in GAE computation\\
    $\hat A_i^j$                      & Advantage of image $I_i^j$ computed via GAE \\    
    $\gamma_{\mathrm{denoise}}$       & Discount factor applied across diffusion timesteps \\
    $\hat A_t^{(i,j)}$ & Advantage at diffusion step $t$ for prompt $P_i^j$ \\
    $w_t$                             & Likelihood ratio used in PPO update \\
    $\epsilon$                        & Clipping threshold in PPO, bounding $w_t$ to ensure stable updates \\
    
    \bottomrule
  \end{tabular}
  \label{tab:notation}
\end{table*}

\section{Pseudo-code}
The pseudo-code of Rhet2Pix for one training round goes as Algorithm \ref{alg:Rhet2Pix}.

\begin{algorithm*}[htb]
  \caption{Pseudo-code for Rhet2Pix}
  \label{alg:Rhet2Pix}
  \textbf{Input}: Rhetorical texts $\{T_i\}_{i=1}^N$, number of stages $C$, diffusion steps $T$, pretrained diffusion model, LLM, reward function $R$.\\
  \textbf{Output}: Optimized policy parameters $\theta$, value parameters $\phi$.
  \begin{algorithmic}[1]
    \STATE \textbf{Multi-stage scene modeling with self-validation}
    \FOR{$i=1$ \TO $N$}
      \STATE $k \leftarrow 1$
      \REPEAT
        \STATE $F_i^{(k)} \leftarrow \mathrm{LLM}(T_i)$
        \STATE $k \leftarrow k + 1$
      \UNTIL{$\mathrm{verify}(F_i^{(k)},T_i)=1$}
      \STATE $F_i \leftarrow F_i^{(k)}$ \COMMENT{Generate validated factors}
      \STATE $P_i^1 \leftarrow \mathrm{LLM}(T_i, F_i)$
      \FOR{$j=2$ \TO $C$}
        \STATE $P_i^j \leftarrow \mathrm{LLM}(P_i^{j-1},\text{next factor})$ \COMMENT{Generate multi-stage prompts}
      \ENDFOR
    \ENDFOR

    \STATE \textbf{Sampling (Two-layer MDP rollout)}
    \STATE Initialize $\mathcal{D}\!\leftarrow\!\emptyset$
    \FOR{$i=1$ \TO $N$}
      \FOR{$j=1$ \TO $C$}
        \STATE $\mathbf{c}\leftarrow \mathrm{Embed}(P_i^j)$
        \STATE Sample $\mathbf{x}_T\sim\mathcal{N}(0,I)$
        \FOR{$t=T$ \TO $1$ \textbf{step} $-1$}
          \STATE $\mathbf{s}_t\leftarrow(\mathbf{c},t,\mathbf{x}_t)$
          \STATE $\mathbf{a}_t\sim \pi_\theta(\cdot\mid\mathbf{s}_t)$
          \STATE $\mathbf{x}_{t-1}\leftarrow \mathbf{a}_t$
          \IF{$t=1$}
            \STATE $r_i^j\leftarrow R(\mathbf{x}_0,P_i^j,F_i)$ \COMMENT{Reward only for the final step}
          \ELSE
            \STATE $r_i^j\leftarrow 0$
          \ENDIF
          \STATE store $(\mathbf{s}_t,\mathbf{a}_t,t,r_i^j)$ in $\mathcal{D}$
        \ENDFOR
      \ENDFOR
    \ENDFOR

    \STATE \textbf{Training}
    \FOR{$i=1$ \TO $N$}
      \FOR{$j=1$ \TO $C$}
        \STATE $\delta_i^j \leftarrow r_i^j + \gamma\,V_\phi(I_i^{j+1}) - V_\phi(I_i^j)$
        \STATE $\hat A_i^j \leftarrow \sum_{l=0}^{C-j} (\gamma\lambda)^l\,\delta_i^{\,j+l}$ \COMMENT{Outer-layer advantage via GAE}
      \ENDFOR
    \ENDFOR
    \FORALL{$(\mathbf{s}_t,\mathbf{a}_t,t,r_i^j)\in\mathcal{D}$}
      \STATE $\hat A_t^{(i,j)} \leftarrow \gamma_{\mathrm{denoise}}^{\,t}\,\hat A_i^j$ \COMMENT{Inner-layer advantage propagation}
      \STATE update transition to $(\mathbf{s}_t,\mathbf{a}_t,t,r_i^j,\hat A_t^{(i,j)})$ in $\mathcal{D}$
    \ENDFOR

    \STATE Shuffle and partition $\mathcal{D}$ into minibatches
    \FORALL{minibatch $\mathcal{B}$}
      \FORALL{$(\mathbf{s}_t,\mathbf{a}_t,t,r_i^j,\hat A_t^{(i,j)})\in\mathcal{B}$}
        \STATE $w_t \leftarrow \dfrac{\pi_\theta(\mathbf{a}_t\mid\mathbf{s}_t)}{\pi_{\theta_{\mathrm{old}}}(\mathbf{a}_t\mid\mathbf{s}_t)}$
        \STATE $\mathcal{L}_{\mathrm{PPO}} \leftarrow \mathbb{E}\bigl[\min(w_t\,\hat A_t^{(i,j)},\,\mathrm{clip}(w_t,1-\epsilon,1+\epsilon)\,\hat A_t^{(i,j)})\bigr]$ \COMMENT{PPO update}
        \STATE $\mathcal{L}_V \leftarrow \mathbb{E}\bigl[(V_\phi(\mathbf{s}_t)-\hat A_t^{(i,j)})^2\bigr]$
      \ENDFOR
      \STATE update $\theta$ by minimizing $\mathcal{L}_{\mathrm{PPO}}$
      \STATE update $\phi$ by minimizing $\mathcal{L}_V$
    \ENDFOR

  \end{algorithmic}
\end{algorithm*}

\section{Hyperparameters}

We list model configuration and training hyperparameters of our method in Table \ref{tab:combined-config}.

\begin{table*}[h]
  \centering
  \caption{Model configuration and training hyperparameters.}
  \begin{tabular}{llc}
    \toprule
    \textbf{Module}               & \textbf{Hyperparameter}         & \textbf{Value}                  \\
    \midrule
                                  & Patch Size                      & 8                               \\
                                  & Transformer Depth               & 1                               \\
    ViT Encoder                   & Embedding Dimension             & 128                             \\
                                  & Attention Heads                 & 4                               \\
                                  & Activation Function             & GELU                            \\
    \midrule
                                  % & Embedding Style                 & \texttt{embed2}                 \\
                                  & Embedding Normalization         & 
                                  % 0 (
                                  Disabled
                                  % )
                                  \\
    Patch Embedding               & Conv1 (Kernel / Stride)         & 8 / 4                           \\
                                  & Conv2 (Kernel / Stride)         & 3 / 2                           \\
                                  & GroupNorm                       & Disabled                        \\
                                  & Patch Embed Output Dim          & 128                             \\
    \midrule
                                  & MLP Expansion Ratio             & 4×                              \\
    Transformer Block             & Dropout Rate                    & 0.0                             \\
                                  & Attention Type                  & Flash Attention                 \\
    \midrule
                                  & Projection Dimension            & 128                             \\
    Spatial Embedding             & Spatial Embedding Dropout       & 0.0                             \\
                                  & Weight Initialization           & \(\mathcal{N}(0,0.02^2)\)       \\
    \midrule
                                  & Input Feature                   & ViT + Spatial Emb               \\
                                  & MLP Hidden Dimensions           & [256, 256, 256]                 \\
    RGB Critic                    & Activation Function             & Mish                            \\
                                  & Residual Style                  & Enabled                         \\
                                  & LayerNorm                       & Disabled                        \\
                                  & Output Dimension                & 1                               \\
    \midrule
                                  & Train Batch Size                & 3                               \\
    Training                      & Gradient Accumulation Steps     & 1                               \\
                                  & Image Conditioning Steps        & 1                               \\
    \midrule
                                  & Denoising Steps                 & 50                              \\
    DDIM Sampling                 & Guidance Scale                  & 5.0                             \\
                                  & Noise Weight (\(\eta\))         & 1.0                             \\
                                  & Scheduler Type                  & DDIM                            \\
    \midrule
                                  & Optimizer                       & AdamW                           \\
                                  & Learning Rate                   & \(3\times10^{-4}\)              \\
    UNet Optimizer                & Betas                           & (0.9, 0.999)                    \\
                                  & Weight Decay                    & \(1\times10^{-4}\)              \\
                                  & \(\epsilon\)                    & \(1\times10^{-8}\)              \\
    \midrule
                                  & Optimizer                       & AdamW                           \\
                                  & Learning Rate                   & \(1\times10^{-3}\)              \\
    Critic Optimizer              & Betas                           & (0.9, 0.999)                    \\
                                  & Weight Decay                    & \(1\times10^{-4}\)              \\
                                  & \(\epsilon\)                    & \(1\times10^{-8}\)              \\
    \bottomrule
  \end{tabular}
  \label{tab:combined-config}
\end{table*}

\section{More experimental results}
\subsection{Quantitative evaluation}
\begin{figure}
    \centering
    \includegraphics[width=0.8\linewidth]{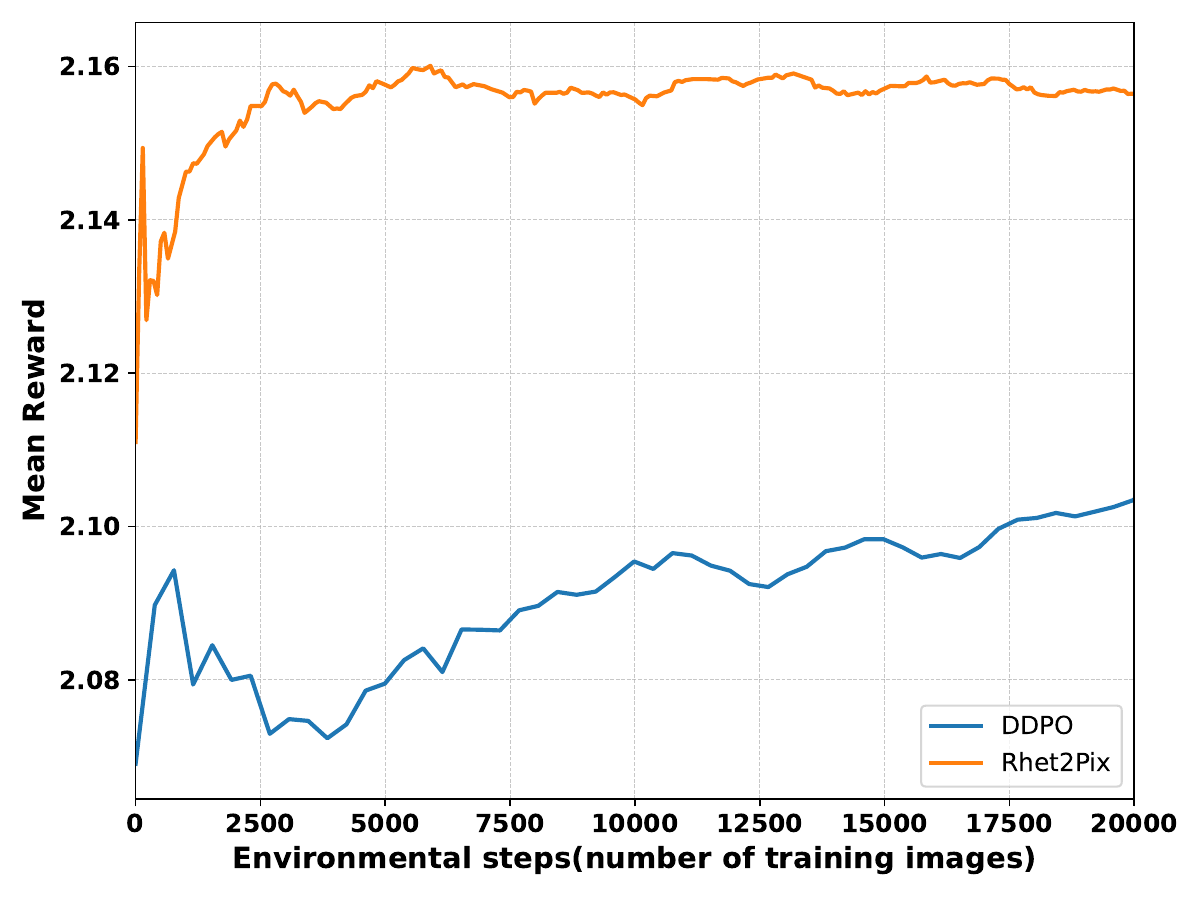}
    \caption{Reward curve of our method and DDPO when fine-tuning the diffusion model.}
    \label{fig:reward_curve}
\end{figure}

We apply our framework to fine-tune the diffusion model. Figure \ref{fig:reward_curve} displays the mean reward curve when fine-tuning the Stable Diffusion model using our method and DDPO as environmental steps (number of training images) increases.  To highlight its underlying progression, both lines are smoothed by the exponential moving average (EMA) with the same parameter. 

Under the same reward function and identical evaluation prompts from the rhetorical dataset, our method attains a higher reward than DDPO. Moreover, it recovers more quickly from the pronounced oscillations and initial declines during the warm-up phase, entering the correct performance regime sooner.

\newpage

\subsection{More samples}

In this section, we present additional samples generated by Rhet2Pix, alongside outputs from several strong baselines, including Stable Diffusion (SD), DDPO, GPT-4o, and Grok-3. All samples of Rhet2Pix are generated from the 50-th epoch from the same network weights in the same GPU.

\begin{figure*}[htbp]
  \centering
  %----- 第一组 subfigure -----
  \begin{subfigure}[b]{\linewidth}
  \captionsetup{labelformat=empty}
    \centering
    % 五张图，每张图都包在一个 minipage 里，上方手写标题
    \begin{minipage}[b]{0.14\linewidth}
      \centering
      {\small \textbf{SD}}\\[.5ex]
      \includegraphics[width=\linewidth]{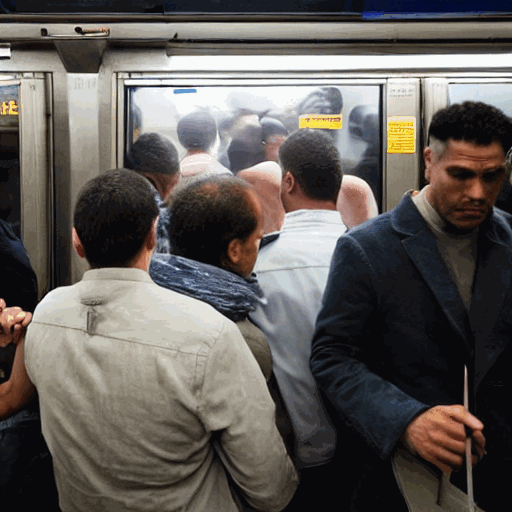}
    \end{minipage}\hfill
    \begin{minipage}[b]{0.14\linewidth}
      \centering
      {\small \textbf{DDPO}}\\[.5ex]
      \includegraphics[width=\linewidth]{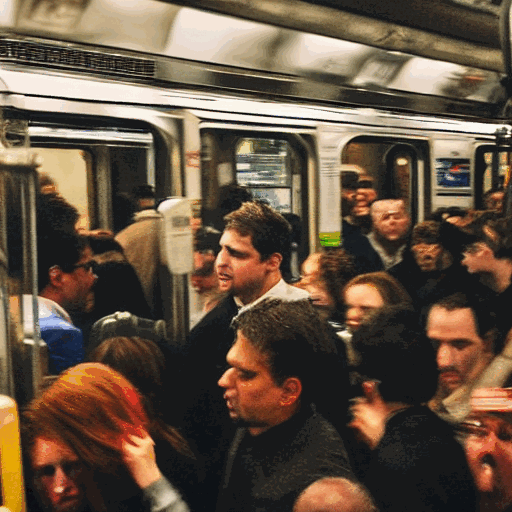}
    \end{minipage}\hfill
    \begin{minipage}[b]{0.14\linewidth}
      \centering
      {\small \textbf{MMaDA}}\\[.5ex]
      \includegraphics[width=\linewidth]{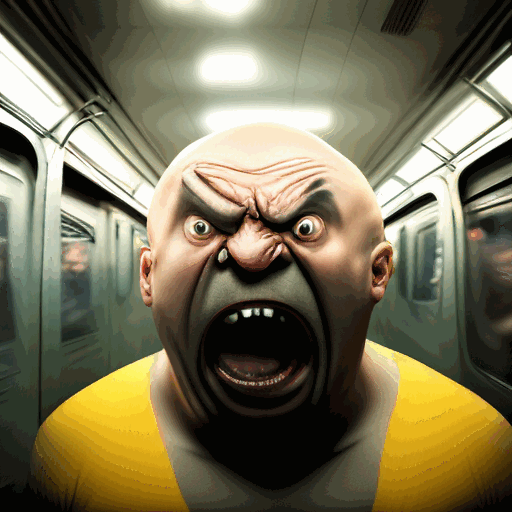}
    \end{minipage}\hfill
    \begin{minipage}[b]{0.14\linewidth}
      \centering
      {\small \textbf{Imagen}}\\[.5ex]
      \includegraphics[width=\linewidth]{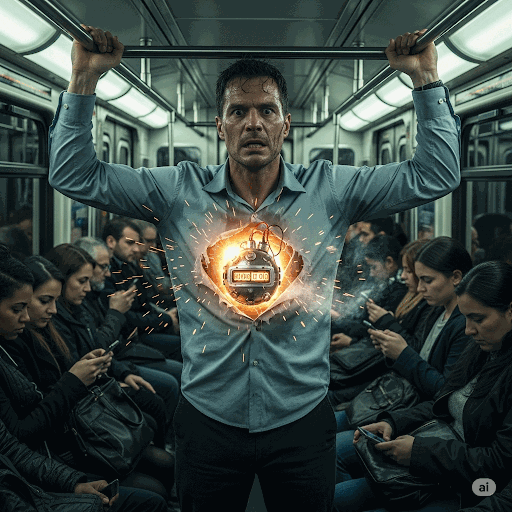}
    \end{minipage}\hfill
    \begin{minipage}[b]{0.14\linewidth}
      \centering
      {\small \textbf{GPT-4o}}\\[.5ex]
      \includegraphics[width=\linewidth]{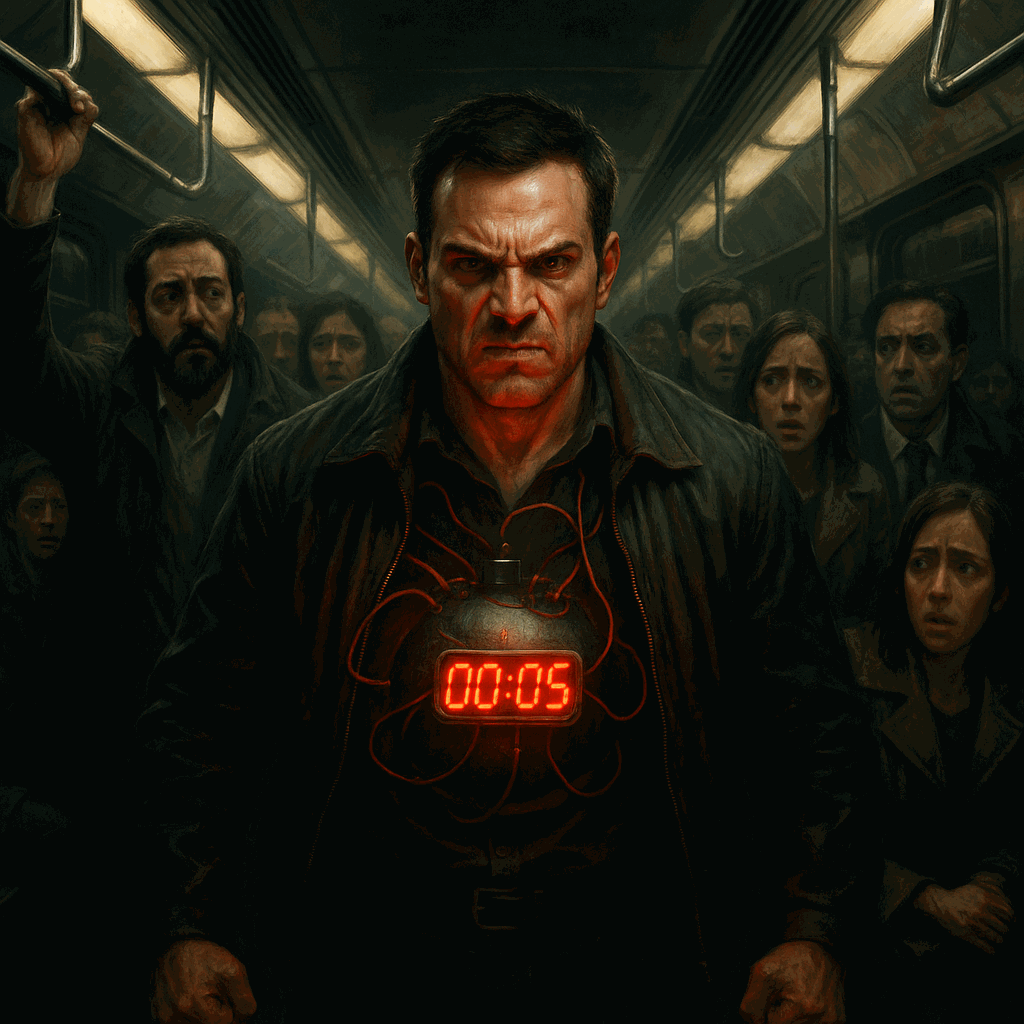}
    \end{minipage}\hfill
    \begin{minipage}[b]{0.14\linewidth}
      \centering
      {\small \textbf{Grok-3}}\\[.5ex]
      \includegraphics[
        width=\linewidth,          % 宽度占满 minipage
        height=\linewidth,         % 高度也等于宽度 -> 正方形
        keepaspectratio=false,     % 允许改变长宽比
        clip,                      % 启用裁剪
        trim=0 5cm 0 3cm           % 左 下 右 上，各裁下 0,3cm,0,3cm
      ]{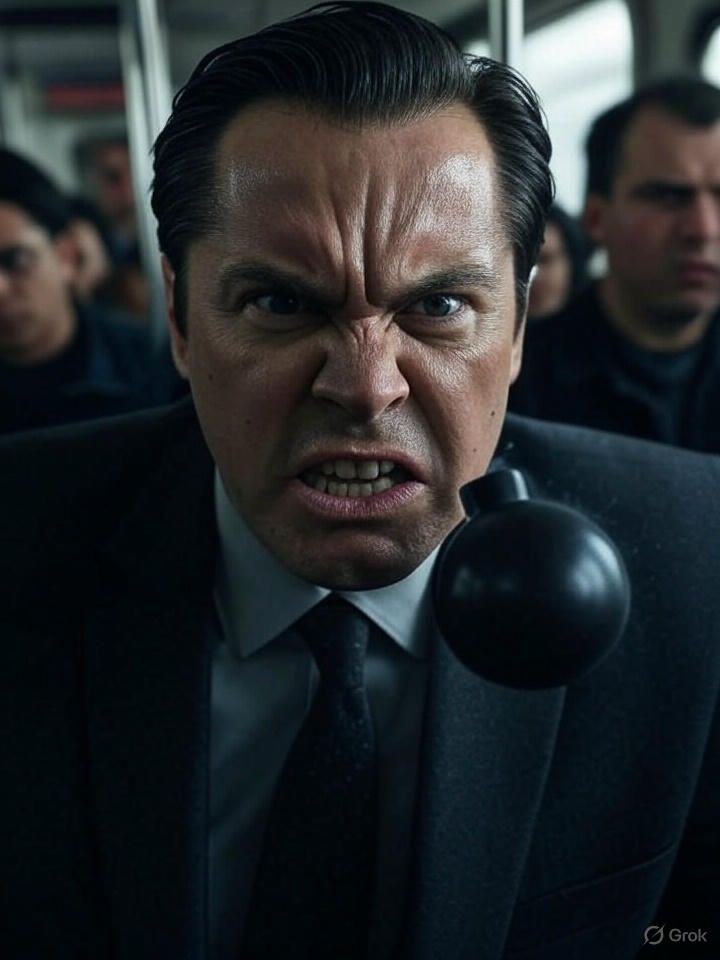}
    \end{minipage}
    \begin{minipage}[b]{0.14\linewidth}
      \centering
      {\small \textbf{Ours}}\\[.5ex]
      \includegraphics[width=\linewidth]{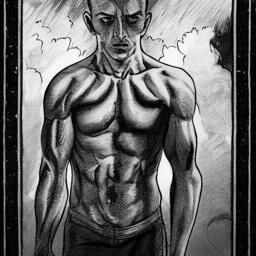}
    \end{minipage}
    \caption{"His anger was a ticking bomb in a crowded subway car."}
    \label{fig:group1}
  \end{subfigure}

  \begin{subfigure}[b]{\linewidth}
  \captionsetup{labelformat=empty}
    \centering
    \includegraphics[width=0.14\linewidth]{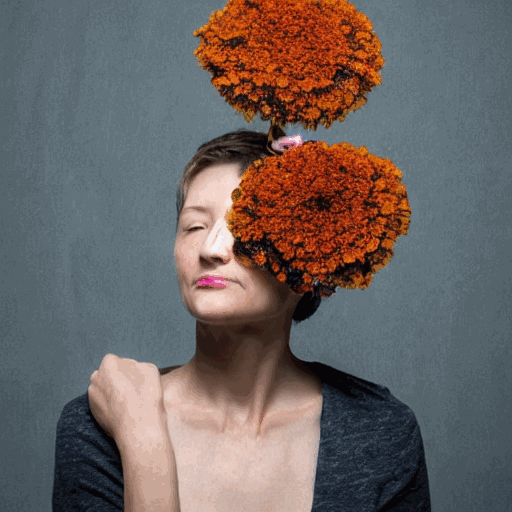}\hfill
    \includegraphics[width=0.14\linewidth]{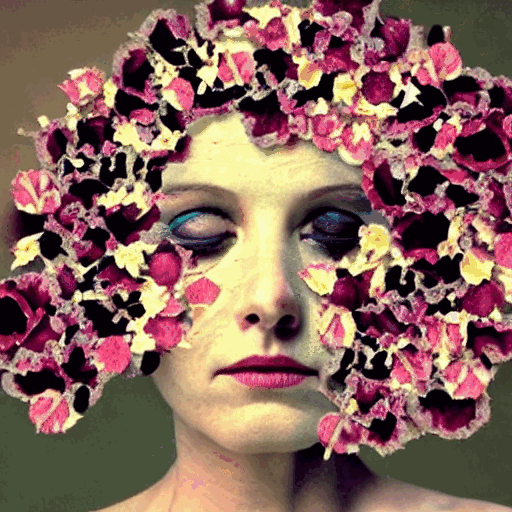}\hfill
    \includegraphics[width=0.14\linewidth]{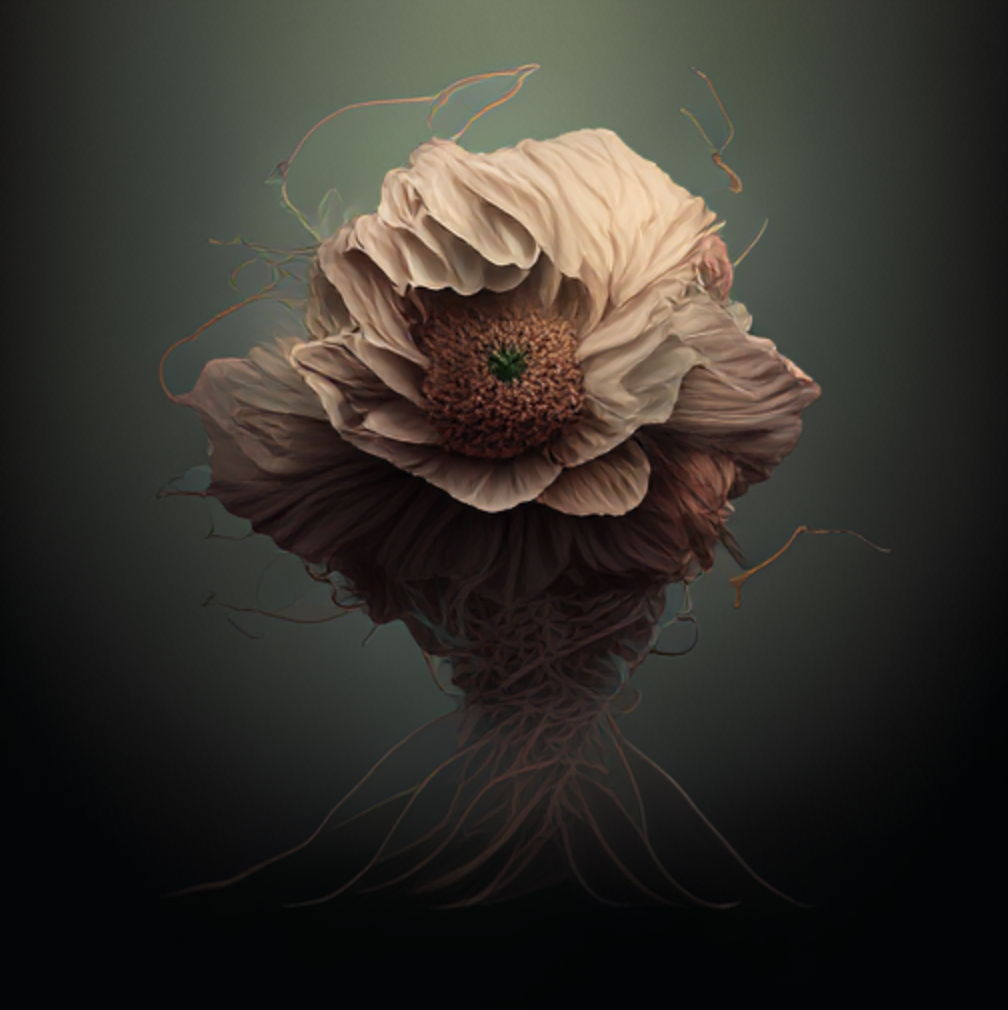}\hfill
    \includegraphics[width=0.14\linewidth]{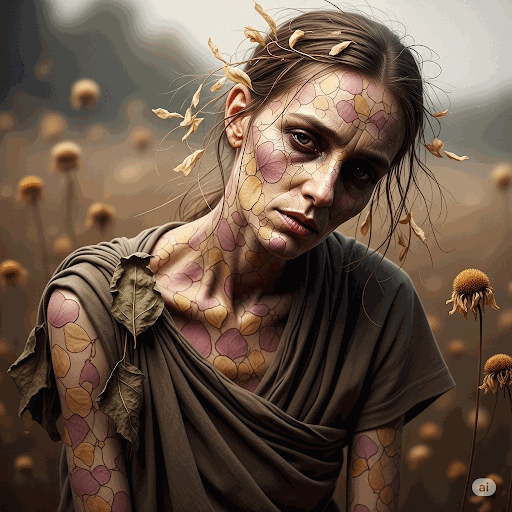}\hfill
    \includegraphics[width=0.14\linewidth]{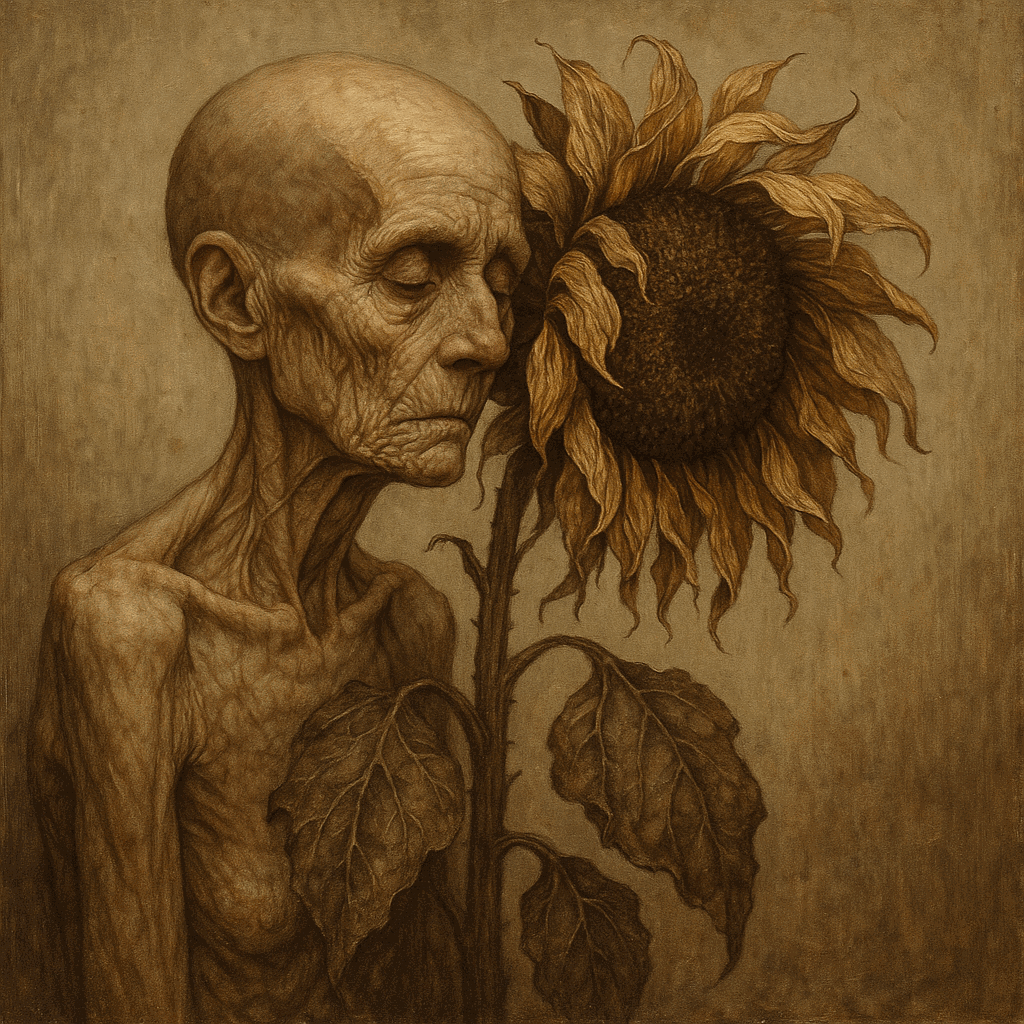}\hfill
    \includegraphics[
        width=0.14\linewidth,     
        height=0.14\linewidth,   
        keepaspectratio=false, 
        clip,           
        trim=0 5cm 0 3cm  
      ]{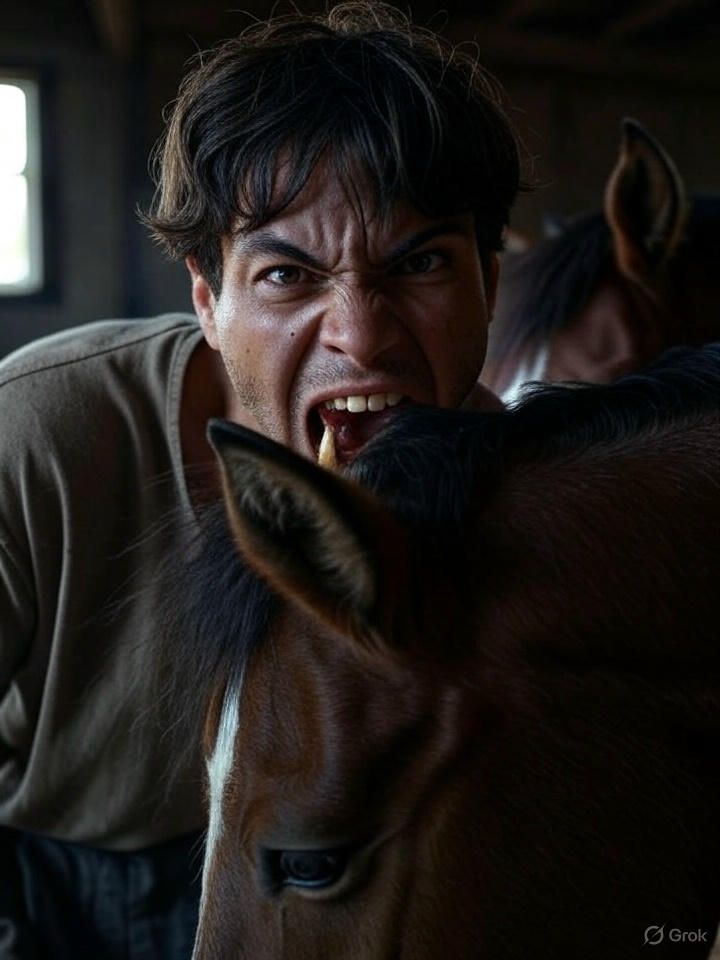}\hfill
    \includegraphics[width=0.14\linewidth]{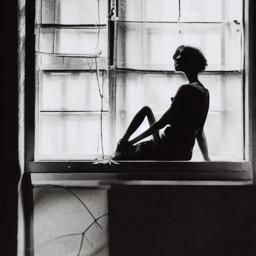}
    \caption{"The cancer made her like a dried flower."}
    \label{fig:group2}
  \end{subfigure}

  \begin{subfigure}[b]{\linewidth}
  \captionsetup{labelformat=empty}
    \centering
    \includegraphics[width=0.14\linewidth]{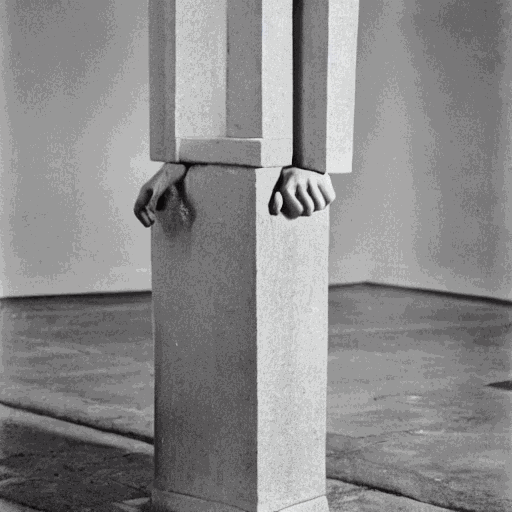}\hfill
    \includegraphics[width=0.14\linewidth]{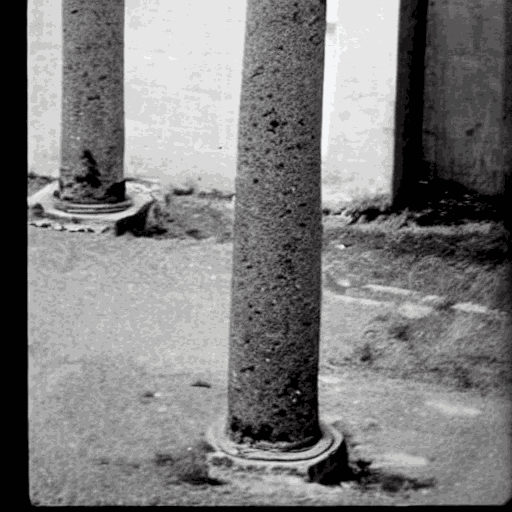}\hfill
    \includegraphics[width=0.14\linewidth]{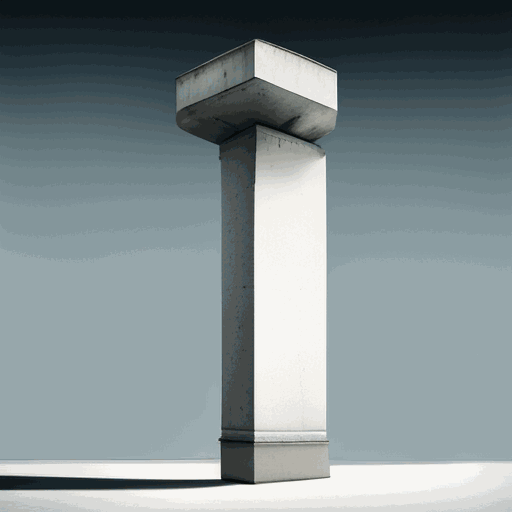}\hfill
    \includegraphics[width=0.14\linewidth]{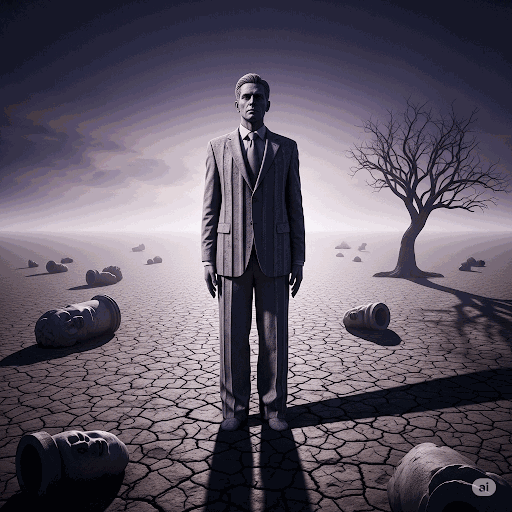}\hfill
    \includegraphics[width=0.14\linewidth]{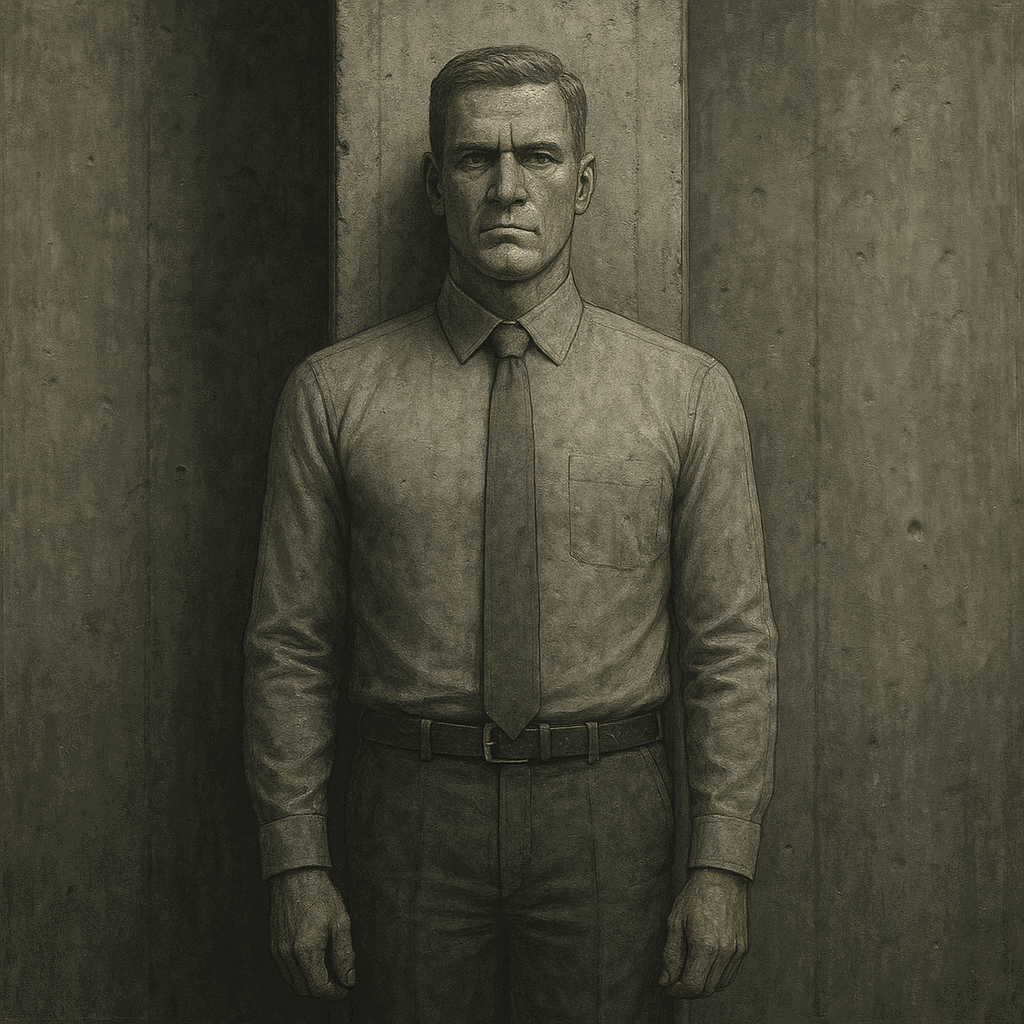}\hfill
    \includegraphics[
        width=0.14\linewidth,     
        height=0.14\linewidth,   
        keepaspectratio=false, 
        clip,           
        trim=0 5cm 0 3cm  
      ]{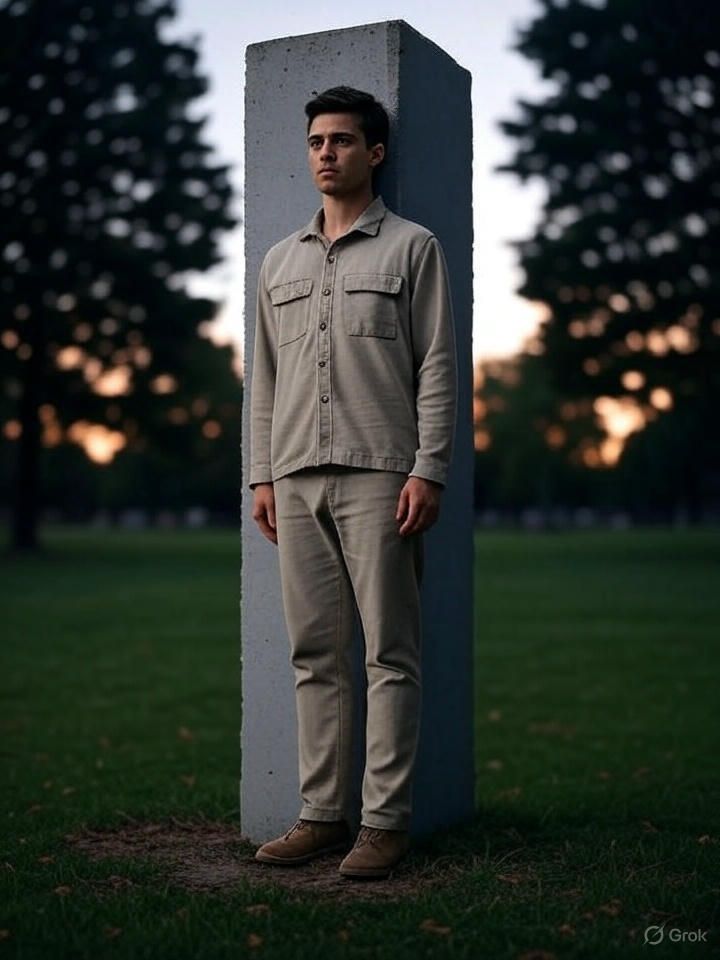}\hfill
    \includegraphics[width=0.14\linewidth]{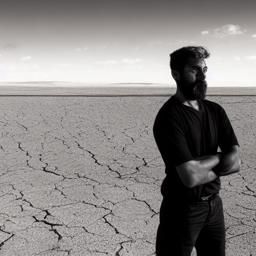}
    \caption{"His posture was a concrete pillar."}
    \label{fig:group2}
  \end{subfigure}

% \vspace{1em}

  \begin{subfigure}[b]{\linewidth}
  \captionsetup{labelformat=empty}
    \centering
    \includegraphics[width=0.14\linewidth]{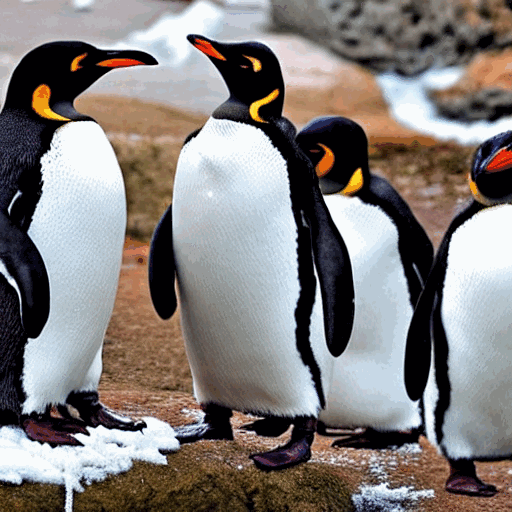}\hfill
    \includegraphics[width=0.14\linewidth]{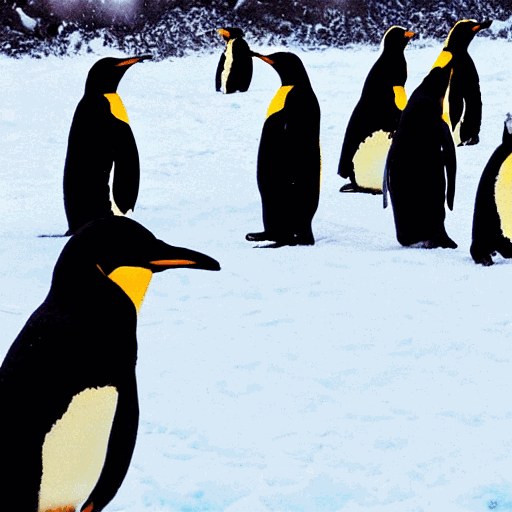}\hfill
    \includegraphics[width=0.14\linewidth]{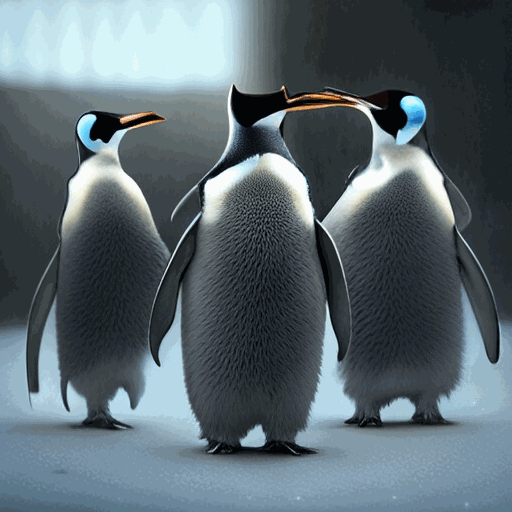}\hfill
    \includegraphics[width=0.14\linewidth]{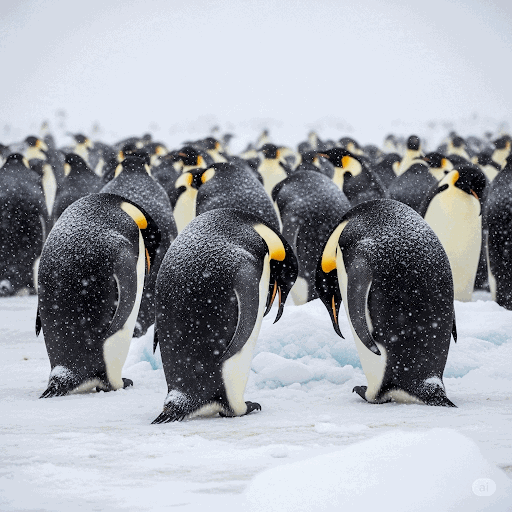}\hfill
    \includegraphics[width=0.14\linewidth]{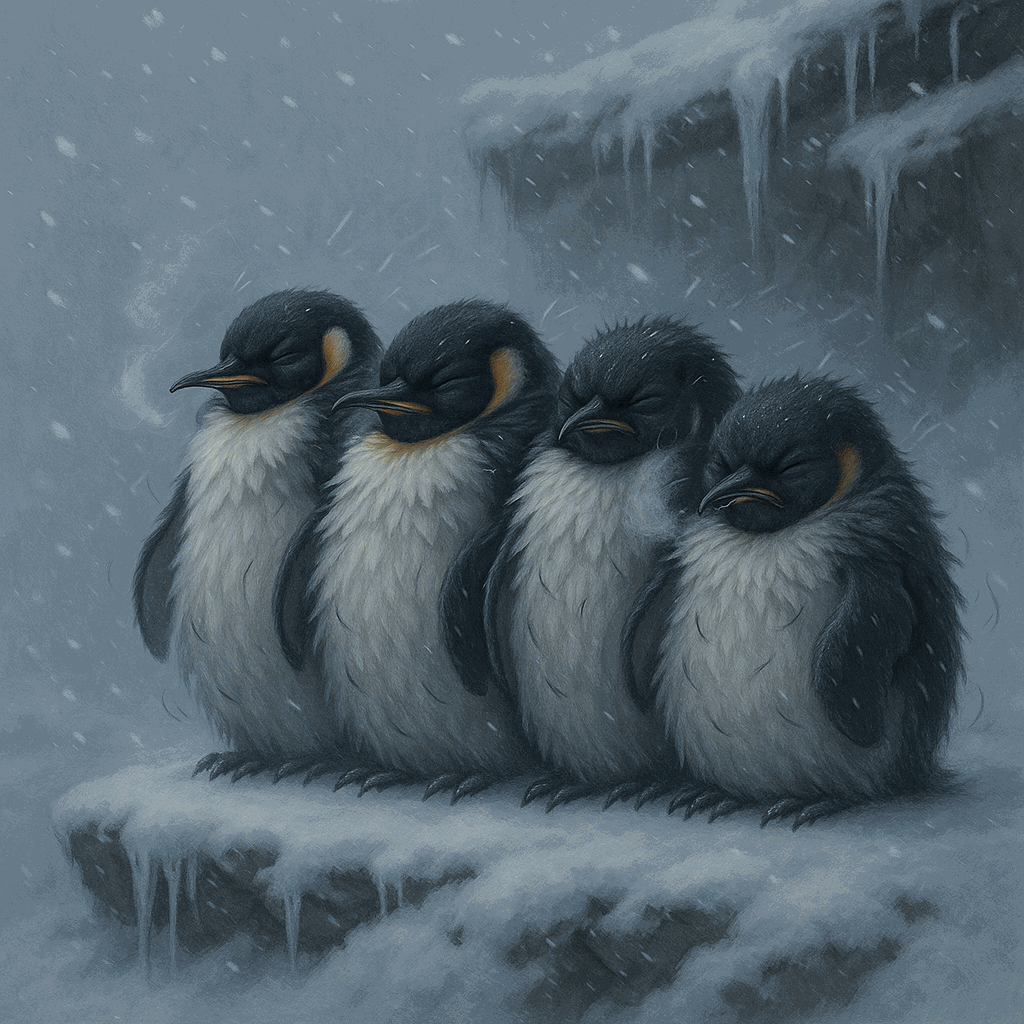}\hfill
    \includegraphics[
        width=0.14\linewidth,     
        height=0.14\linewidth,   
        keepaspectratio=false, 
        clip,           
        trim=0 5cm 0 3cm  
      ]{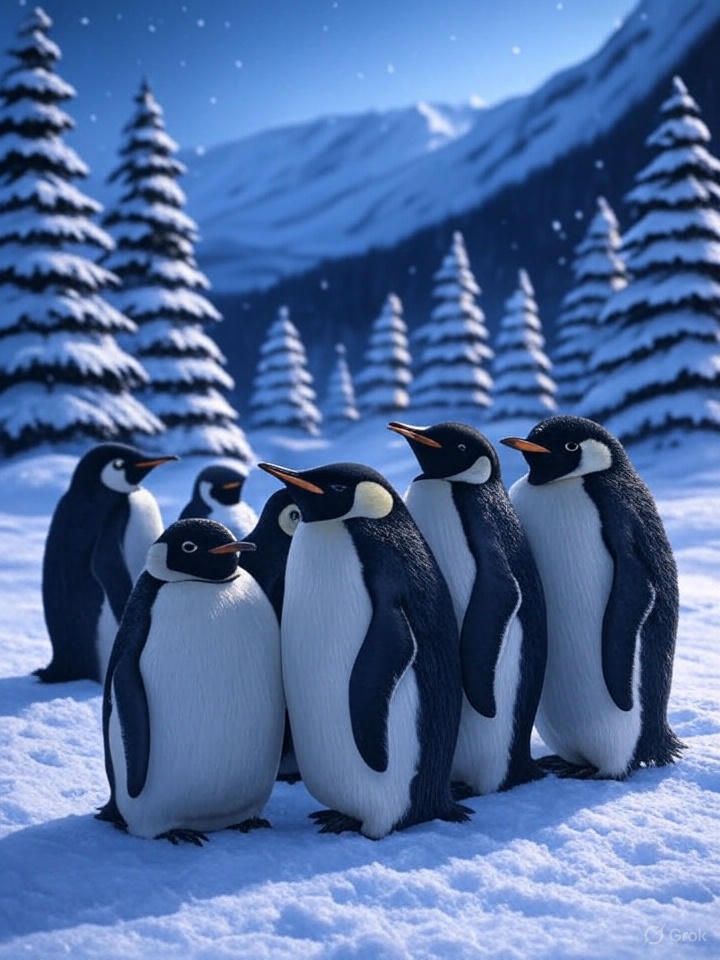}\hfill
    \includegraphics[width=0.14\linewidth]{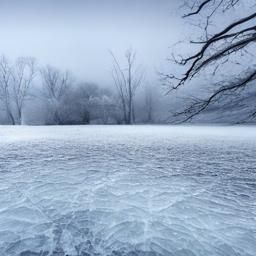}
    \caption{"It's so cold outside, even the penguins are shivering!"}
    \label{fig:group2}
  \end{subfigure}

  % \vspace{1em}

    \begin{subfigure}[b]{\linewidth}
    \captionsetup{labelformat=empty}
    \centering
    \includegraphics[width=0.14\linewidth]{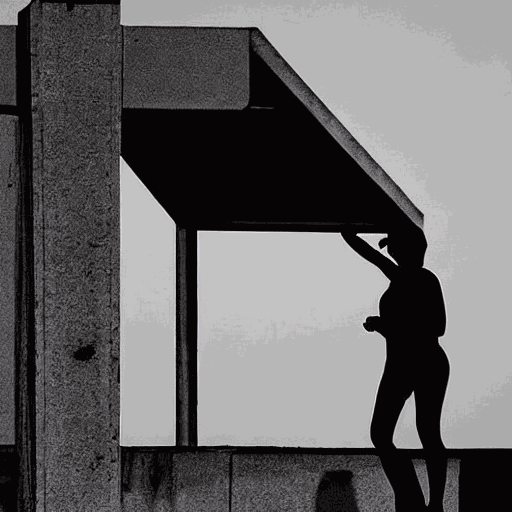}\hfill
    \includegraphics[width=0.14\linewidth]{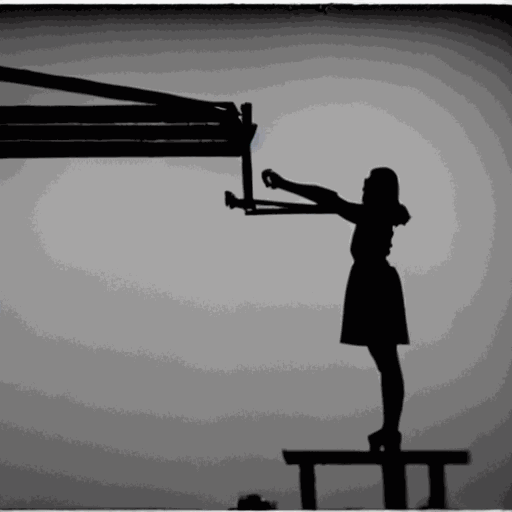}\hfill
    \includegraphics[width=0.14\linewidth]{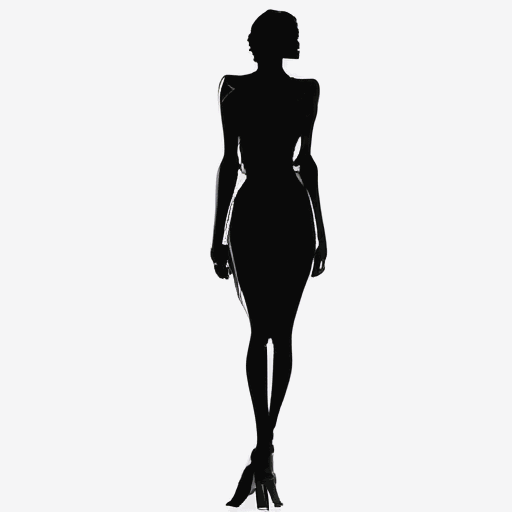}\hfill
    \includegraphics[width=0.14\linewidth]{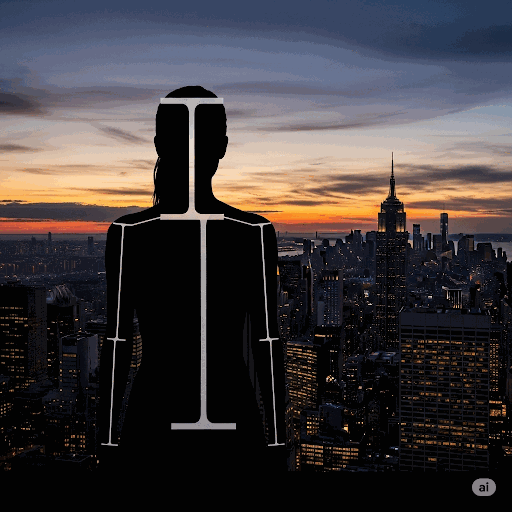}\hfill
    \includegraphics[width=0.14\linewidth]{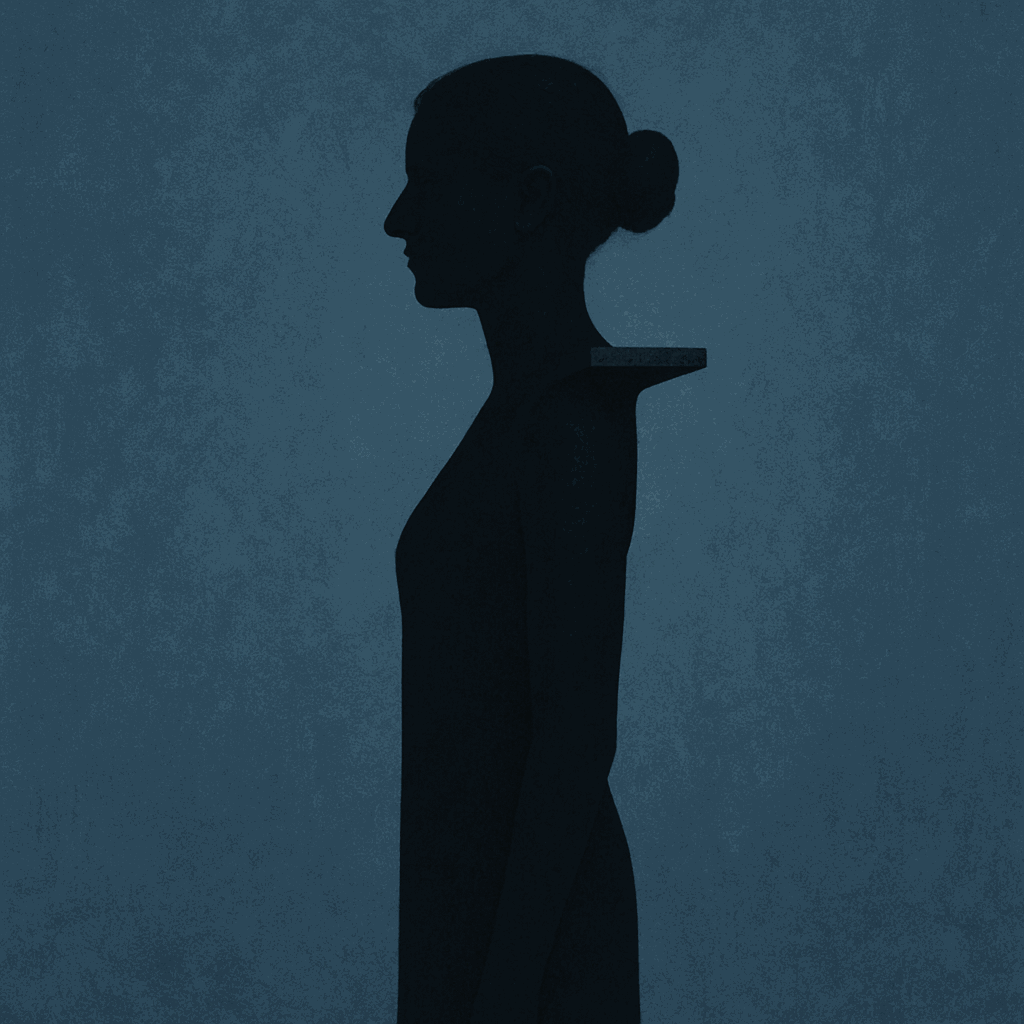}\hfill
    \includegraphics[
        width=0.14\linewidth,     
        height=0.14\linewidth,   
        keepaspectratio=false, 
        clip,           
        trim=0 5cm 0 3cm  
      ]{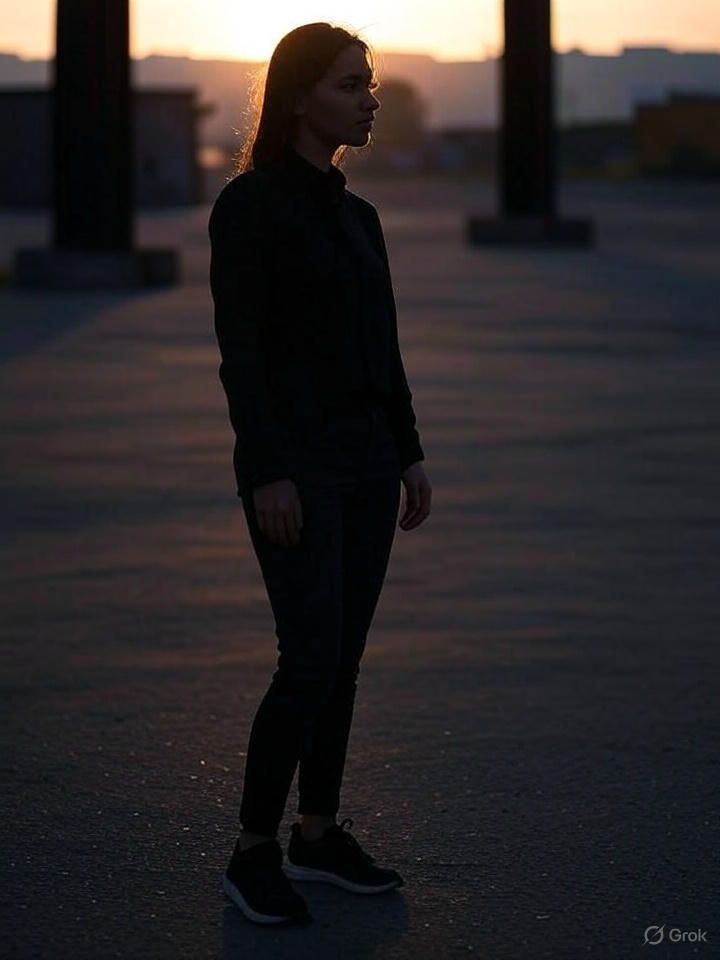}\hfill
    \includegraphics[width=0.14\linewidth]{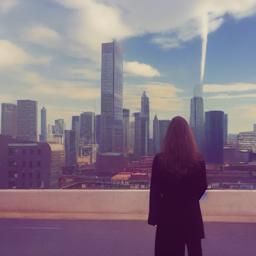}
    \caption{"Her silhouette stood like a steel I-beam."}
    \label{fig:group2}
  \end{subfigure}

  % \vspace{1em}

  \caption{More samples generated by our method compared with other baselines}
  \label{fig:sample1}
\end{figure*}

\begin{figure*}[htbp]
  \centering

  %----- 第二组 subfigure -----

  \begin{subfigure}[b]{\linewidth}
  \captionsetup{labelformat=empty}
    \centering
    \includegraphics[width=0.14\linewidth]{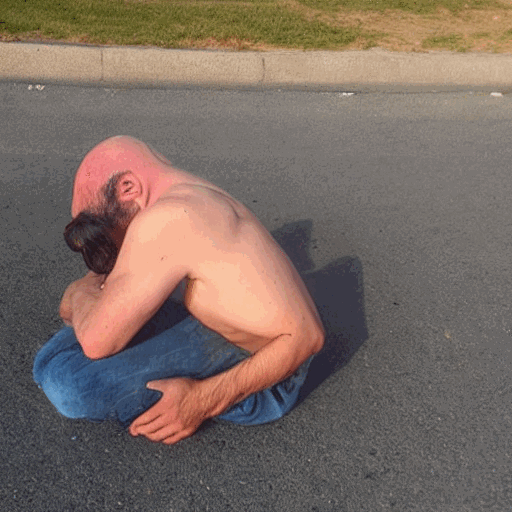}\hfill
    \includegraphics[width=0.14\linewidth]{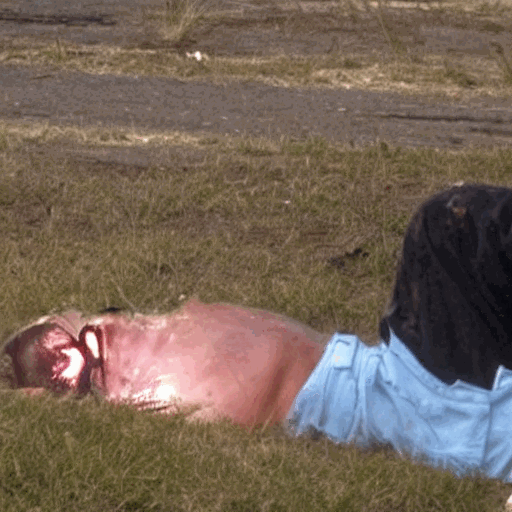}\hfill
    \includegraphics[width=0.14\linewidth]{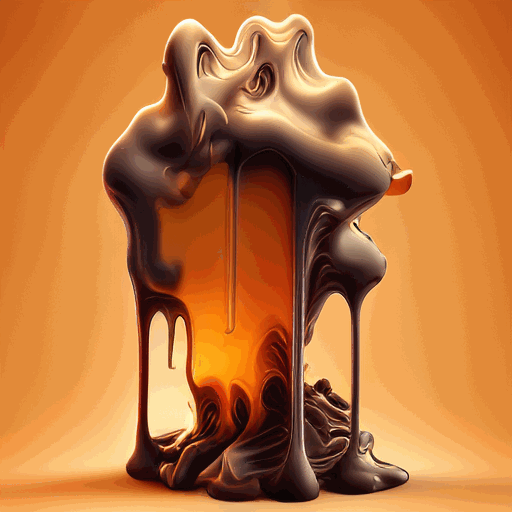}\hfill
    \includegraphics[width=0.14\linewidth]{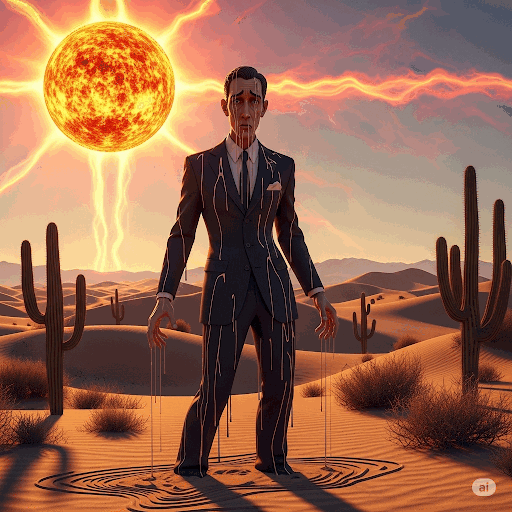}\hfill
    \includegraphics[width=0.14\linewidth]{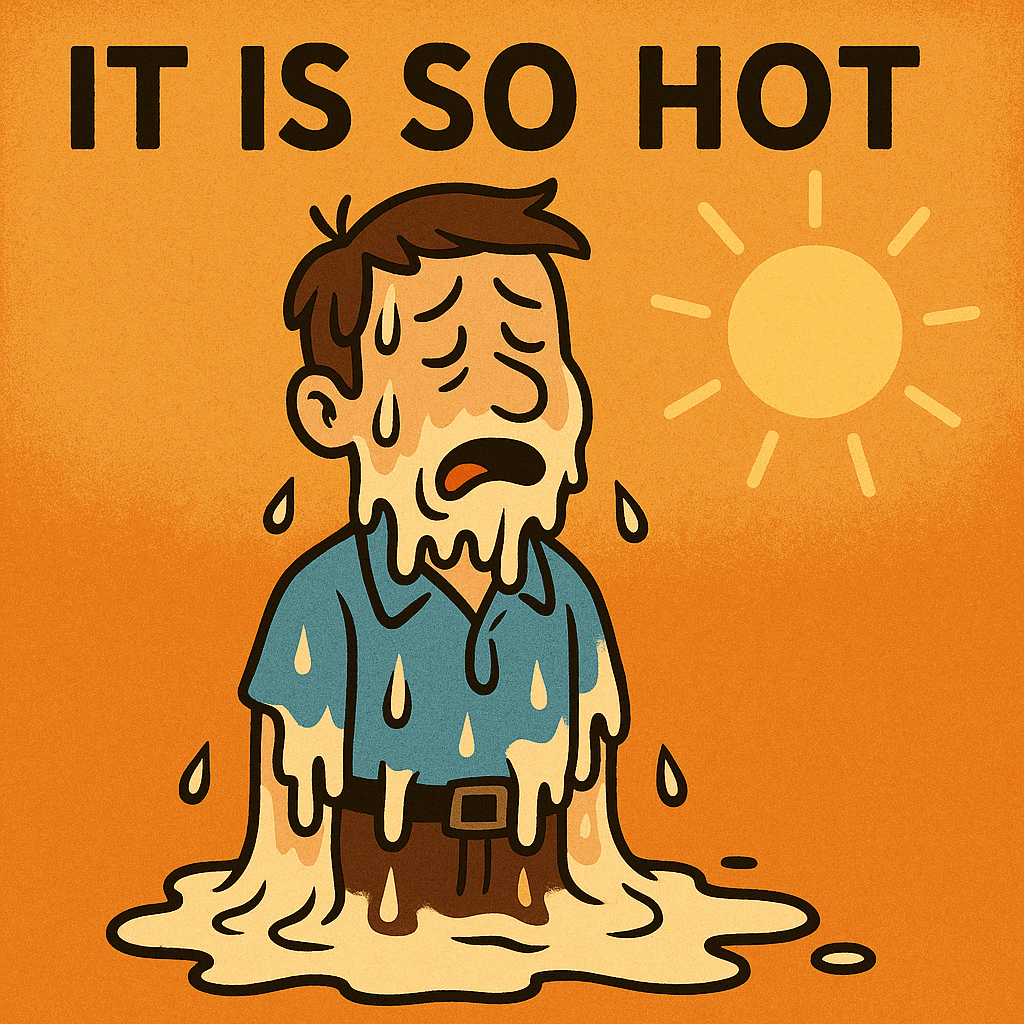}\hfill
    \includegraphics[width=0.14\linewidth]{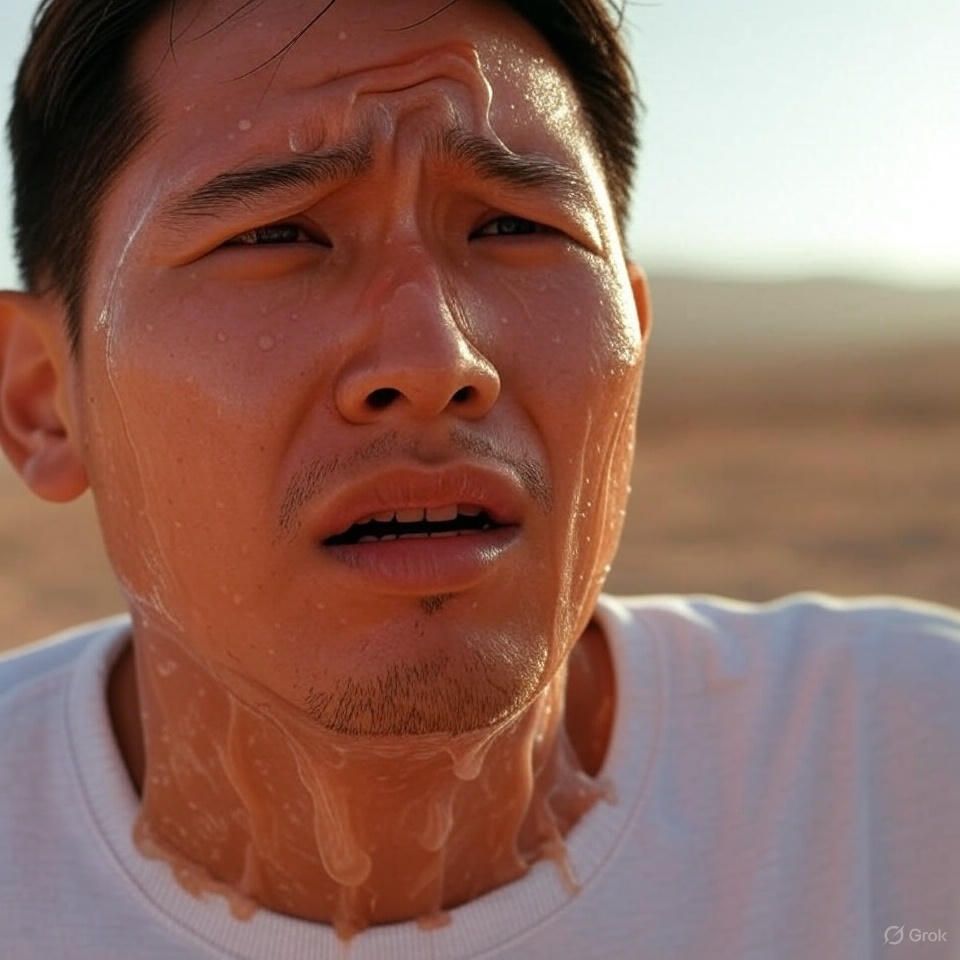}\hfill
    \includegraphics[width=0.14\linewidth]{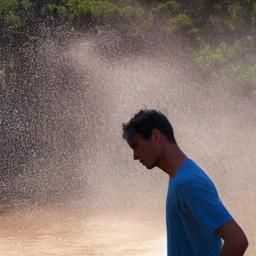}
    \caption{"It is so hot he is melting."}
    \label{fig:group2}
  \end{subfigure}

  \begin{subfigure}[b]{\linewidth}
  \captionsetup{labelformat=empty}
    \centering
    \includegraphics[width=0.14\linewidth]{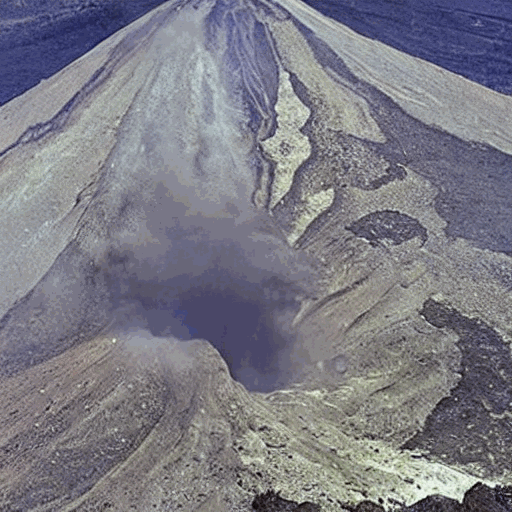}\hfill
    \includegraphics[width=0.14\linewidth]{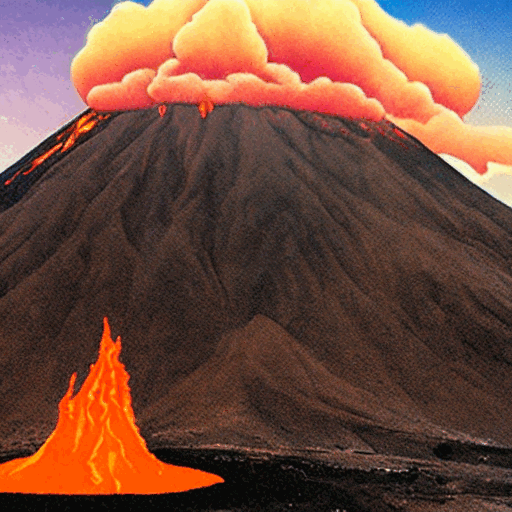}\hfill
    \includegraphics[width=0.14\linewidth]{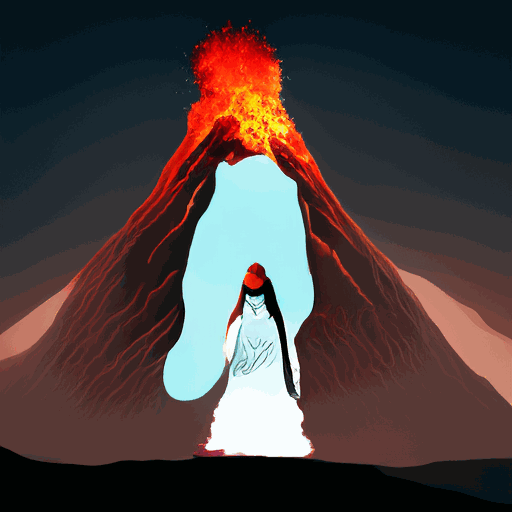}\hfill
    \includegraphics[width=0.14\linewidth]{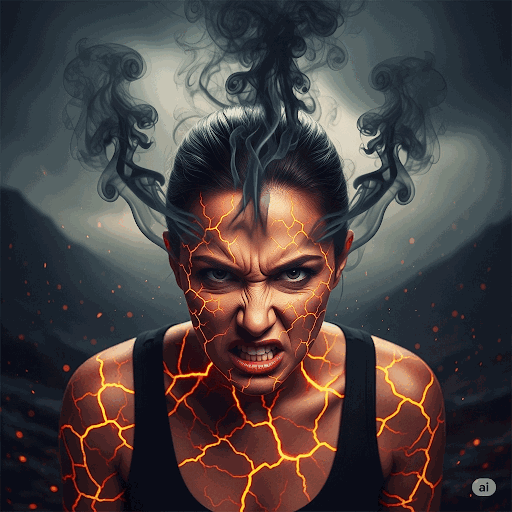}\hfill
    \includegraphics[width=0.14\linewidth]{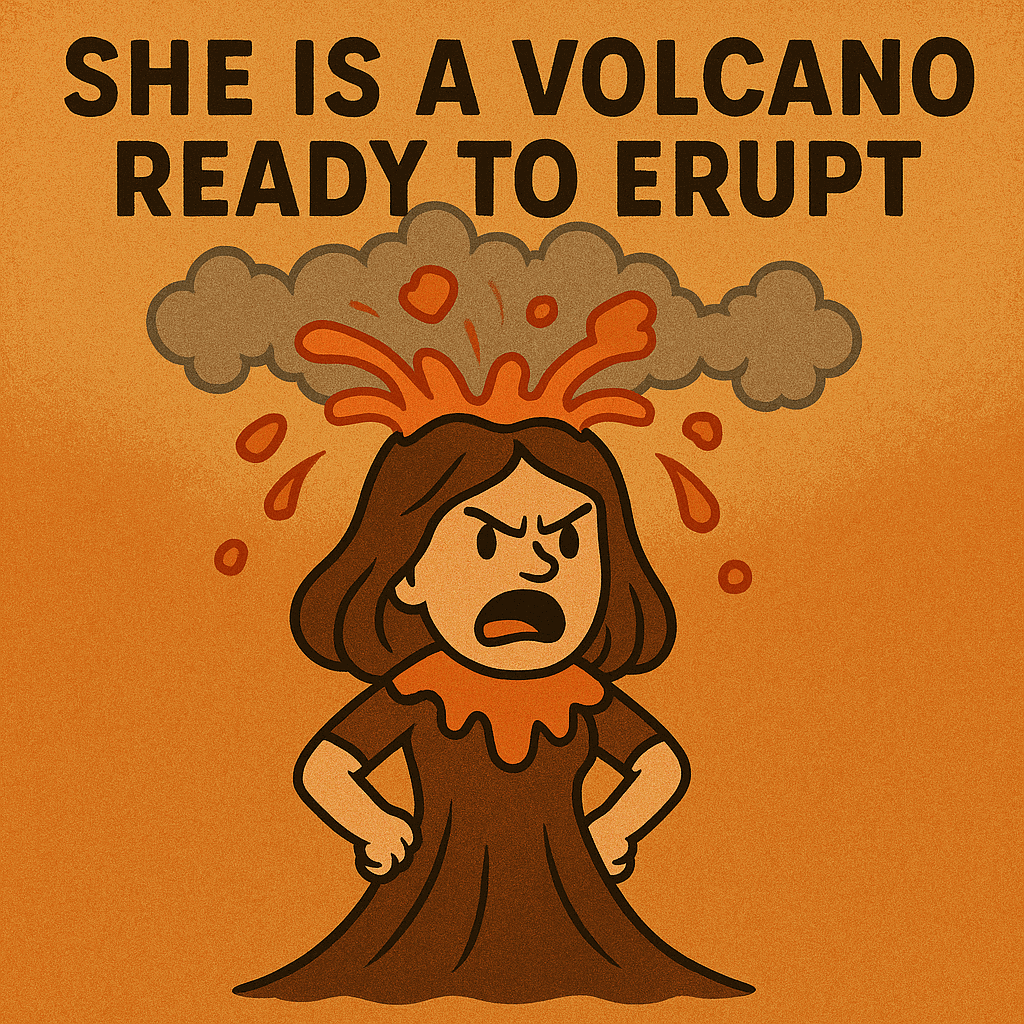}\hfill
    \includegraphics[width=0.14\linewidth]{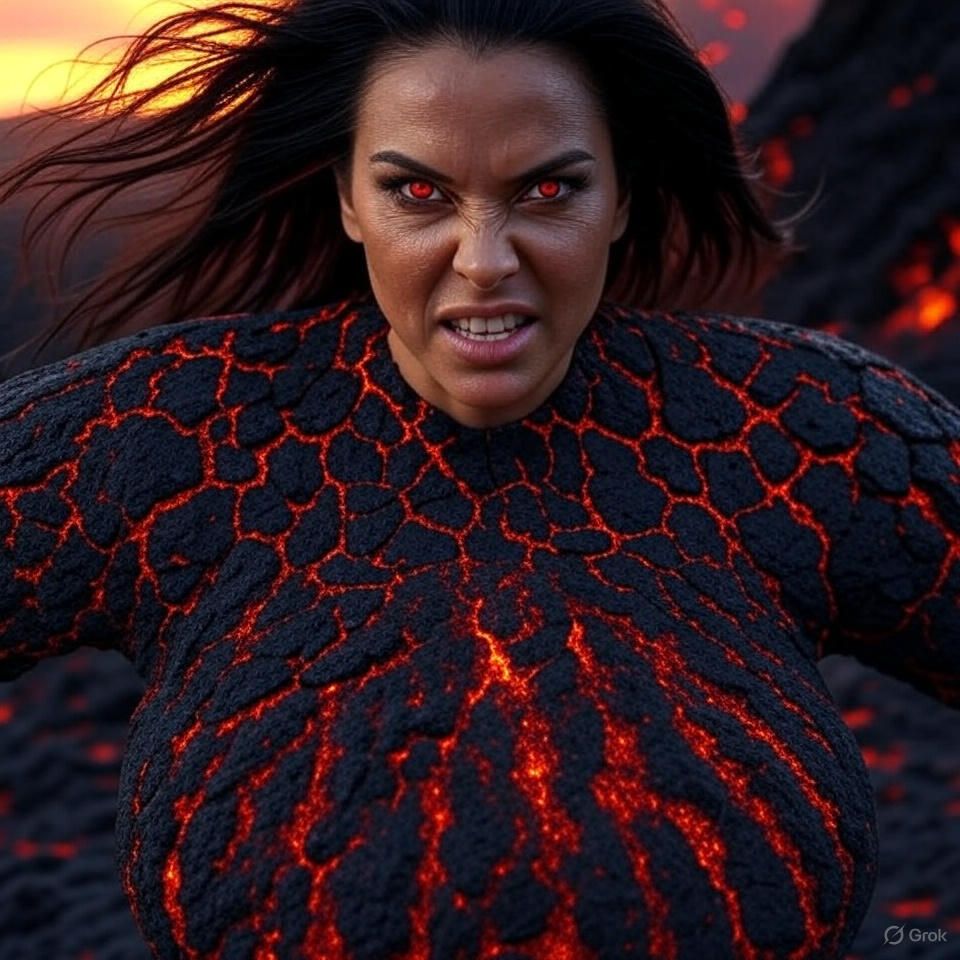}\hfill
    \includegraphics[width=0.14\linewidth]{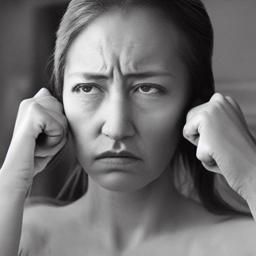}
    \caption{"She is a volcano ready to erupt."}
    \label{fig:group2}
  \end{subfigure}

  \begin{subfigure}[b]{\linewidth}
  \captionsetup{labelformat=empty}
    \centering
    \includegraphics[width=0.14\linewidth]{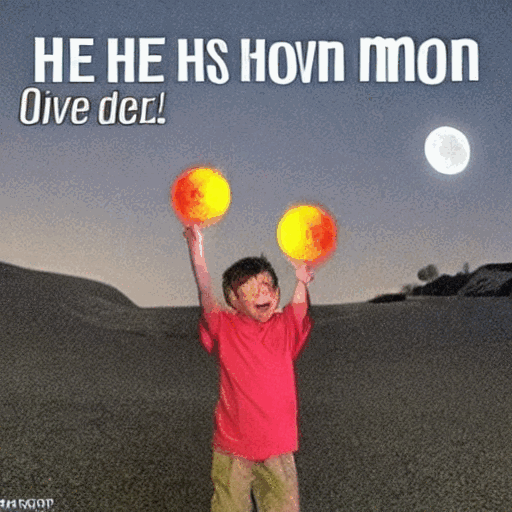}\hfill
    \includegraphics[width=0.14\linewidth]{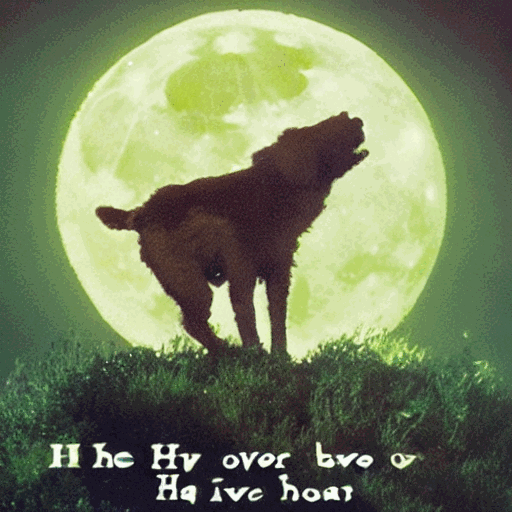}\hfill
    \includegraphics[width=0.14\linewidth]{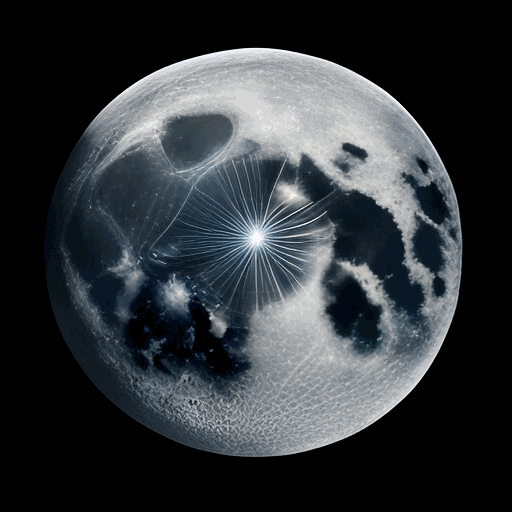}\hfill
    \includegraphics[width=0.14\linewidth]{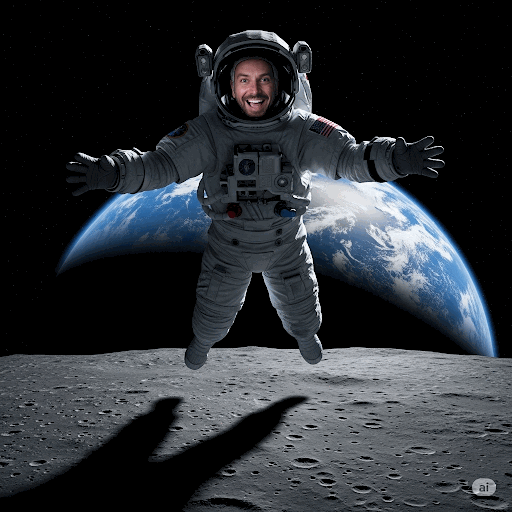}\hfill
    \includegraphics[width=0.14\linewidth]{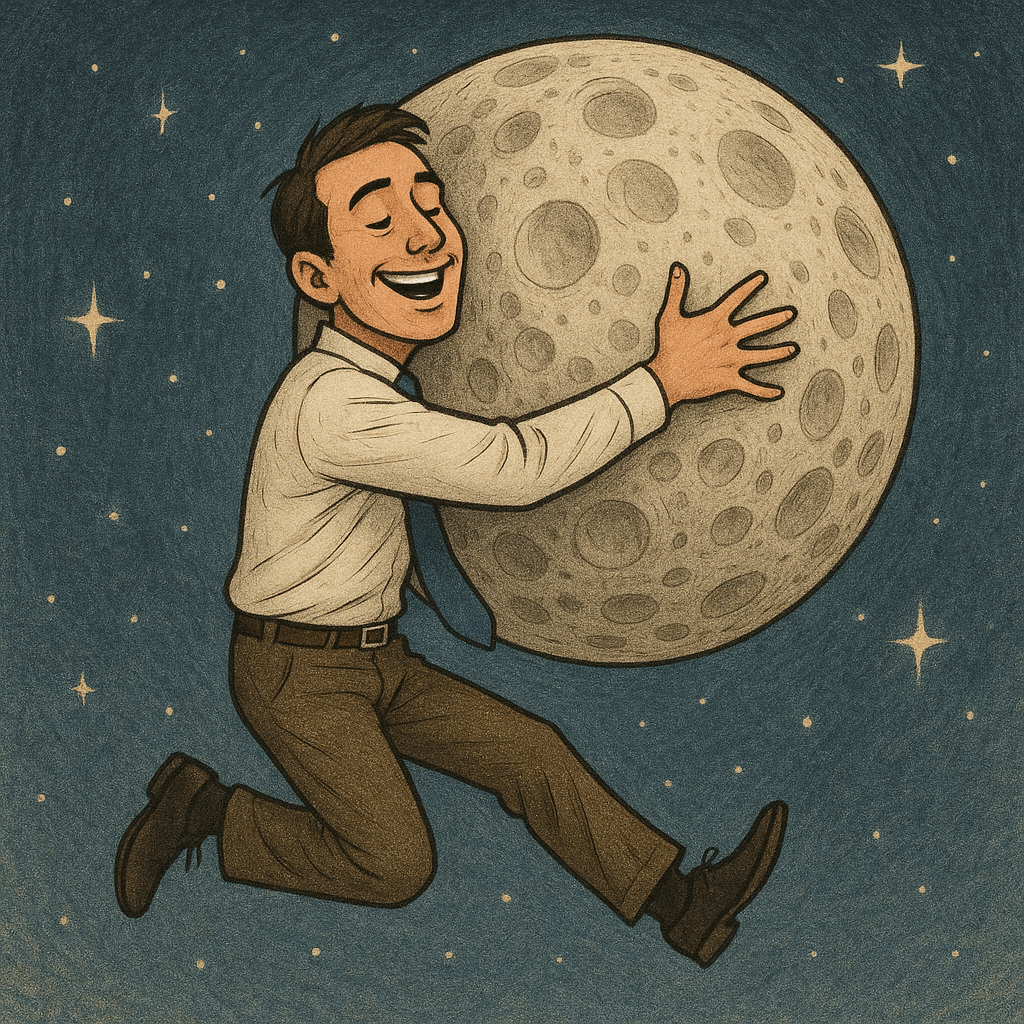}\hfill
    \includegraphics[width=0.14\linewidth]{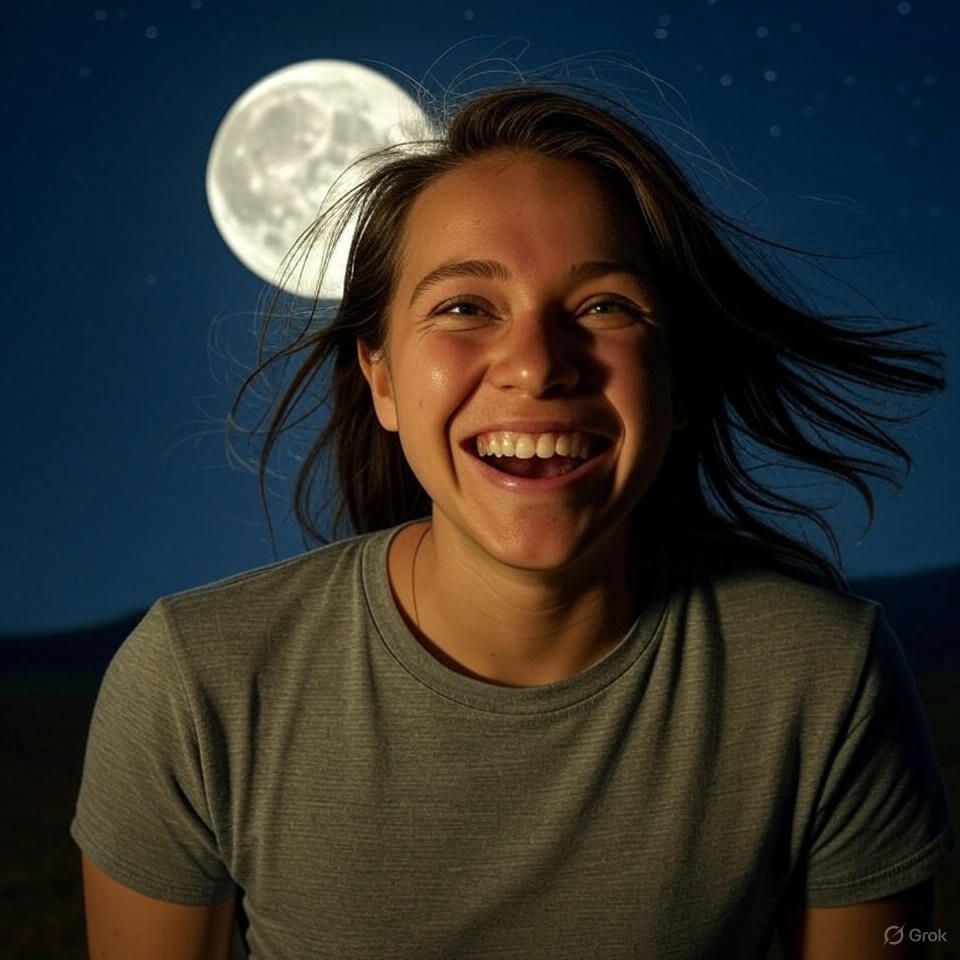}\hfill
    \includegraphics[width=0.14\linewidth]{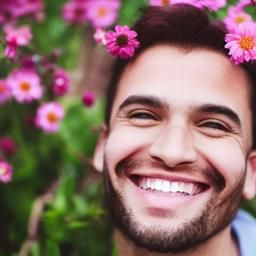}
    \caption{"He is over the moon."}
    \label{fig:group2}
  \end{subfigure}

    \begin{subfigure}[b]{\linewidth}
  \captionsetup{labelformat=empty}
    \centering
    \includegraphics[width=0.14\linewidth]{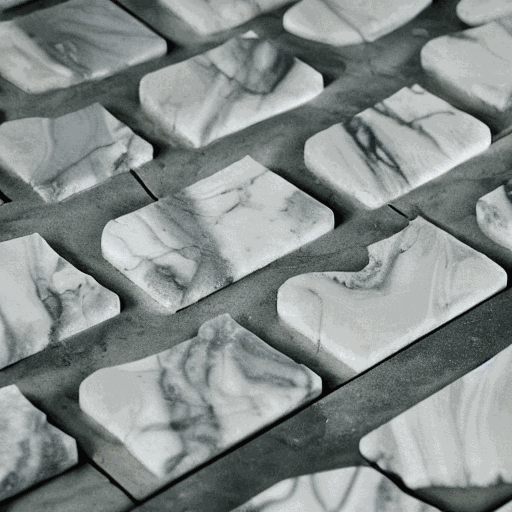}\hfill
    \includegraphics[width=0.14\linewidth]{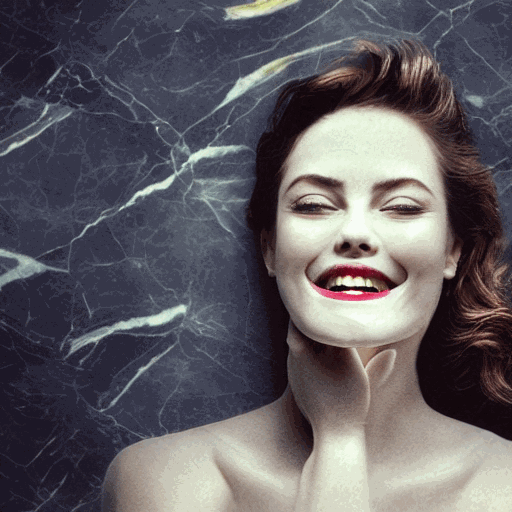}\hfill
    \includegraphics[width=0.14\linewidth]{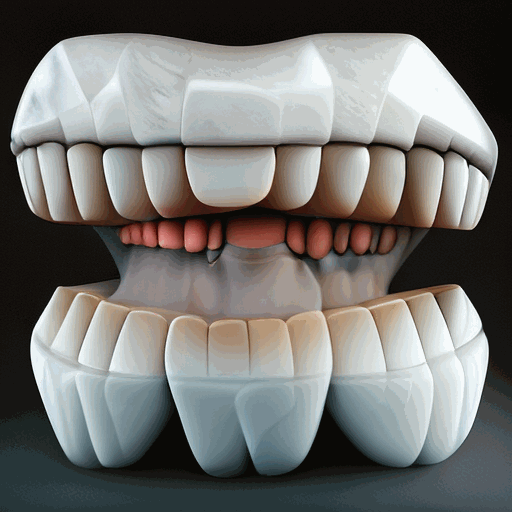}\hfill
    \includegraphics[width=0.14\linewidth]{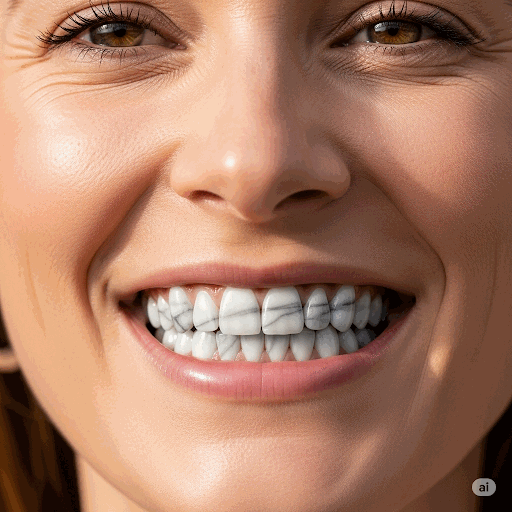}\hfill
    \includegraphics[width=0.14\linewidth]{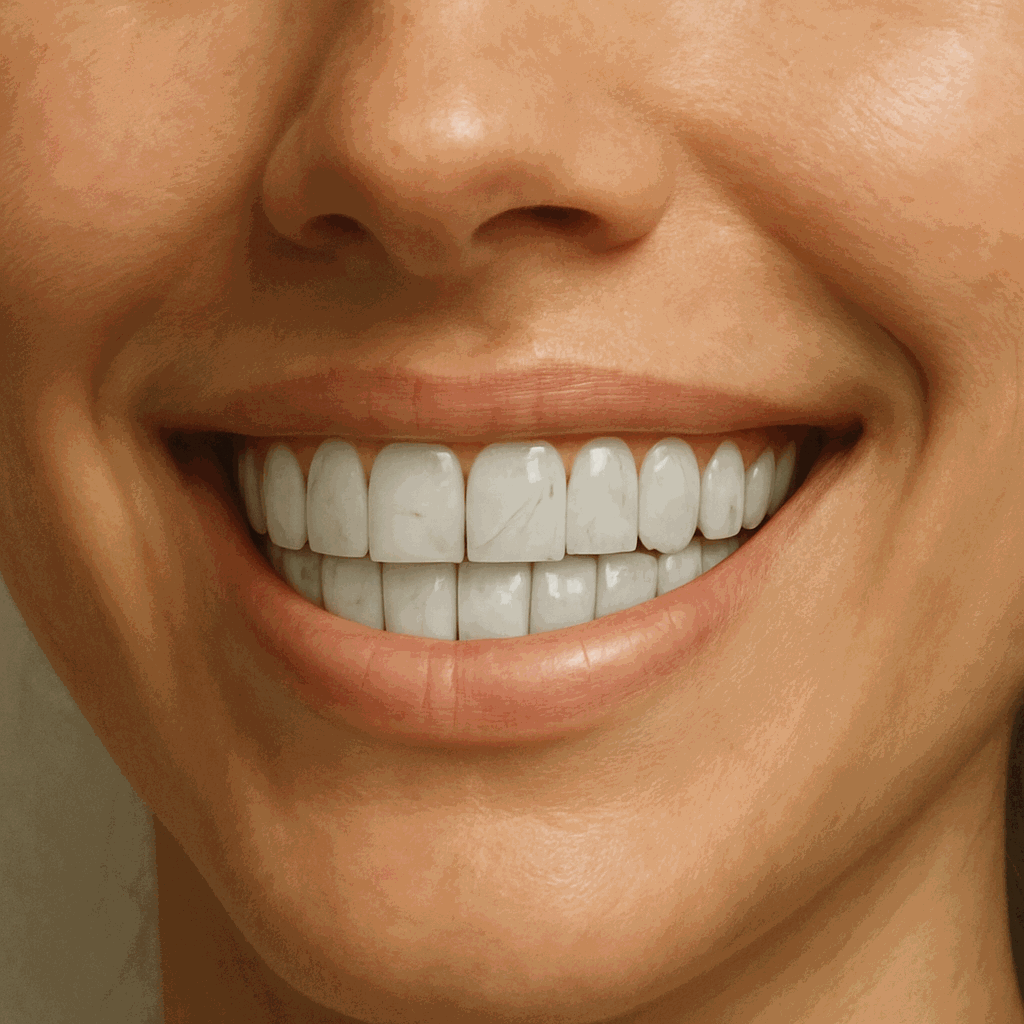}\hfill
    \includegraphics[
        width=0.14\linewidth,     
        height=0.14\linewidth,   
        keepaspectratio=false, 
        clip,           
        trim=0 5cm 0 3cm  
      ]{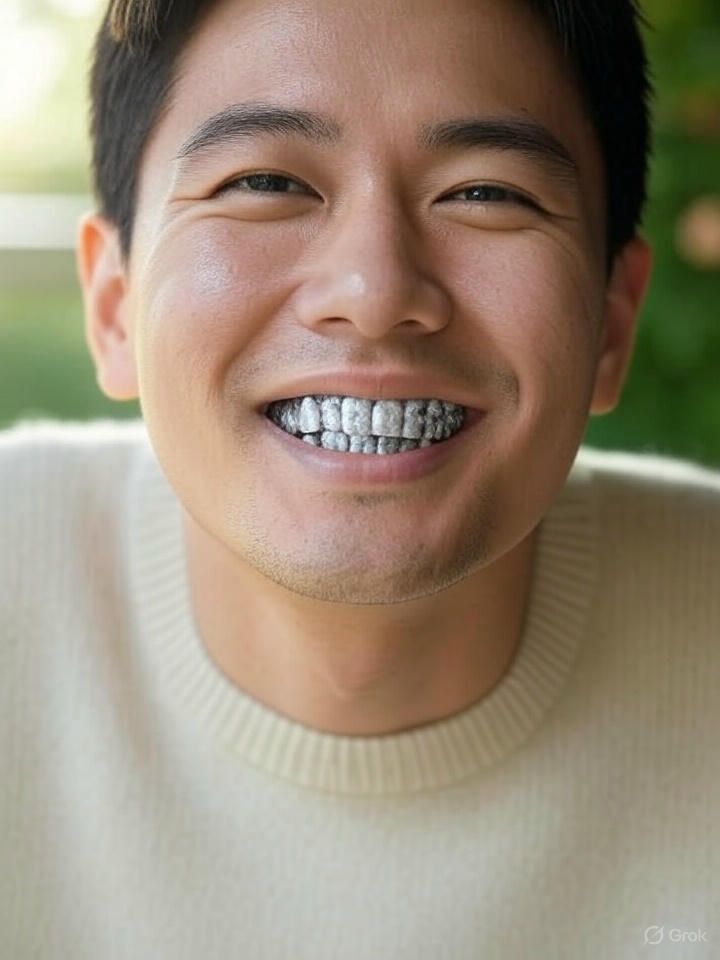}\hfill
    \includegraphics[width=0.14\linewidth]{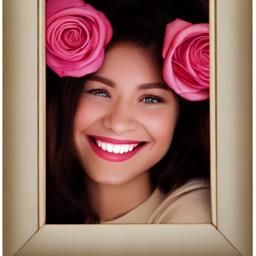}
    \caption{"Her teeth were like polished marble tiles."}
    \label{fig:group2}
  \end{subfigure}

    \begin{subfigure}[b]{\linewidth}
    \captionsetup{labelformat=empty}
    \centering
    \includegraphics[width=0.14\linewidth]{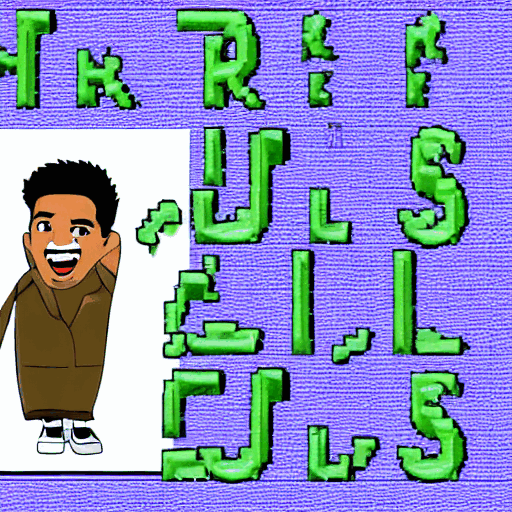}\hfill
    \includegraphics[width=0.14\linewidth]{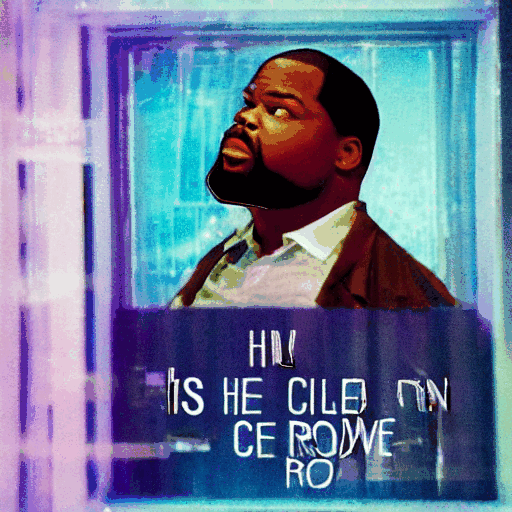}\hfill
    \includegraphics[width=0.14\linewidth]{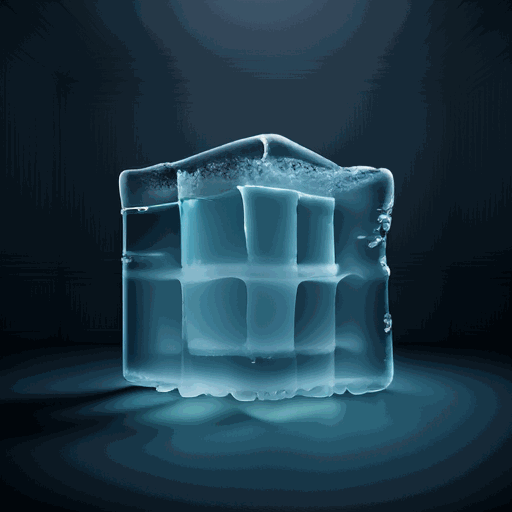}\hfill
    \includegraphics[width=0.14\linewidth]{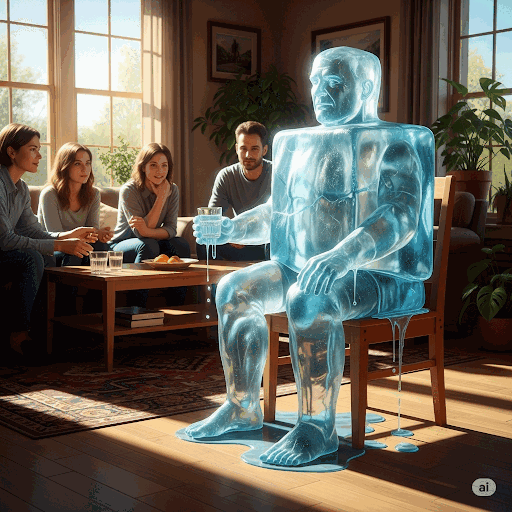}\hfill
    \includegraphics[width=0.14\linewidth]{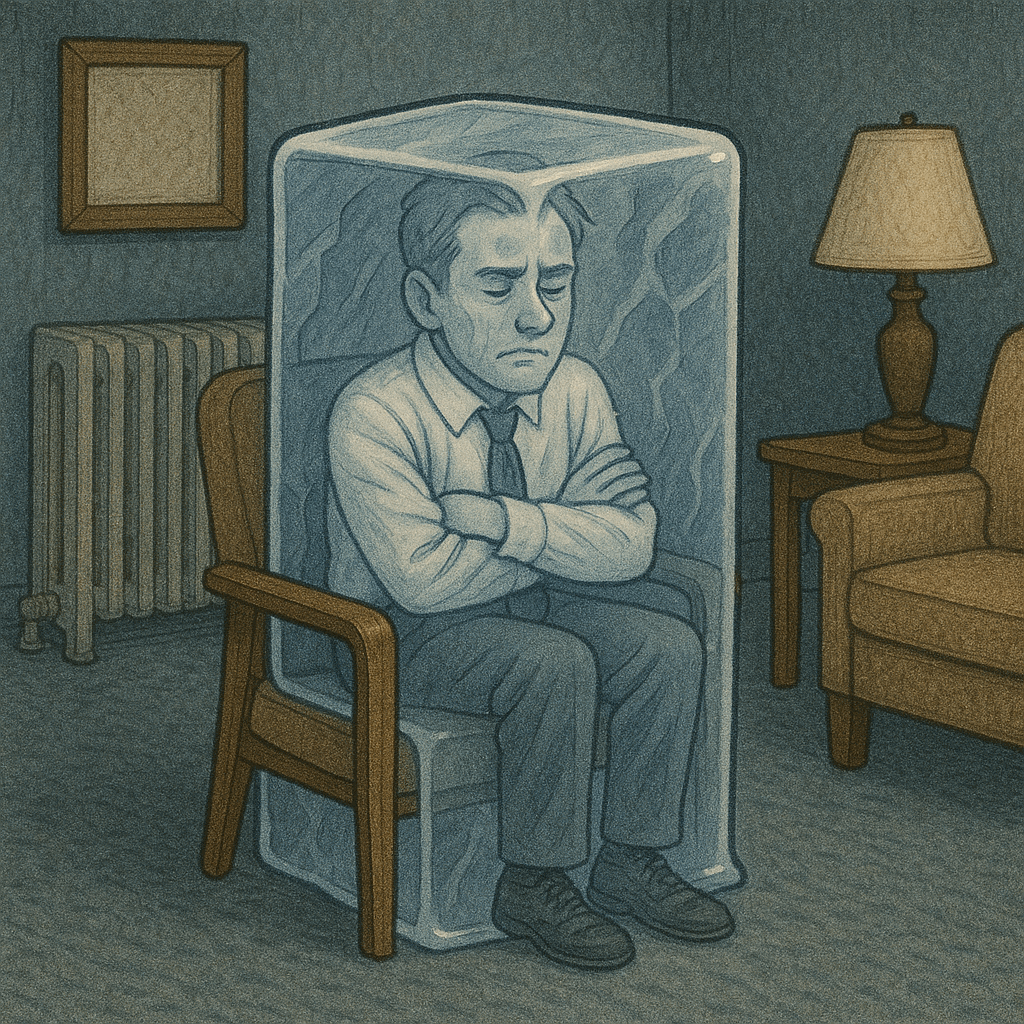}\hfill
    \includegraphics[width=0.14\linewidth]{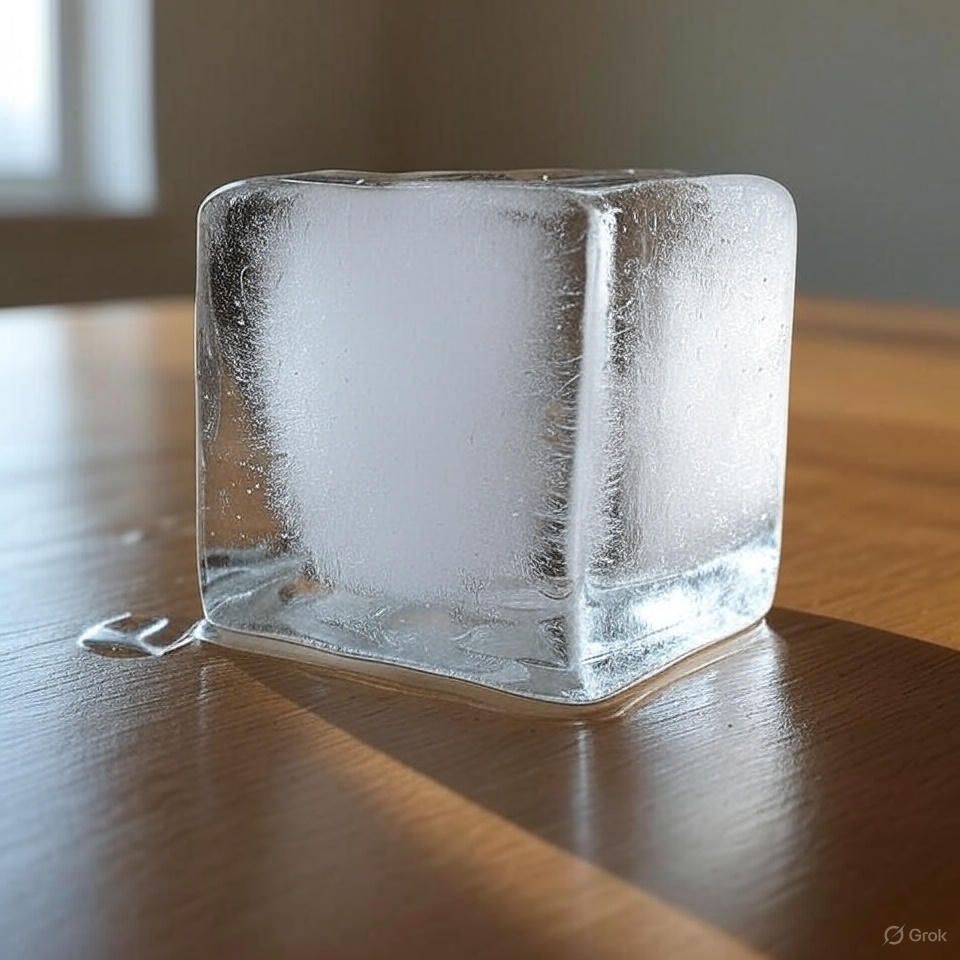}\hfill
    \includegraphics[width=0.14\linewidth]{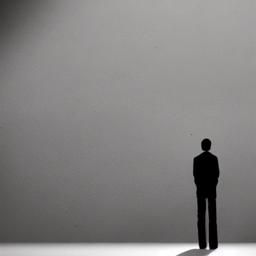}
    \caption{"He is an ice cube in this room."}
    \label{fig:group2}
  \end{subfigure}

  \caption{More samples generated by our method compared with other baselines}
  \label{fig:sample2}
\end{figure*}

As shown in Figure \ref{fig:sample1} and Figure \ref{fig:sample2}, Rhet2Pix produces images with notable visual richness and strong semantic alignment, effectively capturing the rhetorical intent while ensuring that the visuals highlight the correct referents rather than relying on literal or misleading interpretations.

In contrast, baseline models exhibit limitations. Stable Diffusion and DDPO tend to produce generic outputs that match surface-level prompt cues but lack metaphorical understanding. Multimodal LLMs such as GPT-4o and Grok-3 occasionally capture figurative meaning but often struggle to differentiate rhetorical components, resulting in semantically ambiguous depictions.

\newpage

\end{document}